\documentclass[sigconf]{acmart}

\AtBeginMaketitle{%

\def\EDBTISSN{2367-2005}
\def\EDBTISBN{978-3-98318-104-9}
\setcopyright{none}
\copyrightyear{\copyright\ 2026 Copyright held by the owner/author(s).
  Published on \url{OpenProceedings.org} under ISBN \EDBTISBN, series ISSN \EDBTISSN.
  Distribution of this paper is permitted under the terms of the
  Creative Commons license CC-by-nc-nd 4.0}
\acmYear{2026}
\acmDOI{}
\acmISBN{}

\acmConference[EDBT '26]{Extending Database Technology}{24-27 March 2026}{Tampere (Finland)}

\settopmatter{printacmref=false, printccs=false, printfolios=false}

\newsavebox{\ximagebox}
\newlength{\ximageheight}
\newsavebox{\xglyphbox}
\newlength{\xglyphheight}
\newcommand{\xbox}[1]%
  {\savebox{\ximagebox}{#1}%
  \settoheight{\ximageheight}{\usebox{\ximagebox}}%
  \savebox{\xglyphbox}{\color{white}\char32}%
  \settoheight{\xglyphheight}{\usebox{\xglyphbox}}%
  \raisebox{\ximageheight}[0pt][0pt]{\raisebox{-\xglyphheight}[0pt][0pt]{%
    \makebox[0pt][l]{\usebox{\xglyphbox}}}}%
    \usebox{\ximagebox}%
    \raisebox{0pt}[0pt][0pt]{\makebox[0pt][r]{\usebox{\xglyphbox}}}}

\newsavebox{\LogoBox}
\sbox{\LogoBox}{\includegraphics[height=1cm]{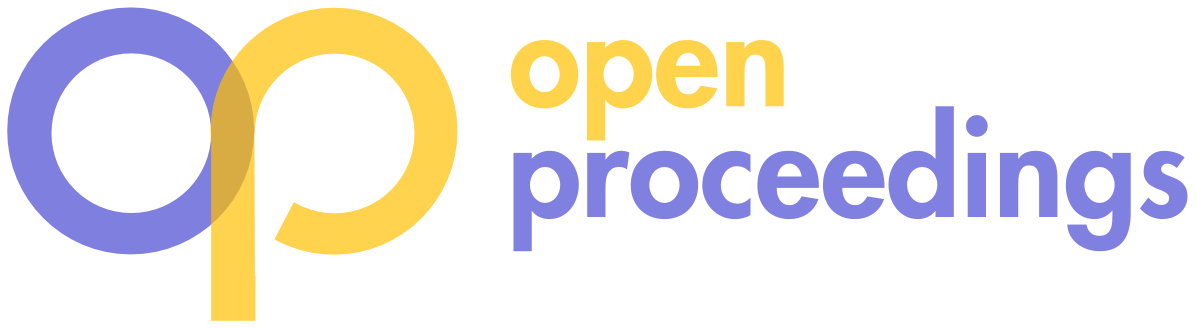}}

\fancypagestyle{firstpagestyle}{%
  \fancyhf{}%
  \fancyfoot{}%
  \fancyhead[L,C]{}
  \fancyhead[R]{\href{https://OpenProceedings.org/}{\raisebox{15pt}[0pt][0pt]{\xbox{\usebox{\LogoBox}}}}}
  }

\hypersetup{%
  pdfcopyright={CC-by-nc-nd},
  pdfpublisher={OpenProceedings.org -- University of Konstanz, Germany},
  pdfcreator={LaTeX with acmart, hyperref, and EDBT modifications},
  pdfvolumenum={29},
  pdfissuenum={2},
  pdfisbn={\EDBTISBN},
  pdfissn={\EDBTISSN}
}

}

\usepackage{amsfonts}
\usepackage[multiple]{footmisc}
\usepackage{booktabs}
\usepackage{dsfont}
\usepackage[utf8]{inputenc}
\usepackage{balance}
\usepackage{siunitx}
\usepackage{multirow}
\usepackage{graphicx}
\usepackage{listings}
\usepackage[ruled,vlined]{algorithm2e}
\usepackage{tcolorbox}
\tcbuselibrary{skins,breakable}
\usepackage{xcolor}
\usepackage{commath}
\usepackage{hyperref}
\usepackage{color}
\usepackage{url}
\usepackage{caption}
\usepackage{alltt}
\usepackage{float}
\usepackage{mathrsfs}
\usepackage{float}
\usepackage{subcaption}
\usepackage{graphicx,fancyvrb}
\usepackage{algpseudocode}
\usepackage{amsmath}
\usepackage{booktabs}

\usepackage{mathtools}
\usepackage{caption}
\usepackage{float}
\usepackage{capt-of}
\usepackage{booktabs}
\usepackage{colortbl}
\usepackage{xcolor}
\usepackage{xfrac}
\usepackage{footnote}

 \usepackage{enumitem}
 \setlist{nosep}
\usepackage{array}
\usepackage{arydshln}

\usepackage{cleveref}

\definecolor{mygreen}{rgb}{0,0.6,0}
\definecolor{myred}{rgb}{0.6,0,0}
\definecolor{mygray}{rgb}{0.5,0.5,0.5}
\definecolor{mymauve}{rgb}{0.58,0,0.82}
\definecolor{myblue}{rgb}{0,0,1}

\usepackage{listings}
\lstnewenvironment{VerbatimText}[1][]{
    
    \lstset{fancyvrb=true,basicstyle=\footnotesize,captionpos=b,xleftmargin=2em,#1}
}{}

\usepackage{booktabs}
\usepackage{pgfplots}
\pgfplotsset{compat=1.7}
\usepackage{pgfplotstable}

\PassOptionsToPackage{hyphens}{url}\usepackage{hyperref}

\usepackage{amsmath}

\newcommand{\ignore}[1]{}
\newcommand{\FuncName}[1]{{\small\ttfamily\scshape #1}}
\definecolor{black}{rgb}{0,0,0}
\definecolor{grey}{rgb}{0.8,0.8,0.8}
\definecolor{red}{rgb}{1,0,0}
\definecolor{green}{rgb}{0,1,0}
\definecolor{darkgreen}{rgb}{0,0.5,0}
\definecolor{darkpurple}{rgb}{0.5,0,0.5}
\definecolor{darkdarkpurple}{rgb}{0.3,0,0.3}
\definecolor{blue}{rgb}{0,0,1}
\definecolor{shadegreen}{rgb}{0.95,1,0.95}
\definecolor{shadeblue}{rgb}{0.95,0.95,1}
\definecolor{shadered}{rgb}{1,0.85,0.85}
\definecolor{shadegrey}{rgb}{0.85,0.85,0.85}
\definecolor{oddRowGrey}{rgb}{0.80,0.80,0.80}
\definecolor{evenRowGrey}{rgb}{0.85,0.85,0.85}
\definecolor{lightpurple}{rgb}{0.88,1.0,1.0}

\newcommand{\model}{\mathcal{M}}
\newcommand{\fairness}{\mathcal{F}}
\newcommand{\budget}{{B}}
\newcommand{\acq}{\data_{exp}}

\newcommand{\RNum}[1]{\uppercase\expandafter{\romannumeral #1\relax}}

\newcommand{\mb}[1]{{\mathbf{#1}}}
\newtheorem{defn}{Definition}[section]

\newtheorem{col}[defn]{Colollary}

\newcommand{\proj}[1]{{\Pi}}
\newcommand{\sel}[1]{{\sigma}}

\newcommand{\cut}[1]{}
\newcommand{\eat}[1]{}

\usepackage[multiple]{footmisc}

\newcommand{\data}{\mb D}
\newcommand{\datapool}{\mathcal{D}}
\newcommand{\dtrain}{\mb D_{train}}
\newcommand{\dtest}{\mb D_{test}}
\newcommand{\dval}{\mb D_{val}}
\newcommand{\datapoint}{\ensuremath{d_i}}

\newcommand{\oparam}{\ensuremath{\theta}^*}
\newcommand{\param}{\ensuremath{{\theta}}}

\SetKwInput{KwInput}{Input}
\SetKwInput{KwOutput}{Output}

\newcommand{\sys}{\textsc{DataSift}\xspace}
\newcommand{\sysinf}{\textsc{Inf}\xspace}
\newcommand{\sysmab}{\textsc{DataSift\_M}\xspace}

\newcommand{\random}{\textsf{Random}\xspace}
\newcommand{\entropy}{\textsf{Entropy}\xspace}
\newcommand{\autodata}{\textsf{AutoData}\xspace}
\newcommand{\slicetuner}{\textsf{SliceTuner}\xspace}
\newcommand{\infs}{\textsf{Inf}\xspace}

\begin{document}


\title{Selective Data Expansion for Model Performance}

\author{Jahid Hasan}
\orcid{0009-0007-0998-7613}
\affiliation{%
  \institution{Purdue University}
  \city{West Lafayette}
  \state{IN}
  \country{USA}
}
\email{hasan89@purdue.edu}

\author{Romila Pradhan}
\orcid{0009-0003-1538-0472}
\affiliation{%
  \institution{Purdue University}
  \city{West Lafayette}
  \state{IN}
  \country{USA}
}

\email{rpradhan@purdue.edu}
\renewcommand{\shortauthors}{Jahid et al.}


\begin{abstract}
  Machine learning systems are increasingly being used in critical decision-making, such as healthcare, finance, and criminal justice. Concerns around system fairness have resulted in several mitigation techniques that emphasize the need for high-quality data to ensure fairer decisions. However, the role of earlier stages in machine learning pipelines in addressing model unfairness remains underexplored.
  We focus on the task of \textit{selective data expansion}--- carefully selecting additional data points from a data pool to add to the training data--- to rapidly improve the fairness of a model learned on the modified data while also preserving model accuracy. Since not all points in the pool are equally beneficial, we propose \sys, a data expansion framework that combines data valuation with multi-armed bandits to identify the most valuable data points for including into training data. Unlike prior methods that mitigate unfairness through data transformation or post-/in-processing, \sys addresses the problem directly by selecting the right data. Over successive iterations, \sys selects a partition, samples a batch of points leveraging influence functions, evaluates their impact, and updates partition utilities accordingly. 
  Empirical evaluation of \sys on multiple real-world and synthetic datasets shows that model unfairness is mostly resolved by including as few as $4\%$ of additional data with at most $2.6\%$ reduction (and as much as $27.4\%$ increase) in accuracy.
\end{abstract}


\maketitle
\section{Introduction}
\label{sec:intro}

With the increasingly widespread use of machine learning (ML) in consequential decision-making domains, such as criminal justice, healthcare, and housing, 
there is an ever-growing need to ensure that ML-based systems do not have adverse implications on society. Designed carefully, these systems can potentially eliminate
the unwanted aspects of human decision-making (e.g., biased decisions). However, irresponsible uses of artificial intelligence (AI) can lead to and reinforce systemic biases, discrimination and other abuses often reflected in
the underlying training data~\cite{propublica-machine-bias,fb-housing,self-driving-cars}. 
%
A number of fairness metrics have been introduced over the past decade to quantify the discrimination exhibited by the ML-based systems~\cite{verma2018fairness,mehrabi_survey_2022}. Simultaneously, the need to debias these systems has given rise to several bias mitigation techniques (see~\cite{10.1145/3616865,mehrabi_survey_2022} for recent surveys on fairness and bias in machine learning).

The focus on data-centric AI has spotlighted the importance of \textit{data quality} in improving machine learning performance~\cite{doi:10.1137/1.9781611977653.ch106,DBLP:journals/corr/abs-2112-06439,10.1007/s00778-022-00775-9,10.1145/3580305.3599553}. In a recent survey, data science practitioners have reported feeling the most control over their data during earlier stages in the data science pipeline such as data collection and curation~\cite{10.1145/3290605.3300830}. Several recent works have particularly recognized the significance of augmenting the training data from accessible data sources, termed as \textit{data expansion}, for improving machine learning model performance~\cite{NEURIPS2023_0506ad3d,pmlr-v139-liu21f}. Data expansion differs from the task of data acquisition which focuses on identifying additional data that should be acquired from existing/new sources~\cite{li_data_2021,10.1145/3654934,chai_selective_2022, 10.1145/3448016.3452792}. Most of the existing works on data acquisition to enhance machine learning models focus on traditional performance metrics (e.g., model accuracy, loss)~\cite{chai_selective_2022,li_data_2021, 10.1145/3448016.3452792} or model confidence~\cite{10.1145/3654934}. 
%
Each of these solutions is tailored to the specific performance metric (e.g., distance-based clustering for accuracy, sampling by decision boundary distance for confidence, power-law region~\cite{mishra2021accelerating} observing for loss) and is therefore not directly applicable to the equally important metric of model fairness where recent research has highlighted the importance of different stages of data science pipelines in combating fairness violations~\cite{10.1145/3510003.3510057,10.1145/3468264.3468536}.

\begin{figure}[t]
    \centering
    \includegraphics[scale=0.30]{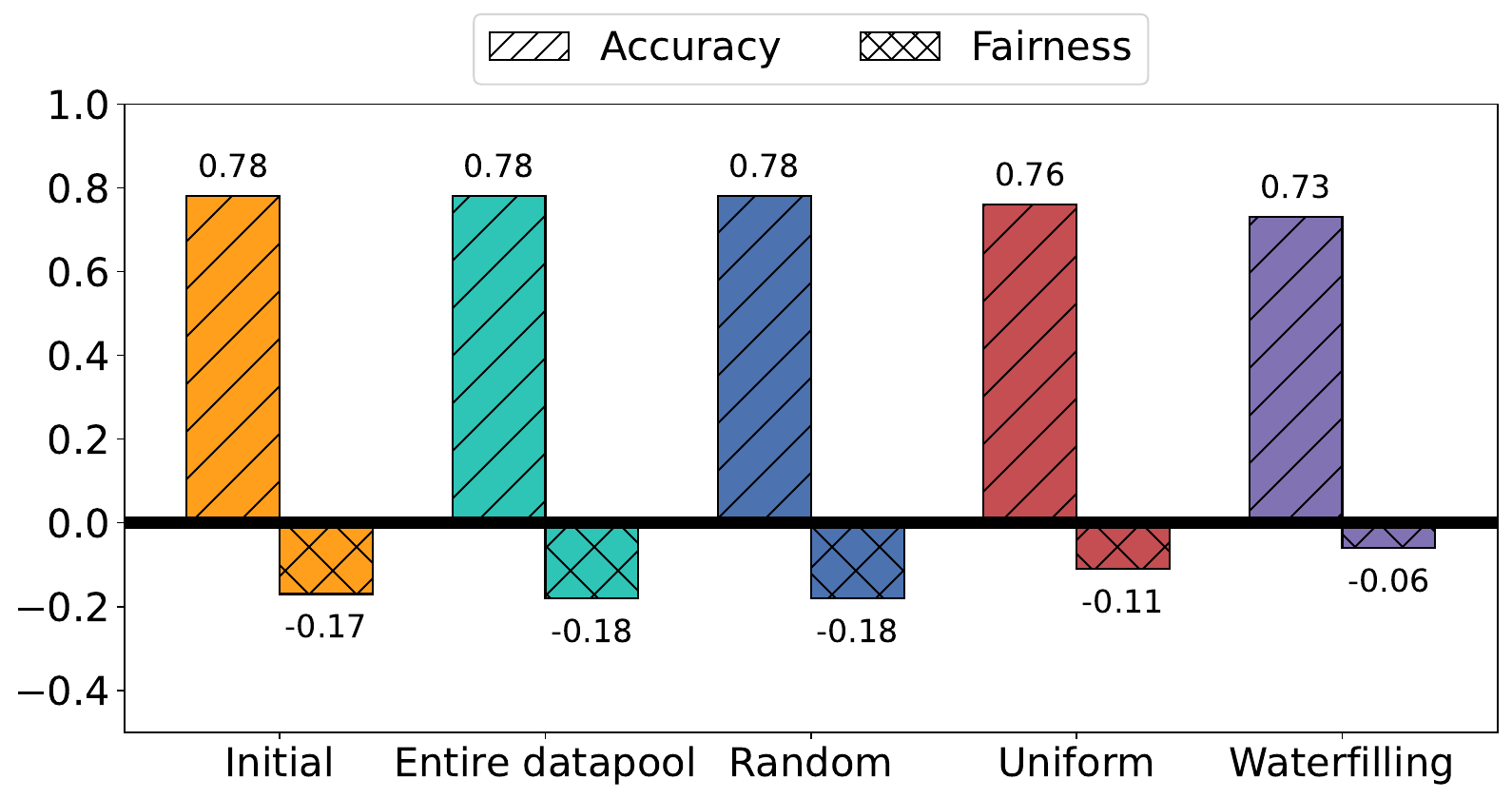}
    \caption{Impact of different data selection strategies on model fairness (statistical parity) and classification accuracy. Lower magnitude of fairness indicates a less biased and more equitable model overall.
    }
    \label{fig:ex}
\end{figure}

The
following example illustrates the need for adding useful additional data to mitigate unfairness in model decisions.
\begin{example}
Consider an organization that deploys an automated algorithm to predict whether an individual qualifies as high income—a critical step for determining eligibility for social benefits reserved for higher-income groups. Although the model demonstrates high predictive accuracy on historical data, it systematically underestimates the likelihood of high income for female applicants, despite qualifications comparable to those of male applicants. This results in a documented demographic disparity of nearly $17\%$ in outcomes~\cite{mehrabi_survey_2022}, raising serious concerns about fairness and equity in resource allocation.
To address these inequities, a data scientist working in the organization considers expanding the training set with additional samples from the AdultIncome dataset~\cite{dua2017uci} (see Table~\ref{tab:datasets_summary} and Section~\ref{sec:dataset_description} for more details) with the same schema. However, the pool itself is demographically skewed: women are underrepresented overall, and especially so in the high-income category. As shown in Figure~\ref{fig:ex}, simply adding the entire data pool perpetuates the same disparity. This behavior persists when the initial training dataset is expanded with  $20\%$ of data chosen randomly from the data pool. With an equal selection of data from the two demographic groups (Uniform), we observe a modest improvement in fairness but at the cost of accuracy. However, an even deliberate selection of data that enforces demographic balance in the training set (Waterfilling) results in a $65\%$ improvement in fairness (but causes a sharp $6\%$ decline in accuracy).
These results highlight the inherent trade-off between fairness and accuracy, and the importance of selective data expansion that preserves both. 
(Indeed, Figure~\ref{fig: AdultIncome_DataSift} shows that through strategic reordering of data selection, \sys achieves substantial fairness improvements over the baselines while preserving accuracy, using only 8\% of the data pool.
Further details are provided in Section~\ref{sec:exp}).
\label{ex:example}
\end{example}
\begin{figure*}[t]
    \centering   \includegraphics[width=\textwidth]{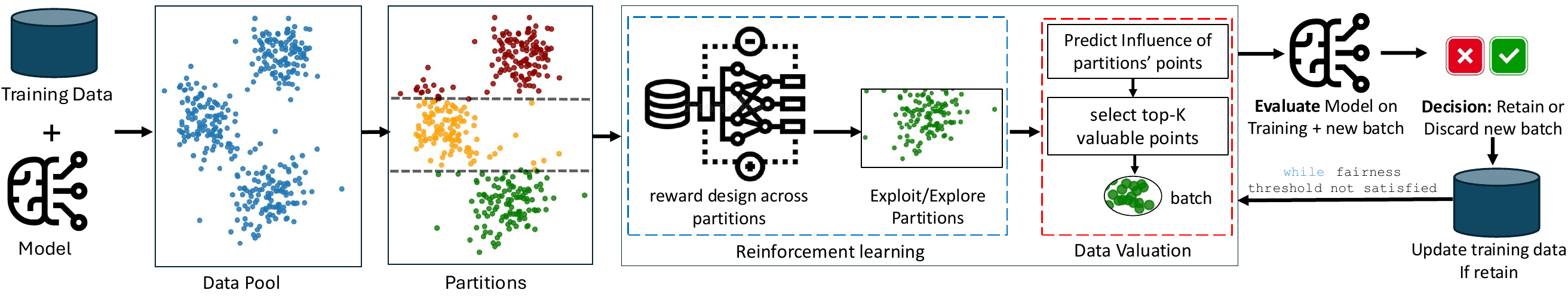}
    \caption{Overview of \sys for selective data expansion aimed at improving model fairness. The blue dashed box highlights \sys with a reinforcement learning–based multi-armed bandit (MAB) framework for selective data expansion, while the red dashed component incorporates data valuation to prioritize informative and fairness-enhancing samples.}
    \label{fig:framework}
\end{figure*}

Disparate representation rates of demographic groups in training data are a primary source of fairness violations~\cite{asudeh_assessing_2019,shahbazi2023representation,shahbazi2024coverage,nargesian2021tailoring}.  
As described in Example~\ref{ex:example}, while ensuring equal representation of demographic groups can partially improve fairness, it often comes at the expense of accuracy.  
This paper, therefore, addresses a central question: which additional data points must be added to the training data to improve fairness while simultaneously preserving the accuracy of the resulting model?

Given a data pool $\datapool$, our goal is to add a subset of points that maximally improves fairness while preserving accuracy. This can be cast as a \textit{subset selection} problem, which is NP-hard in general~\cite{886461319970101}.  
We focus on determining the best subset that has up to $K$ data points (typically, $K \ll |\datapool|$).
However, not all such subsets improve fairness, and identifying the best one requires evaluating $\binom{|\datapool|}{k}$ possibilities ($k \in [1,K]$). 
This exhaustive search is prohibitively expensive due to the exponential number of subsets and the retraining cost per evaluation ($\mathcal{O}(|\datapool|^K \times Tr)$, where $Tr$ is the cost of retraining). 
An alternative solution constructs the subset by sequentially adding up to $K$ \textit{best data points} rather than finding the best subset. The na\"ive approach requires evaluating each data point in the data pool individually and selecting the top-$K$ most valuable data points with an $\mathcal{O}(|\datapool| \times Tr)$ complexity. To reduce the number of times the model is retrained, data are often evaluated in batches rather than one at a time. The challenge then is to construct batches such that we do not unnecessarily evaluate data points that do not improve fairness at all. 


We propose \sys, a framework that integrates data valuation with multi-armed bandits (MAB)~\cite{vermorel2005multi}, a special case of reinforcement learning~\cite{sutton2018reinforcement}, to efficiently evaluate the utility of data points based on their joint impact on model fairness and accuracy when incorporated into training.  
\sys reduces the search space by systematically partitioning the data pool into smaller subsets and determining the order of incorporating into training data through a principled balance between \textbf{exploring} new partitions and \textbf{exploiting} influential data points within the selected partition. Partitioning can be performed automatically (e.g., clustering) or guided by domain knowledge (e.g., stratification by state, demographics, or timeframes). 
Related to our work, AutoData~\cite{chai_selective_2022} proposed an MAB-based technique to selectively acquire data from heterogeneous data sources to enhance the accuracy of a learned model, but has limitations: (a)~its reward formulation is explicitly accuracy-centric and does not account for the inherent trade-off between fairness and utility, making its direct application to fairness-aware data acquisition difficult; (b)~within each partition, it acquires random batches, which limits systematic exploration of beneficial data points; and (c)~
it retrains a model to evaluate each such batch, rendering it inefficient for large datasets and complex models.

To address these limitations, \sys employs an upper confidence bound (UCB)~\cite{auer2002using} strategy guided by a \textit{reward score} that jointly accounts for both \textbf{accuracy} and \textbf{fairness} when selecting a partition for expansion. Within a selected partition, \sys further leverages data valuation to prioritize candidate data points based on their estimated impact on model fairness, and selects the top-$k$ points for acquisition. To ensure computational efficiency, we employ influence functions~\cite{doi:10.1080/00401706.1980.10486199,pmlr-v70-koh17a} to approximate the impact of including individual data points on fairness metrics, thereby avoiding repeated full model retraining for each candidate batch.

\vspace{1mm}\noindent\textbf{Summary of contributions.}
Our main contributions can be summarized as follows:
\begin{itemize}[label=$\bullet$,leftmargin=*]
    \item We formalize the problem of \textbf{selective data expansion for improving the fairness} of an ML model in classification tasks. (Section~\ref{sec:prelim})
    \item We present \sys, a system that casts the task of selective data expansion for model fairness as a \textbf{multi-armed bandit} (MAB) problem and solves it using the upper confidence bound (UCB) algorithm based on a novel reward score that addresses both model fairness and accuracy.  
    (Section~\ref{sec:mab})
    \item To carefully construct a batch for addition, \sys incorporates the concept of \textbf{data valuation} and leverages \textit{influence functions} to estimate the importance of data points toward model fairness, which in turn speeds up \sys. (Section~\ref{sec:valuation})
    \item We conduct extensive experiments on six real-world datasets to demonstrate the effectiveness of \sys in rapidly improving model fairness while preserving accuracy and present trade-offs between effectiveness and efficiency of the proposed methods. (Section~\ref{sec:exp})
\end{itemize}

\section{Preliminaries}
\label{sec:prelim}
This section formally defines the terminology and problems we addressed in this paper.

\vspace{1mm}\noindent\textbf{Classification.}
We consider a binary supervised learning task defined on data domain $\Gamma=\{\mathcal{X}, \mathcal{Y}\}$, where $\mathcal{X}$ denotes the feature space over $p$ features and $\mathcal{Y} = \{0, 1\}$ denotes the binary label space.
%
Suppose there is a conditional distribution $p(y \mid \mb x)$ defined over $\Gamma$, where $\mb x \in \mathcal{X}, y \in \mathcal{Y}$. Given training dataset $\dtrain= \{\datapoint\}_{i=1}^n = \{\mb x_i, y_i\}_{i=1}^n \in \Gamma$, the learning task is to train a classifier $\model$ that represents a distribution $g$ that captures the target distribution $p$ as closely as possible. $\model$ learns a function $f: \mathcal{X} \rightarrow \hat{Y}$ that associates each data point $\mb x$ with a prediction $\hat{y} = f(\mb x) \in \{0,1\}$, and is evaluated on $\dtest \in \Gamma$.
%
Learning algorithm $f$ trains on $\dtrain$ to learn the optimal parameters $\oparam \in \mathbb{R}^p$ that minimize the empirical loss $\mathcal{L} (\dtrain, \param) = \frac{1}{n} \sum_{i=1}^n L(\datapoint, \param)$. 

\vspace{1mm}\noindent\textbf{Algorithmic group fairness.}
Given a binary classifier $\model: \mathcal{X} \rightarrow \hat{Y}$ and a protected attribute $S \in \mathcal{X}$  (such as gender, race, age, etc.), we interpret $\hat{Y}=1$ as a favorable (positive) prediction and $\hat{Y}=0$ as an unfavorable (negative) prediction. We assume the domain of $S$, \textsf{Dom}($S$) $= \{0, 1\}$ where $S = 1$ indicates a privileged and $S=0$ indicates a protected group (e.g.,
males and non-males, respectively). We select the widely popular setting of binary classification with binary protected attributes; however, solutions presented in this paper are easily extensible to settings for multi-class classification and intersectional fairness. 
%
Algorithmic group fairness mandates that individuals belonging to different groups must be treated similarly. The notion of similarity in treatment is captured by different associative notions of fairness such as statistical parity (a.k.a. demographic parity), predictive parity, and equalized odds~\cite{verma2018fairness, mehrabi_survey_2022, chouldechova2017fair}.
%
%
%
For example, for the widely-used group fairness metric of statistical parity, model $\model$ satisfies statistical parity if both the protected and the privileged groups have the same probability of being predicted the positive outcome i.e., $P(\hat{Y} = 1|S = 0) = P(\hat{Y} = 1| S = 1)$.
Although these commonly used metrics are tailored to binary classification, \sys is not inherently tied to binary classification; instead, it relies on a data valuation signal and an acquisition policy that are agnostic to the label structure. 
Extending \sys to multi-class or multi-label classification primarily requires redefining the fairness objective for a multi-label setting, for example, by aggregating class-wise fairness violations or applying one-vs-rest formulations. As such, the framework naturally generalizes to more complex prediction settings beyond binary classification, which is orthogonal to \sys's mechanism.
We define the chosen fairness metric by $\fairness$ and quantify fairness in the predictions over $\dtest$ by a model trained on $\data$ by $\fairness_\data$. 
For example, for statistical parity, $\fairness_\data = P(\hat{Y} = 1|S = 0) - P(\hat{Y} = 1| S = 1)$ quantifies the difference in the probabilities of protected and privileged groups having a positive outcome. If $\fairness_\data < 0$, the model is biased against the protected group while $\fairness_\data > 0$ indicates the model is biased against the privileged group. A higher value of $|\fairness_\data|$ indicates lower fairness (greater disparity) in the model's predictions. 

\vspace{1mm}\noindent\textbf{Data Pool.}
In our framework, data pool $D$ denotes an auxiliary collection of candidate examples that are available for potential acquisition but are not included in the initial training set. 
The pool may be constructed from multiple heterogeneous sources or from a single homogeneous source; in this work, we focus on the latter setting, where $D$ is derived from a single source and thus homogeneous. 
In practice, such pools naturally arise from historical archives, held-out operational data, or newly arriving data in operational pipelines.
Additionally, for tabular learning tasks, $D$ can be formed through externally discovered relational datasets obtained from data lakes or Web APIs that expose tables with partial attribute overlap with the training data~\cite{fernandez2018aurum, nargesian2020organizing, nargesian2018table}. These datasets may require lightweight schema alignment, with missing attributes handled using standard techniques such as NULL imputation. For vision tasks, the pool can be formed from publicly available benchmarks or images retrieved through Web-based search APIs using task-relevant labels. 
The choice and composition of the data pool directly influence the expansion process, as they determine which examples are evaluated and selected. A diverse pool increases the likelihood of identifying points that improve utility or mitigate fairness gaps, while a skewed or limited pool may restrict attainable gains. Importantly, our method does not assume that the pool follows the same distribution as the training data and remains effective under distributional shifts, as demonstrated in Section~\ref{subsec:hyperparameter}. The acquisition strategy relies on the observed contribution of candidate subsets rather than global distributional alignment. Finally, we assume a uniform cost for incorporating any point from the pool, consistent with scenarios where the pool has already been acquired.
\vspace{1mm}\noindent\textbf{Problem Definition.} Given a model 
trained on $\dtrain$, fairness metric $\fairness$, and data pool $\datapool$, 
we address the problem of determining additional data points $\acq \subset \datapool$ that must be added to $\dtrain$ such that 
the model learned on $\dtrain \cup \acq$ is 
fairer 
than the original model learned on $\dtrain$ alone (i.e., $|\fairness_{\dtrain \cup \acq}| < |\fairness_{\dtrain}|$).

\section{Data Expansion Framework}
\label{sec:framework}

In this section, we introduce \sys, a framework for acquiring data points that enhance the fairness of a learned model. Section~\ref{sec:mab_acq} formulates the task as a multi-armed bandit (MAB) problem, which identifies the partition with the highest potential for fairness improvement but not the specific points to select. To address this, Section~\ref{sec:valuation} introduces a data valuation-based batch selection strategy for targeted acquisition. By integrating these components, \sys jointly improves fairness and overall model performance.


\subsection{\textbf{Multi-Armed Bandit (MAB) framework}} 
\label{sec:mab}
The Multi-Armed Bandit (MAB) ~\cite{vermorel2005multi} maps a framework for sequential decision-making under uncertainty. This problem can be framed by the metaphor of a gambler (or `Agent') choosing which of several slot machines (or `Arms') to play in each attempt to maximize the total prize over a series of trials. Considering that gamblers have some knowledge about each slot machine from initial trials, they are faced with the question of which machine to select next. The multi-armed bandit framework is ideal for solving this dilemma. MAB can formally be defined as: at each state $t$, the agent selects an arm $i$ from a set of $K$ available arms and receives a reward $r_t$ from a distribution associated with that arm, which is generally unknown to the agent. MAB aims to determine which arm to pull next to maximize the cumulative reward $R$ after $T$ rounds, guided by the principle of balancing \textit{exploration} and \textit{exploitation}. 

\textit{Exploration} involves choosing unexplored options to gather new information about their potential rewards. By exploring, the agent reduces uncertainty about less-known arms, potentially uncovering actions that provide higher rewards than initially expected. On the other hand, \textit{exploitation} involves selecting the action that currently offers the highest reward based on the agent's existing knowledge. While exploitation maximizes short-term gains by considering known information, it may lead to suboptimal 
long-term outcomes if the agent overlooks better options that have not been sufficiently explored. At the same time, excessive exploration may waste resources on testing suboptimal actions, thus missing opportunities for immediate reward maximization.

Thus, the agent must balance between exploring and exploiting to maximize cumulative rewards eventually. To find a trade-off between exploration and exploitation, making optimal short-term decisions based on available information is crucial to solving this dilemma. This trade-off is central to various algorithms, such as Thompson sampling~\cite{agrawal2012analysis}, $\epsilon$-greedy\cite{kuleshov2014algorithms}, and Upper Confidence Bound (UCB)~\cite{auer2002using} designed to 
maximize the agent's long-term performance. Due to its computational flexibility, we adopt the UCB algorithm for our problem, while noting that alternative algorithms could also be applied to \sys. Similar to other base algorithms for MAB, the UCB algorithm does not always yield exact optimal results, but its performance is near-optimal~\cite{bubeck2013bandits}.

Next, we will map the problem of data expansion for improving model fairness to the MAB framework and discuss our approach.

\subsection{\hspace{-2mm}\textbf{Mapping data expansion for fairness to Multi-Armed Bandit (MAB)}}
\label{sec:mab_acq}
To cast the data expansion problem to the MAB framework, we first split the data pool into several disjoint partitions, i.e., $\datapool=\bigcup_{i=1}^{g} C_i$ where $g$ is the number of partitions. 
These partitions could be obtained by clustering $\datapool$ using existing clustering algorithms such as multivariate Gaussian Mixture Model (GMM)~\cite{figueiredo2002unsupervised}, $k$-means~\cite{jain2010data}, hierarchical~\cite{murtagh2012algorithms}, and DBSCAN~\cite{ester1996density}  or partitioning the data pool based on some criterion or by considering the data pool as a collection of data sources. Partitioning the data pool offers two advantages: i) ensures systematic exploration of overlooked data composition critical for fairness, ii) and it reduces computational complexity by restricting selection to smaller, structured subsets rather than the entire data pool.
However, each partition \( C_i \) is then treated as an \textit{arm} in the multi-armed bandit (MAB) framework. In the \( k \)-th iteration, the algorithm selects a partition \( C_k \) and samples a batch from \( C_k \) for evaluation. We will discuss the batch selection process in more detail in Section~\ref{sec:valuation}. 
Subsequently, the selected batch is merged with the current training dataset, and the fairness of the resultant model trained on the updated training dataset is reported. The change in the model fairness before and after adding the batch determines whether the batch should be retained or discarded and whether the selected partition is rewarded or penalized.
The reward/penalty score is utilized to select the subsequent partition to evaluate more data points.
The blue-box part in Figure \ref{fig:framework} illustrates the data expansion task, framed as a multi-armed bandit problem.

\vspace{2mm}\noindent\textbf{Reward/Penalty score.} 
The algorithm scores the chosen partition according to the change in model fairness after merging the batch with existing training data. The reward for partition $C_i$ at iteration $k$ is defined as $r_i^k$. 
If fairness improves, the partition is rewarded, otherwise penalized. 
However, in addition to allocating reward/penalty scores to the chosen partition, the \textit{full feedback}~\cite{slivkins2019introduction} MAB algorithm also assigns a portion of scores to other partitions that could have been selected. This approach efficiently accelerates the process by reducing the number of evaluations. Several studies on MAB-based data expansion for improving model performance~\cite{chai_selective_2022, wang2024optimizing} consider the distance between partitions when assigning rewards or penalties to other partitions. The intuition behind this scoring is based on the assumption that partitions that are closer to the selected partition have a higher likelihood of getting selected and, hence, should be rewarded similarly.
This assumption holds for performance metrics such as accuracy and confidence because closer partition centroids indicate that the partition shares similar characteristics. However, this setting might not hold for model fairness because partitions that are close might have extremely different compositions over sensitive attributes and, therefore, might impact fairness differently.
In other words, the \textit{base rate difference} of the partition plays a significant role in fairness. Recall that the base rate difference for dataset $D$ is defined as: 
$\Delta BR_{D} = P(Y=1 \mid S=0) - P(Y=1 \mid S=1)$.
In the context of fairness, the selection of a partition indicates an improvement in model fairness as a result of inherent lower base rate difference among the different demographic groups in the partition. 
Consequently, other partitions with similar base rates should be rewarded higher than those with worse base rates. 
In fact, we show that
incorporating intra-partition base rate differences among the different demographic groups is much more effective in the proper distribution of the reward among partitions, 
resulting in significantly improved fairness (see Section~\ref{subsec:hyperparameter} for more details). 

Focusing solely on the base rate difference, however, can degrade the overall accuracy of the learned model. We, therefore, propose a \textit{novel reward score} that caters to both fairness and accuracy with a balance parameter to split the reward scores among the different partitions. When partition $C_i$ is evaluated, the reward score for each partition $C_j$ is computed as follows:

\begin{equation}
    r_j = \frac{\Delta \fairness}{1 + |\Delta BR_{C_j}|}+\beta\frac{\Delta acc}{1+dist(C_i, C_j)} 
    \label{eq:reward_eq}
\end{equation}
where $\Delta \fairness = \fairness_{\dtrain^{k-1} \cup b_i} - \fairness_{\dtrain^{k-1}}$ denotes the change in model fairness after adding batch $b_i$ to training data up to $k-1$ iterations denoted by $\dtrain^{k-1}$, while $\Delta acc$ stands for accuracy change. The parameter $\beta$ is used to balance the contributions of fairness and accuracy in the reward scores. 
Additionally, $\Delta BR_{C_j}$ indicates the intra-partition base rate difference in $C_j$ and $dist(C_i, C_j)$ is the normalized Euclidean distance between the two partitions computed over the partition centroids.  A positive value in $r_j$ incurs a reward, while a negative value incurs a penalty.

\vspace{1mm}\noindent\textbf{Aggregated reward/penalty score.} The multi-armed bandit approach aggregates the prior reward and penalty information up to $k$ iterations to make an informed decision in the next iteration. Let $R_i^k$ represent the aggregate score of partition $C_i$ from iteration $1$ to $k$, defined as:
$R_i^k=\frac{1}{n_i^k} \sum_{j=1}^k r_i^j$,
where $r_i^j$ represents the reward score of $C_i$ at the $j$-th iteration, and $n_i^k$ denotes the number of times $C_i$ is selected and rewarded a positive score up to the $k$-th iteration. 

\vspace{1mm}\noindent\textbf{Upper Confidence Bound~(UCB).} The UCB algorithm is designed to adaptively adjust the trade-off between exploration and exploitation over time~\cite{auer2002using, auer2000using}, which is achieved by incorporating a measure of uncertainty or confidence in the estimated rewards of each arm. This measure is used to guide the decision-making process, allowing the algorithm to explore arms with potentially high but uncertain rewards while also exploiting arms with known high rewards. Due to its deterministic nature, it has been widely used in the field of MAB, Reinforcement Learning, and Recommendation Systems~\cite{carpentier2011upper, nguyen2019recommendation, auer2000using, garivier2011upper}. We determine the UCB score for partition $i$ in the $k$-th iteration as:

\begin{equation}
U_i^k=R_i^k+\alpha \sqrt{2 \ln \left( \frac{n^k}{n_i^k+1}\right)}
\label{eq:UCB_eq}
\end{equation}

where $\alpha$ is a pre-defined parameter that maintains the balance between exploration and exploitation, and $n^k =\sum_{i=1}^g n_i^k$ for the total number of partitions $g$.
The first term in Equation~\ref{eq:UCB_eq} represents exploitation, while the latter pertains to exploration. A partition with a higher reward will have a higher exploitation score, whereas one selected less frequently will have a higher exploration score.

\subsubsection{Role of Batch Selection.} MAB plays a central role in identifying the partition with the highest expected reward. Once a partition is selected, however, sampling a batch from it is non-trivial: while the partition may be promising overall, not every randomly chosen subset will improve model performance. AutoData~\cite{chai_selective_2022} adopts a random batch selection strategy, but over time, this approach performs no better than naive random acquisition from the entire pool—even when combined with MAB (see results in Figure~\ref{fig: Datasift Vs Others}).

To facilitate more effective batches, the following section introduces a data valuation approach for enhanced batch selection.

\subsection{Valuation-based Batch Selection}
\label{sec:valuation}
To carefully construct a batch to merge from a chosen partition, we leverage the idea of data valuation, which has been successfully used in explainable AI~\cite{pmlr-v97-ghorbani19c,pmlr-v70-koh17a} to quantify the contribution of training data points toward the performance of the learned model. 
Due to their effectiveness in accurately estimating the contribution of data points and faster online computation time, we use first-order \textit{influence functions}~\cite{pmlr-v70-koh17a,doi:10.1080/00401706.1980.10486199} to approximate the importance of training data points toward model fairness. Based on the computations, we construct the batch to include the chosen partition.

\subsubsection{Influence functions}
\label{sec:inf}
Recall from Section \ref{sec:prelim} that 
$\theta^*$ is the 
set of optimal parameters that 
minimize the empirical risk, i.e.,

\begin{equation}
\theta^*=\underset{\theta \in \Theta}{\operatorname{argmin}} \hspace{1mm}\mathcal{L}(\theta)=\underset{\theta \in \Theta}{\operatorname{argmin}} \hspace{1mm} \frac{1}{n} \sum_{i=1}^n L\left({d}_i, \theta\right) \label{eq:Loss}
\end{equation}

To incorporate influence functions, we assume that the empirical risk function $\mathcal{L}(\theta)$ is twice-differentiable and strictly convex. Under these conditions, we can guarantee the Hessian matrix $\mathcal{H}_\theta$ exists and is positive definite, and therefore, its inverse $\mathcal{H}_\theta^{-1}$ also exists. 
These assumptions are applicable to a wide range of classification algorithms such as logistic regression, support vector machines, and feed-forward neural networks.

Let $\nabla_\theta \mathcal{L}(\theta)$ and $\mathcal{H}_\theta=\nabla_\theta^2 \mathcal{L}(\theta)=\frac{1}{n} \sum_{i=1}^n \nabla_\theta^2 L\left({d}_i, \theta\right)$ be the gradient and the Hessian of the loss function, respectively. The influence of up-weighting a data point $\mathbf{d} \in \dtrain$ by $\epsilon$ on the model parameters is computed as:

\begin{equation}
I_\theta(\mathbf{d})=\left.\frac{d \theta_\epsilon^*}{d \epsilon}\right|_{\epsilon=0}=-\nabla_\theta^2 \mathcal{L}\left(\theta^*\right)^{-1} \nabla_\theta L\left(\mathbf{d}, \theta^*\right)=-\mathcal{H}_\theta^{-1} \nabla_\theta L\left(\mathbf{d}, \theta^*\right) \label{eq:par_influence}   
\end{equation}

To add a data point $\mathbf{d}$ to training data, we up-weight it by $\epsilon = \frac{1}{n}$. Therefore, the influence of $\mathbf{d}$ on model parameters can be linearly approximated by computing $d \theta_\epsilon^* \approx \frac{1}{n} I_\theta(\mathbf{d})$.

Using the chain rule of differentiation, we can estimate the effect of up-weighting data point $\mathbf{d}$ on any function $f(\theta)$ as:

\begin{equation}
\mathcal{I}_f(\mathrm{d})=\left.\frac{d f\left(\theta_\epsilon^*\right)}{d \epsilon}\right|_{\epsilon=0}=\left.\frac{d f\left(\theta_\epsilon^*\right)}{d \theta} \frac{d\left(\theta_\epsilon^*\right)}{d \epsilon}\right|_{\epsilon=0}=\nabla_\theta f\left(\theta^*\right)^{\top} \mathcal{I}_\theta(\mathrm{d}) \label{eq:fair_inf}
\end{equation}

When $f = \mathcal{F}$, we obtain the first-order influence of a single training data point $\mathbf{d}$ on model fairness by approximating $d \fairness_\epsilon^* \approx \frac{1}{n} \mathcal{I}_\fairness(\mathrm{d})$. 

\subsubsection{Batch construction}
\label{sec:batch_construction}
Given a data point $\mathbf{d} \in \dtrain$, the influence function approximation in Equation~\ref{eq:fair_inf} estimates the effect of up-weighting $\mathbf{d}$ on model fairness.
To generalize this notion to the unlabeled data pool $\datapool$, we first compute first-order influence estimates on $\dtrain$ and train a polynomial Ridge regression model to approximate the fairness influence as a function of data features. The regressor is trained using features from $D_{\text{train}}$, with targets given by the estimated fairness influence obtained via influence functions. The learned regression model $\mathcal{R}$ is then applied to unseen data points in each partition $C_i \subset \datapool$ to predict their expected fairness influence on the current model $\mathcal{M}$. Data points within each partition are subsequently ranked in descending order of predicted influence, as formalized in \hyperref[fn:sortpartitions]{\FuncName{SortPartitions()}}.
At each iteration, a batch $b$ is constructed by selecting the top-$K$ data points with the highest predicted influence on fairness. 
\vspace{0.8em} 
\SetKwFunction{SortPartitions}{\FuncName{SortPartitions}}
\SetKwProg{Fn}{Function}{:}{}

\phantomsection\label{fn:sortpartitions}
\Fn{\SortPartitions{$C=\{C_1,\dots,C_g\}, \dtrain, \mathcal{M}$}}{
  $I \gets \textsc{get\_influence}(\dtrain,\mathcal{M})$ as in Eq.~\ref{eq:fair_inf}\;\\
  $\dtrain \gets \{(x, y, I(x)) : (x,y) \in \dtrain\}$\;\\
  $\mathcal{R} \gets \text{TrainRegressor}(\dtrain)$\;\\
  \For{$i \gets 1$ \KwTo $g$}{
    \For{$x \in C_i$}{
      $\hat{I}(x) \gets \mathcal{R}(x)$\;
    }
    $Sorted\_C[i] \gets \text{Sort}\!\left(C_i;\, \hat{I}(x)\ \text{desc.}\right)$\;
  }
  \KwRet{$Sorted\_C$}\;
}
\vspace{2mm}

Note that a batch $b\subseteq \datapool$ constructed in the above manner does not explicitly account for the inherent correlations between data points in $b$. As such, the estimated influence of the batch computed simply by adding the first-order influences of individual data points might not be exact, i.e., $\fairness_{\datapool \cup b}$ is not necessarily equal to $\fairness_{\datapool} + \frac{|b|}{n}\sum_{\mathbf{d} \in b} \mathcal{I}_\fairness(\mathrm{d})$. Therefore, the selected batch is not guaranteed to be optimal.
Identifying the optimal batch within a selected partition can be formulated as a \textit{subset selection} problem, which is NP-hard~\cite{886461319970101}. In practice, however, employing influence function approximations as a heuristic for batch construction has been shown to yield near-optimal performance (see Section~\ref{sec:valuation_exp}).



\subsection{DataSift Framework}
\label{sec:datasift}
In this section, we present \sys, a data expansion framework designed to enhance model performance. Figure~\ref{fig:framework} provides an overview. At a high level, given a biased model trained on an initial dataset and an available data pool, \sys first systematically partitions the pool into smaller subsets to reduce the search space. It then employs a multi-armed bandit (MAB) strategy to identify the partition with the highest potential reward (blue-dashed box in Figure~\ref{fig:framework}). In each round, \sys applies data valuation to extract the top-$K$ influential data points to form a batch from the selected partition (red-dashed part in Figure~\ref{fig:framework}). The model is retrained with the newly acquired batch and re-evaluated. Based on the observed performance, \sys decides whether to retain the batch—updating the training data—or discard it. 
This process iterates until the stopping criteria are met, ultimately yielding a fair model trained on the updated data.

\vspace{1mm}\noindent\textbf{Early stopping criteria.} The Algorithm~\ref{alg:MAB_algorithm} is designed to acquire data points from the available pool in order to maximize improvements in model fairness. In practice, however, we observe that the model often achieves near-fairness (i.e., a parity difference close to zero) after incorporating only a fraction of the pool. Continuing to evaluate and add further data in such cases leads to unnecessary computational overhead. Moreover, if the entire pool were added, all methods would eventually converge to the same performance, obscuring the benefit of selective expansion. To address this, we define an expansion budget, $\budget$, representing a fraction of the data pool, and introduce a fairness threshold, $\tau$, which serves as an early stopping criterion in Algorithm~\ref{alg:MAB_algorithm}.
The expansion process is halted and the data points added so far are returned if: (a)~model fairness is within threshold $\tau$, or (b)~expansion budget \budget~is exhausted.
\begin{algorithm}[t]
\caption{\sys Framework}
\label{alg:MAB_algorithm}
\KwIn{$C=\{C_1,\dots,C_g\}, \dtrain, \dtest, \mathcal{M}, \tau, K, \budget$}
\KwOut{$\acq$}

\nl $Sorted\_C \gets \SortPartitions(C, \dtrain, \mathcal{M})$\;

\nl \For{$i \gets 1$ \KwTo $g$}{ $R_i,n_i,U_i \gets 0$ }

\nl $\acq \gets [\,]$\; $k \gets 0$\; $\dtrain^0 \gets \dtrain$\; $\fairness\gets \fairness(\dtrain)$\;

\nl \While{$|\fairness| \ge \tau$ \textbf{and} $\budget>0$}{

  \nl $k\gets k+1$;\quad \\
  \nl $i \gets \arg\max_j U_j^k$;\quad\\ 
  \nl sample $b_i \subset Sorted\_C[i]$, $|b_i|=K$\;

  \nl $\Delta f \gets \fairness(\dtrain^{k-1}\cup b_i) - \fairness(\dtrain^{k-1})$\;

  \nl \If{$|\Delta f|>0$ \textbf{and} $|\fairness(\dtrain^{k-1}\cup b_i)| \le |\fairness|$}{
    \nl $\acq \gets \acq \cup b_i$\;
    \nl $\dtrain^k \gets \dtrain^{k-1} \cup b_i$\;
    \nl $Sorted\_C[i] \gets Sorted\_C[i] \setminus b_i$\;
    \nl $\fairness \gets \fairness(\dtrain^k)$\;
    \nl $\budget \gets \budget - K$\;
  }

  \nl \For{$j \gets 1$ \KwTo $g$}{
    \nl $r_j \gets \dfrac{\Delta f}{1+|\Delta BR_{C_j}|} + \beta\dfrac{\Delta acc}{1+dist(C_i,C_j)}$\;
    \nl \textbf{update} $R_j^k, n_j^k, U_j^k$\;
  }
}
\nl \Return{$\acq$}
\end{algorithm}

\textbf{Algorithm~\ref{alg:MAB_algorithm}} provides the pseudocode of \sys, which is designed to enhance fairness without degrading accuracy. The algorithm takes as input the set of partitions, training and test data, the base model, the fairness threshold $\tau$, the batch size $K$, and the expansion budget $B$, and outputs the acquired subset of data points.  
\textbf{Lines 1--3:} The procedure begins with the initialization of key variables and parameters.  
\textbf{Lines 4--17:} The main loop iterates until either the specified fairness threshold is reached or the expansion budget is exhausted.  
\textbf{Lines 6--8:} In each iteration, the partition with the maximum UCB score is selected, from which a batch of size $K$ with the highest predicted influence values is sampled and subsequently evaluated.  
\textbf{Lines 9--14:} The sampled batch is incorporated into the training data only if it leads to an improvement in fairness relative to all previous iterations, after which the training set and the corresponding partition are updated.  
\textbf{Lines 15--17:} The algorithm then updates the reward and penalty values for each partition, along with the aggregated and UCB scores.  
Finally, the updated training set is returned.

Algorithm~\ref{alg:MAB_algorithm} has a computational complexity of 
$\mathcal{O}(I \times (|C| + E + t))$
where \( I \) denotes the number of required evaluations (iterations), $|C|$ is the maximum partition size,
\( E \) is the evaluation complexity for each batch, and \( t \) refers to the complexity of score calculation. The maximum number of evaluations, \( I \), is bounded by \( I = A + R \leq \frac{N}{K} \), where \( A \) and \( R \) represent the number of accepted and rejected batches, respectively, \( N \) is the size of the data pool, and \( K \) is the batch size. Moreover, 
\( A \leq \frac{B}{K} \)
where \( B \) is the expansion budget .

Note that the classical formulation of influence functions to compute the valuation of data points limits the applicability of the approach to parametric models with convex and twice-differentiable loss functions. While \sys is instantiated using parametric models with convex, differentiable losses to enable efficient influence estimation, this assumption is not fundamental to the framework. The key requirement is the  relative ranking of data utility, which is all \sys requires for acquisition decision. In practice, influence-based estimates can be extended beyond this setting using standard approximations such as local second-order smoothing, damped Hessian inverses, or surrogate losses, which preserve relative data utility rankings~\cite{pmlr-v70-koh17a}. Besides, prior work on data valuation has shown that Data Shapley approximations~\cite{pmlr-v97-ghorbani19c} can provide model-agnostic utility estimation, albeit at a higher computational cost. Thus, the parametric setting adopted in this work represents a practical and efficient instantiation of \sys rather than a strict limitation.
We also evaluate a model-agnostic variant of \sys in Section~\ref{sec:exp}, denoted as \sysmab, which omits data valuation during expansion and instead relies on random batch sampling.

Table~\ref{tab:running_eaxample} presents an example illustrating the first five iterations out of 25 for the decision-making process of \sys on a small data pool (\(|\datapool|=50\)) from the ACSIncome dataset~\cite{ding2021retiring}. The pool was partitioned into an optimal number of clusters, and, with a mini-batch size of two instances, the table reports the evaluation process. It highlights how \sys selects clusters and batches while balancing exploration and exploitation. Overall, in this small example, \sys reduced model unfairness by 60\%.


\section{Experimental Evaluation}
\label{sec:exp}


\begin{table*}[t]
\caption{Illustrative example of cluster and batch selection decisions of \sys on ACSIncome with $|\datapool|=50$ and batch size $2$. $\Delta\fairness$ is the round-wise fairness change after adding a batch; $U$ is the UCB score. \textcolor{darkgreen}{Dark green} indicates improved fairness after adding the selected batch; \textcolor{red}{red} indicates a decline. The bold $U$ value indicates the cluster selected next.}
\label{tab:running_eaxample}

\renewcommand{\arraystretch}{1}
\setlength{\tabcolsep}{10pt}

\resizebox{\textwidth}{!}{%
\fontsize{12pt}{14pt}\selectfont
\begin{tabular}{@{}c r r r r r r r r c@{}}
\toprule
& \multicolumn{2}{c}{\textbf{$C_1$}}
& \multicolumn{2}{c}{\textbf{$C_2$}}
& \multicolumn{2}{c}{\textbf{$C_3$}}
& \multicolumn{2}{c}{\textbf{$C_4$}}
& \\

\cmidrule(lr){2-3}
\cmidrule(lr){4-5}
\cmidrule(lr){6-7}
\cmidrule(lr){8-9}

\textbf{Evaluation}
& $\Delta\fairness$ & $U_1$
& $\Delta\fairness$ & $U_2$
& $\Delta\fairness$ & $U_3$
& $\Delta\fairness$ & $U_4$
& \textbf{Decision} \\

\midrule

1
& \textcolor{red}{-0.0118 $\downarrow$} & -0.0118
& & -0.0073
& & -0.0068
& & \textbf{-0.0050}
& \textcolor{red}{$\Delta\fairness \downarrow$}--Discard batch, Next Cluster $\rightarrow C_4$ \\

2
& & 0.0393
& & 0.0382
& & 0.0373
& \textcolor{darkgreen}{0.0366 $\uparrow$} & \textbf{0.0991}
& \textcolor{darkgreen}{$\Delta\fairness \uparrow$}--Retain batch, Next Cluster $\rightarrow C_4$ \\

3
& & 0.0595
& & 0.0537
& & 0.0510
& \textcolor{darkgreen}{0.0043 $\uparrow$} & \textbf{0.0976}
& \textcolor{darkgreen}{$\Delta\fairness \uparrow$}--Retain batch, Next Cluster $\rightarrow C_4$ \\

4
& & \textbf{0.0800}
& & 0.0712
& & 0.0673
& \textcolor{darkgreen}{0.0110 $\uparrow$} & 0.0800
& \textcolor{darkgreen}{$\Delta\fairness \uparrow$}--Retain batch, Next Cluster $\rightarrow C_1$ \\

5
& \textcolor{darkgreen}{0.0050 $\uparrow$} & 0.1257
& & \textbf{0.1871}
& & 0.1819
& & 0.0999
& \textcolor{darkgreen}{$\Delta\fairness \uparrow$}--Retain batch, Next Cluster $\rightarrow C_2$ \\

\bottomrule
\end{tabular}%
}
\end{table*}

This section presents experiments that evaluate the effectiveness of \sys. 
We seek to answer the following research questions: \textbf{RQ1:} How effective is \sys compared to existing methods in selecting additional data points to improve model fairness across different machine learning models? 
\textbf{RQ2:} What is the benefit of incorporating MAB and data valuation in \sys? \textbf{RQ3:} How effective is \sys with respect to the different hyperparameters and design choices? \textbf{RQ4:} How efficient and scalable are \sys with respect to varying dataset sizes?


\subsection{Experimental Setup}
\label{sec:exp:setup}
\subsubsection{Datasets.}\label{sec:dataset_description} We consider several real-world datasets popular in the fair machine learning literature. \textbf{AdultIncome~\cite{dua2017uci}} dataset used to predict whether an individual's annual income exceeds $\$50,000$ by analyzing a range of demographic and socio-economic factors, including several sensitive attributes, such as age, sex, and race. \textbf{Credit~\cite{GiveMeSomeCredit}} dataset used to forecast the likelihood of delaying credit payments using financial and demographic history. 
American Community Survey (ACS)-based datasets~\cite{ding2021retiring} that provide a suite of datasets (including \textbf{ACSIncome, ACSPublicHealth, ACSMobility}, and \textbf{ACSEmployment}) for predicting different outcomes such as income level, public health status, mobility information, and employment status. 
The ACS datasets are highly skewed toward California (CA) state. Unless otherwise specified, our analysis primarily focuses on evaluating CA data for the year 2018. 
Table~\ref{tab:datasets_summary} summarizes the datasets, all of which exhibit demographic imbalance. Such imbalance is common in real-world applications and creates inherent challenges for fair model training, making them well-suited for evaluating our proposed approach.

\begin{table*}[t] 
\centering
\caption{Summary of datasets.}
\renewcommand{\arraystretch}{1.4} 
\setlength{\tabcolsep}{10pt}
\resizebox{\textwidth}{!}{
\fontsize{12pt}{14pt}\selectfont
\begin{tabular}{lccccccc}
\toprule
\rowcolor[HTML]{E8E8E8} 
\textbf{Dataset} &  \textbf{Adult Income} & \textbf{Credit} & \textbf{ACSPublicHealth} & \textbf{ACSMobility} & \textbf{ACSEmployment} & \textbf{ACSIncome} \\
\midrule
\rowcolor[HTML]{F5F5F5} 
Size & $45,222$ & $150,000$ & $138,554$ & $80,329$ & $378,817$ & $250,847$ \\
No. of features&  8 & 10 & 19 & 22 & 16 & 11 \\
Protected Group ($\mathbf{S}=0$) & Female & Age < 35 & Female & African-American & African-American & Female \\
Population of ($\mathbf{S}=0$) (\%) & 33 & 12.8 & 55.9 & 5.2 & 4.8 & 47.2 \\
\rowcolor[HTML]{F5F5F5} 
Positive Labels in ($\mathbf{S}=0$) (\%) & 11 & 11 & 35 & 73 & 39 & 34 \\
Predictive Task & income> \$50K? & serious delay in 2 years? & has public insurance coverage? & moved address last year? & employed? & income> \$50K? \\
\bottomrule
\end{tabular}
}
\label{tab:datasets_summary} 
\end{table*}

\subsubsection{Competing methods} \label{sec: datasift_types} We compared our proposed approach with several algorithms suited for data expansion: 

\noindent\textbf{\textsf{Random}.} This na\"ive baseline method randomly selects a batch from the data pool in each iteration.

\noindent\textbf{\textsf{Entropy}.} This method selects the $B$ data points with the highest predictive uncertainty~\cite{6773024}. Entropy of data point $d_i$ is calculated as:
$H(d_i)= -\left(p_i \log_2(p_i)+(1-p_i) \log_2(1-p_i)\right)
$
where $p_i$ is the probability that $d_i$ is predicted by the model to have a positive outcome. A higher entropy value indicates that the model is less confident about its prediction for the data point. In each round, it selects the top-$K$ data points with the highest entropy values.


\noindent\textbf{\textsf{AutoData}~\cite{chai_selective_2022}.} This method uses the MAB framework for data acquisition to improve model accuracy. For comparison, we transform the constraint of this algorithm from accuracy to fairness.

\noindent\textbf{\textsf{SliceTuner}~\cite{10.1145/3448016.3452792}.} This method fits per-slice loss–data curves and solves a budget-constrained convex program that balances accuracy and fairness by penalizing slices with above-average losses. For comparison, we consider observing the fairness metric and providing slices into four groups: the cross product of the binary sensitive attribute and the binary target outcome.

\noindent\textbf{\sys.} This method represents our proposed solution described in Algorithm~\ref{alg:MAB_algorithm} that integrates the multi-armed bandit approach with data valuation.

\noindent\textbf{\sysmab.} This method is a variant of \sys that relies solely on the multi-armed bandit(MAB) approach, where batches are selected at random instead of data valuation.

\noindent\textbf{\sysinf}. This method focuses solely on data valuation--computes the influence of data points in the training data using Equation~\ref{eq:fair_inf} and learns a regression model to estimate the influence of points in the data pool on model fairness. Each iteration selects the top-$K$ data points with the calculated highest influence.

\subsubsection{Fairness metrics} There are several metrics used to assess the fairness of a trained model, including \textit{statistical parity}, \textit{true positive rate}, and \textit{predictive parity}~\cite{verma2018fairness, mehrabi_survey_2022, chouldechova2017fair}. Our algorithm effectively addresses any of these fairness metrics. 
Unless stated otherwise, all experiments are conducted using statistical parity, with some results reported under true positive rate parity and predictive parity.

\subsubsection{Settings}
\noindent
We divided the entire dataset into three parts: $\dtrain$ (Training), $\dtest$ (Test),  $\dval$ (Validation) and $\datapool$ (Data pool) in the ratio of $1:2:2:10$. All bandit decisions, fairness checks, and stopping criteria are now computed on $D_{\text{val}}$, while $D_{\text{test}}$ is used only once at the very end to report accuracy and fairness. While test and data pool were split randomly, the training dataset was sampled to obtain an initial biased model. We partition the data pool using Gaussian Mixture Model (GMM)~\cite{figueiredo2002unsupervised} and find the optimal number of partitions ranges between two and seven using the Bayesian Information Criterion (BIC)~\cite{neath2012bayesian}. 
Note that \sys is similarly effective to the other choice of clustering algorithm. We demonstrate its performance with multiple methods, including k-means~\cite{jain2010data} and DBSCAN~\cite{ester1996density}, Hierarchical-BIRCH~\cite{zhang1997birch}.
The model was trained over $\dtrain$ and evaluated on $\dtest$. The expansion budget \(B\) was set to \(20\%\) of the data pool, $|\budget|=.2*|\datapool|$ and the batch size, $|K| = 10\%$ of the $\budget$. The target fairness threshold $\tau$ was set at $0.01$.
The trade-off parameter between exploration and exploitation (\(\alpha\) in Equation~\ref{eq:UCB_eq}) was set to \(0.1\) as~\cite{chai_selective_2022}. The balance parameter \(\beta\)~(in equation~\ref{eq:reward_eq}) was heuristically fixed at \(1\). We considered three ML algorithms: logistic regression, a support vector machine, a feed-forward neural network with one layer and ten nodes.
We used the PyTorch \cite{paszke2019pytorch} or sklearn \cite{pedregosa2011scikit} implementation of these algorithms.


\begin{figure}[b]
    \centering

    \begin{subfigure}[b]{0.48\columnwidth}
        \centering
        \includegraphics[width=\linewidth]{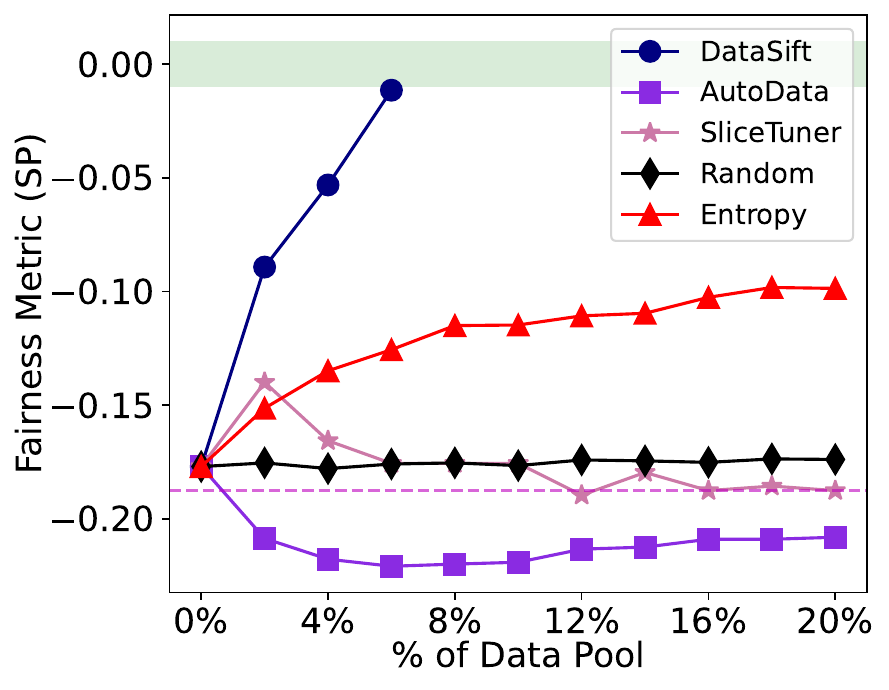}
        \caption{AdultIncome}
        \label{fig: AdultIncome_DataSift}
    \end{subfigure}\hfill
    \begin{subfigure}[b]{0.47\columnwidth}
        \centering
        \includegraphics[width=\linewidth]{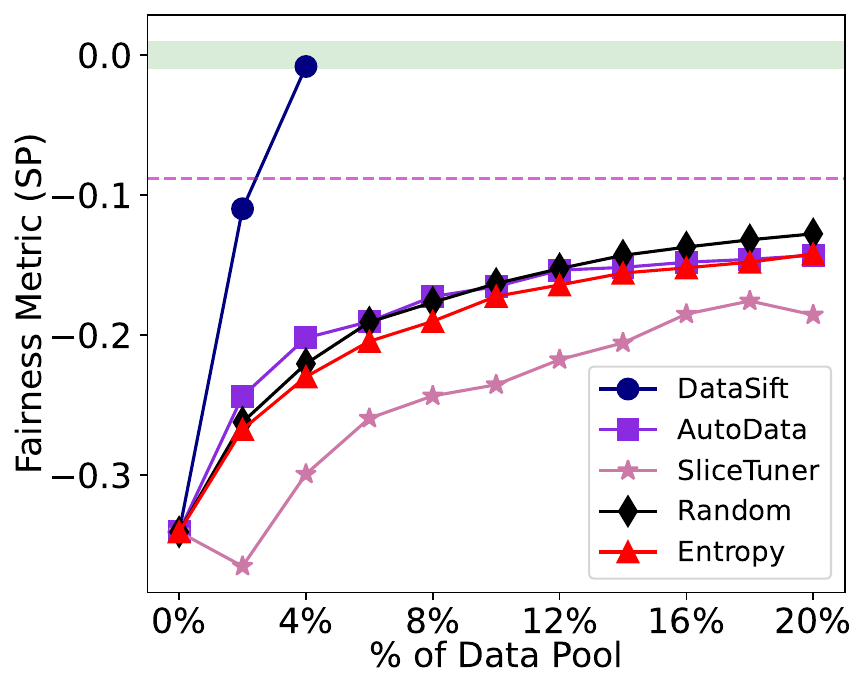}
        \caption{ACSIncome}
        \label{fig: ACSIncome_DataSift}
    \end{subfigure}


    \begin{subfigure}[b]{0.48\columnwidth}
        \centering
        \includegraphics[width=\linewidth]{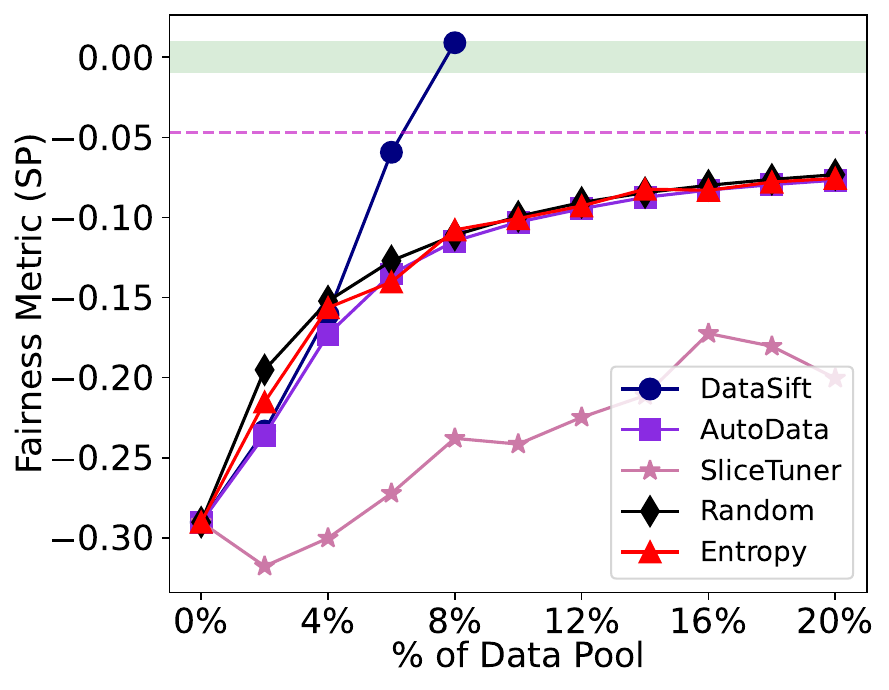}
        \caption{ACSEmployment}
        \label{fig: ACSEmployment_DataSift}
    \end{subfigure}\hfill
    \begin{subfigure}[b]{0.48\columnwidth}
        \centering
Mu        \includegraphics[width=\linewidth]{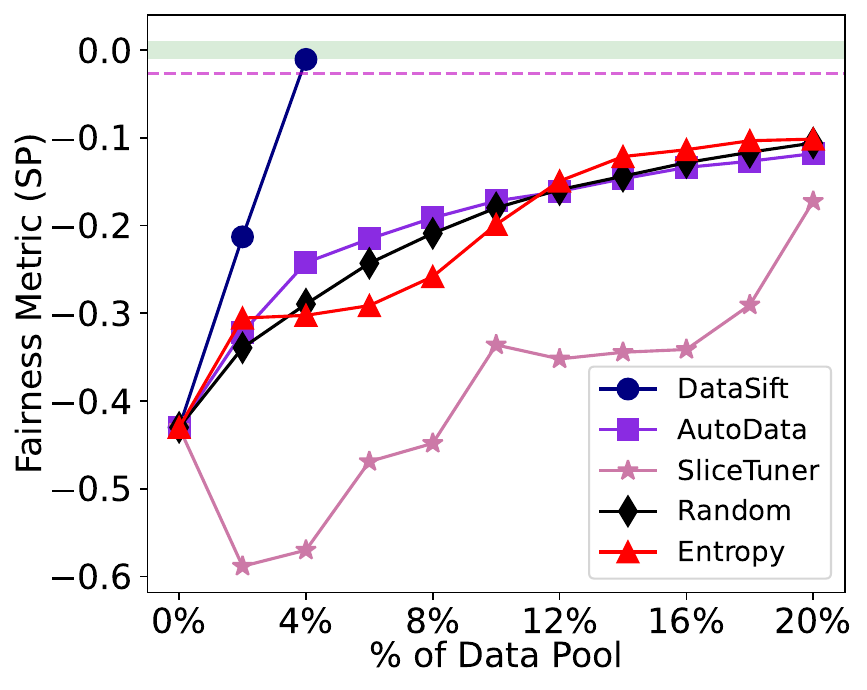}
        \caption{ACSPublicHealth}
        \label{fig: ACSPublicHealth_DataSift}
    \end{subfigure}


    \begin{subfigure}[b]{0.48\columnwidth}
        \centering
        \includegraphics[width=\linewidth]{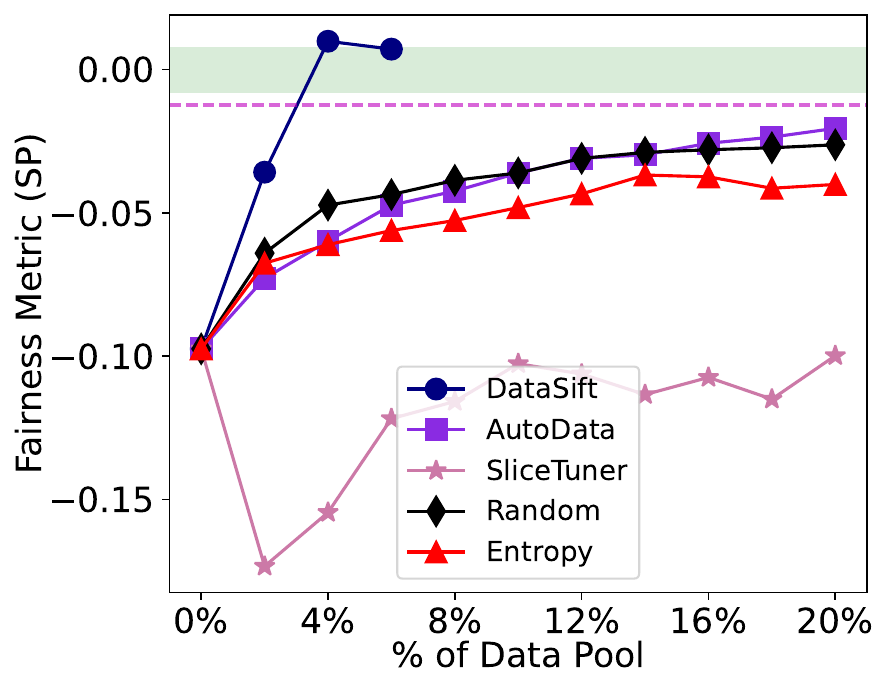}
        \caption{ACSMobility}
        \label{fig: ACSMobility_DataSift}
    \end{subfigure}\hfill
    \begin{subfigure}[b]{0.48\columnwidth}
        \centering
        \includegraphics[width=\linewidth]{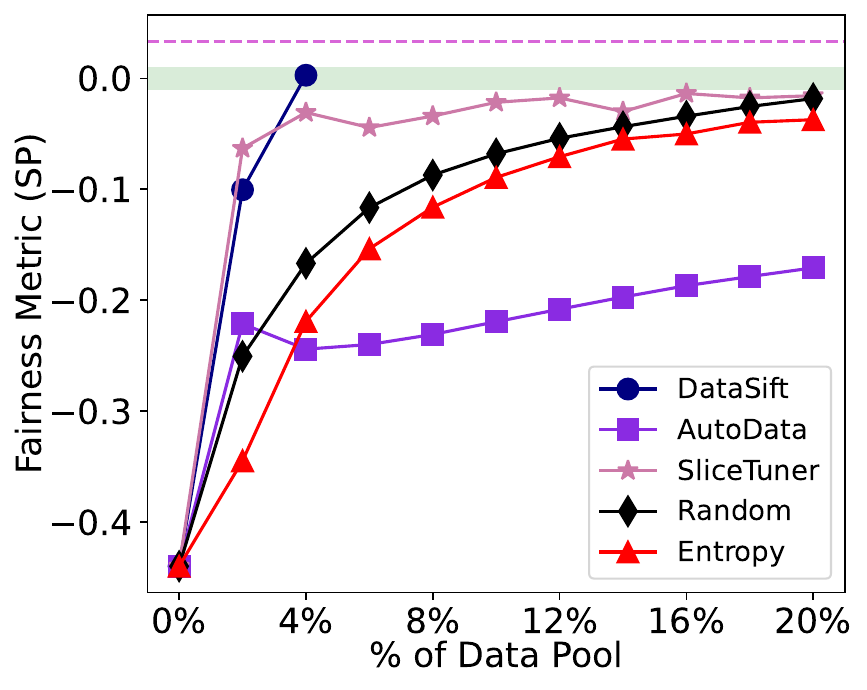}
        \caption{Credit}
        \label{fig: Credit_DataSift}
    \end{subfigure}

    \caption{Comparing \sys with baselines to highlight the effectiveness in achieving fairness. 
    The green shaded area represents the defined zone of unfairness.}
    \label{fig: Datasift Vs Others}
\end{figure}


\subsection{Effectiveness of \sys}
\label{sec:exp:effectiveness}

In this set of experiments, we address RQ1 by comparing the performance of \sys with four competing methods: \random, \entropy, \slicetuner, and \autodata. Unless otherwise specified, all methods are evaluated using Logistic Regression across all datasets. Figure~\ref{fig: Datasift Vs Others} summarizes the results, where the x-axis denotes the fraction of the data pool acquired and the y-axis reports the fairness metric. Values near zero correspond to a fair model: negative values indicate bias against the protected group, while positive values indicate bias favoring the protected group. We define the fairness region as $|\fairness| < \tau = 0.01$, highlighted in green, with each marker in the figure corresponding to the acquisition of a batch.


The baselines \random and \entropy do not incorporate fairness considerations, instead selecting data points without leveraging prior information. Their trajectories are therefore similar across datasets: while adding more points yields marginal fairness improvements relative to the initial model—primarily because of the addition of new data points— they consistently fail to meet the fairness threshold.


\autodata leverages prior information for partition selection but samples batches randomly within partitions. Even when updated with a fairness constraint, it fails to improve fairness due to its distance-centric reward formulation. Consequently, its performance mirrors random selection in most cases. This limitation is most evident in AdultIncome and Credit (Figures~\ref{fig: AdultIncome_DataSift} and~\ref{fig: Credit_DataSift}), where the proportion of positive labels is low (Table~\ref{tab:datasets_summary}). In these settings, random, fairness-agnostic selection from partitions worsens fairness relative to others. effectively injecting additional bias.


Adapting \slicetuner for fairness is non-trivial, as its design relies on power-law distributions. For completeness, we configured it with four slices defined by the cross-product of a binary sensitive attribute and a binary target attribute. While this reduces training loss, it fails to improve fairness across datasets. As shown in Figure~\ref{fig: Datasift Vs Others}, \slicetuner consistently suffers throughout the expansion process, highlighting that demographic representation based expansion alone is insufficient. An exception arises in \textbf{Credit} (Figure~\ref{fig: Credit_DataSift}), where the data composition allows \slicetuner to acquire more protected-positive samples, leading to incidental fairness gains as a byproduct of its loss minimization.


In contrast, \sys integrates MAB-based partition selection with data valuation to identify fairness-influential data points within each partition. As illustrated in Figure~\ref{fig: Datasift Vs Others}, \sys consistently achieves superior performance across all datasets, eliminating unfairness entirely while outperforming all competing methods at nearly every iteration. The sole exception is \textbf{ACSEmployment} (Figure~\ref{fig: ACSEmployment_DataSift}), where \sys shows only marginal gains during the first two iterations; once partition selection changes, however, it rapidly identifies near-optimal batches and surpasses all baselines.

\begin{figure}[t]
    \centering
    \begin{subfigure}[b]{0.23\textwidth}
        \centering
        \includegraphics[width=\textwidth]{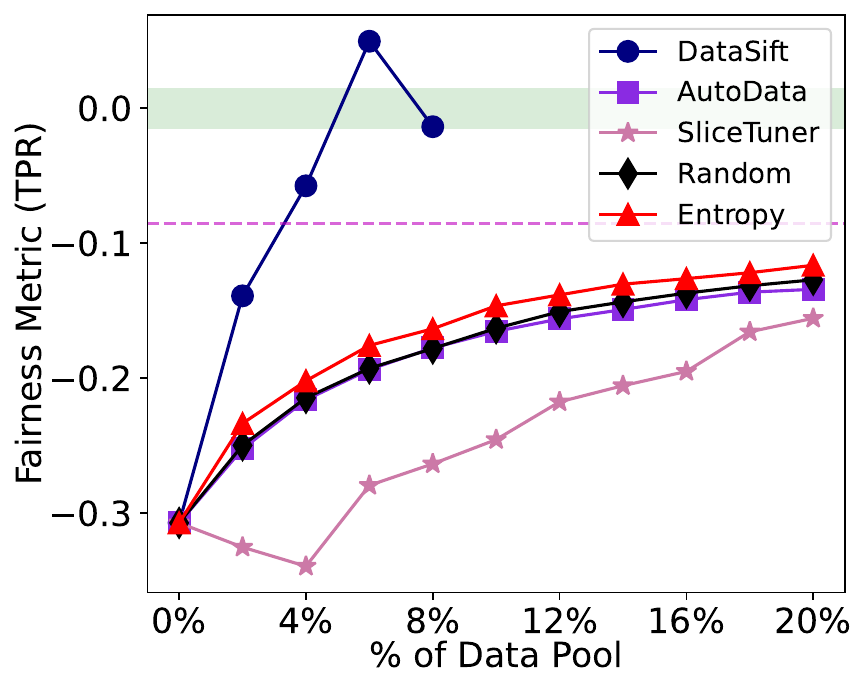}
        \caption{True Positive Rate}
        \label{fig: DataSift_Inf_Income_TPR}
    \end{subfigure}\hfill
    \begin{subfigure}[b]{0.23\textwidth}
        \centering
        \includegraphics[width=\textwidth]{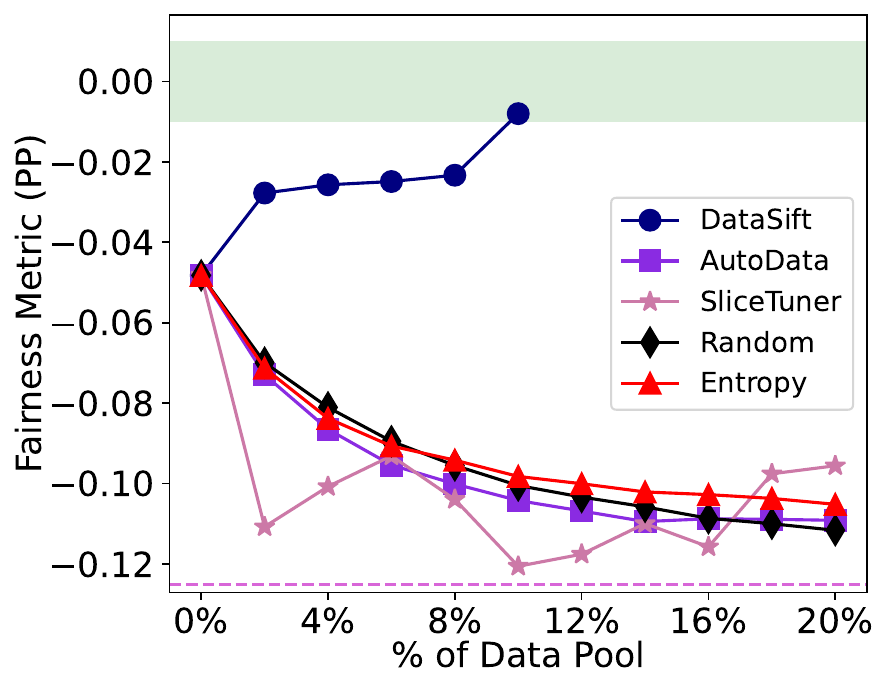}
        \caption{Predictive Parity}
        \label{fig: DataSift_Inf_Income_PP}
    \end{subfigure}
    \caption{Evaluation of \sys over different fairness metrics (\textsf{ACSIncome dataset, magenta dashed line: $\fairness_{\dtrain \cup \datapool}$}).}
    \label{fig: SYS_INF main metric comparisn}
\end{figure}

Overall, \sys attains zero unfairness while using only a small fraction of the data pool. Specifically, it requires just 4--6\% of the data to produce a fully fair model, whereas competing methods consume nearly 20\% without reaching the fairness threshold, ultimately converging to the naïve adding full data pool baseline (dashed magenta line). Adding the entire data pool is computationally costly and still fails to meet the fairness threshold in all cases. By contrast, \sys achieves both fairness and accuracy with far less data, demonstrating its efficiency and effectiveness. 
Note that \sys adopts a \emph{plug-and-play} design for the choice of fairness metric~$\mathcal{F}$. Beyond statistical parity, we evaluate \sys under true positive rate (Figure~\ref{fig: DataSift_Inf_Income_TPR}) and predictive parity (Figure~\ref{fig: DataSift_Inf_Income_PP}) on the ACS Income dataset, with additional results reported in Appendix~\ref{sec: Appendix_metric}. In all the cases, \sys outperforms the baselines and reaches the target fairness threshold more efficiently.

\begin{figure}[t]
    \centering
    \begin{subfigure}[b]{0.23\textwidth}
        \centering
        \includegraphics[width=\textwidth]{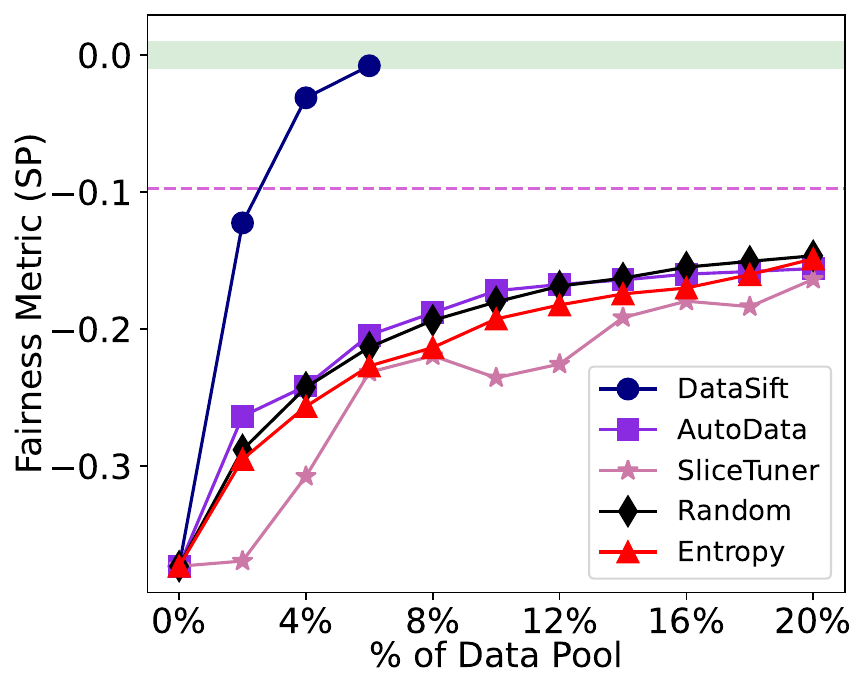}
        \caption{Support Vector Machine}
        \label{fig: DataSift_Inf_SVM}
    \end{subfigure}\hfill
    \begin{subfigure}[b]{0.23\textwidth}
        \centering
        \includegraphics[width=\textwidth]{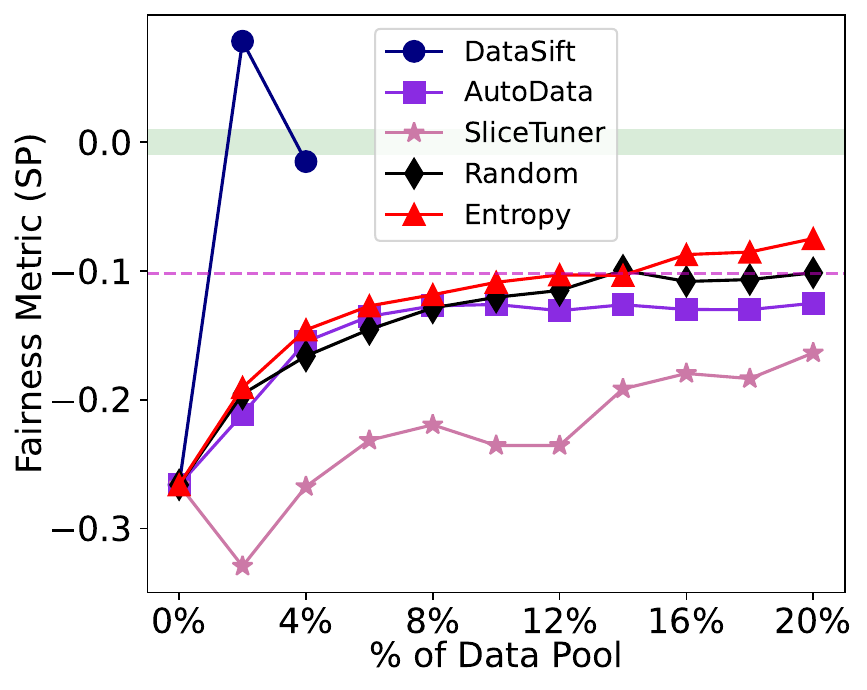}
        \caption{Neural Network}
        \label{fig: DataSift_Inf_NN}
    \end{subfigure}
    \caption{\sys's evaluation across parametric models.}
    \label{fig: SYS_INF parametric model comparisn}
\end{figure}

\vspace{1mm}
\noindent{\textbf{Model Generalization.}} 
The reliance on influence functions limits the applicability of \sys to parametric models. To evaluate robustness within this scope, Figure~\ref{fig: SYS_INF parametric model comparisn} reports results on the \textbf{ACSIncome} dataset for two additional models: Support Vector Machines (SVM) and Neural Networks. For SVM (Figure~\ref{fig: DataSift_Inf_SVM}), \sys requires one additional batch compared to Logistic Regression (Figure~\ref{fig: ACSIncome_DataSift}), yet the overall trajectory and outcome remain closely aligned.
The neural network case (Figure~\ref{fig: DataSift_Inf_NN}) is more dynamic: the model initially shows bias toward the protected group, but after the first batch acquisition, the fairness metric shifts toward the privileged group. In this scenario, \sys adapts effectively, making informed acquisition decisions that progressively mitigate bias and ultimately yield a fair model.
These findings highlight the robustness of \sys in correcting unfairness regardless of its direction or the demographic group affected, underscoring its versatility across parametric models. To extend applicability further, model-agnostic valuation methods such as Data Shapley~\cite{pmlr-v97-ghorbani19c} could be incorporated, enabling generalization to non-parametric models.

\vspace{1mm}
\noindent{\textbf{Effect on model accuracy.}}  To improve fairness, algorithms prioritize diversifying the training set by adding data from partitions that capture under-represented subgroups. This prevents the model from overfitting to privileged groups while reducing bias against marginalized ones. However, as Table~\ref{tab:datasets_summary} indicates, most datasets are dominated by the privileged group; thus, solely fairness-focused acquisition can risk accuracy degradation.
Our reward design explicitly balances this trade-off, enabling \sys to maintain accuracy while achieving fairness. Table~\ref{tab:accuracy_datasift_vs_base} reports the final accuracy after data acquisition. \sys often improves accuracy relative to baselines and, in datasets such as \textbf{AdultIncome} and \textbf{Mobility}, preserves accuracy within an acceptable tolerance. Methods such as \entropy and \slicetuner, which prioritize reducing model uncertainty, achieve slightly higher accuracy but at the expense of fairness, while \autodata suffers accuracy degradation as its constraint is modified to account for fairness. 
\begin{table}[h]
    \centering
    \caption{Accuracy of Logistic Regression models across datasets trained after adding selectively acquired data points from the corresponding pool as determined by each method. (Bold: highest, Underlined: second highest)}
    \label{tab:accuracy_datasift_vs_base}
    \renewcommand{\arraystretch}{1.2}
    \setlength{\tabcolsep}{2pt}
    \fontsize{7pt}{7pt}\selectfont
    \begin{tabular}{@{}lrrrrrr@{}}
        \toprule
        \textbf{Dataset} & \multicolumn{6}{c}{\textbf{Methods}} \\
        \cmidrule(lr){2-7}
        & \textbf{Initial} &  \textbf{Random} & \textbf{Entropy} & \textbf{AutoData} & \textbf{SliceTuner} & \textbf{DataSift} \\ \midrule
        \textbf{AdultIncome}   & 0.80  & 0.78 & 0.77 & 0.71 & \textbf{0.80} & \underline{0.78} \\
        \textbf{ACSIncome}     & 0.70 & \textbf{0.78} & 0.77 & \underline{0.78} & 0.77 & 0.77 \\
        \textbf{Employment}    & 0.69 & \underline{0.75} & \textbf{0.75} & 0.74 & 0.74 & 0.73 \\
        \textbf{PublicHealth}  & 0.65 & \textbf{0.69} & 0.68 & 0.68 & \underline{0.68} & 0.68 \\
        \textbf{Mobility}      & 0.77 & \underline{0.77} & \textbf{0.77} & 0.77 & 0.75 & 0.75 \\
        \textbf{Credit}        & 0.73 & 0.93 & \underline{0.93} & 0.92 & 0.91 & \textbf{0.93} \\
        \bottomrule
    \end{tabular}
     
\end{table}

\begin{figure}[t]
    \centering

    \begin{subfigure}[t]{\columnwidth}
        \centering
        \begin{minipage}{0.48\columnwidth}
            \centering
            \includegraphics[width=\linewidth]{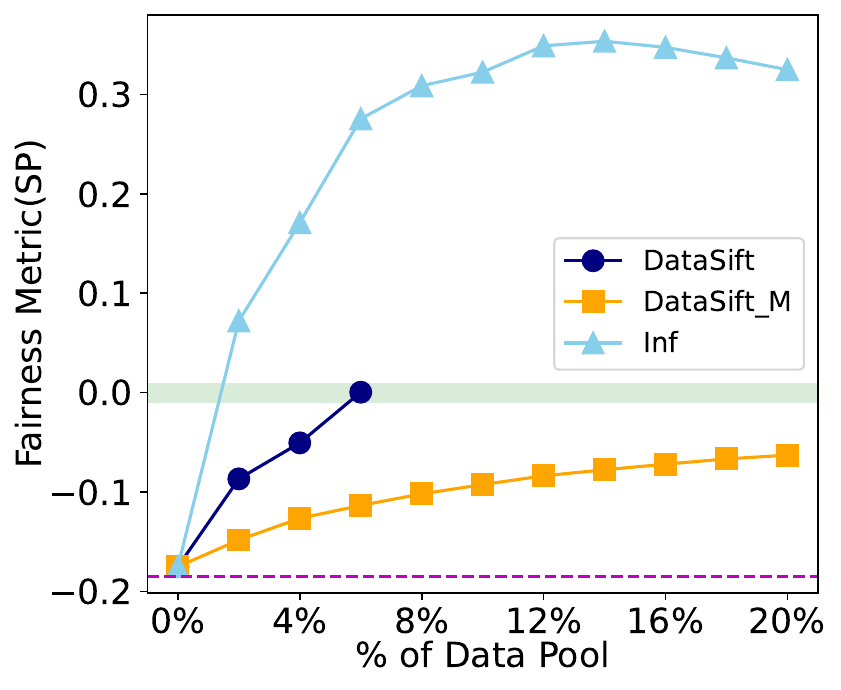}
        \end{minipage}\hfill
        \begin{minipage}{0.48\columnwidth}
            \centering
            \includegraphics[width=\linewidth]{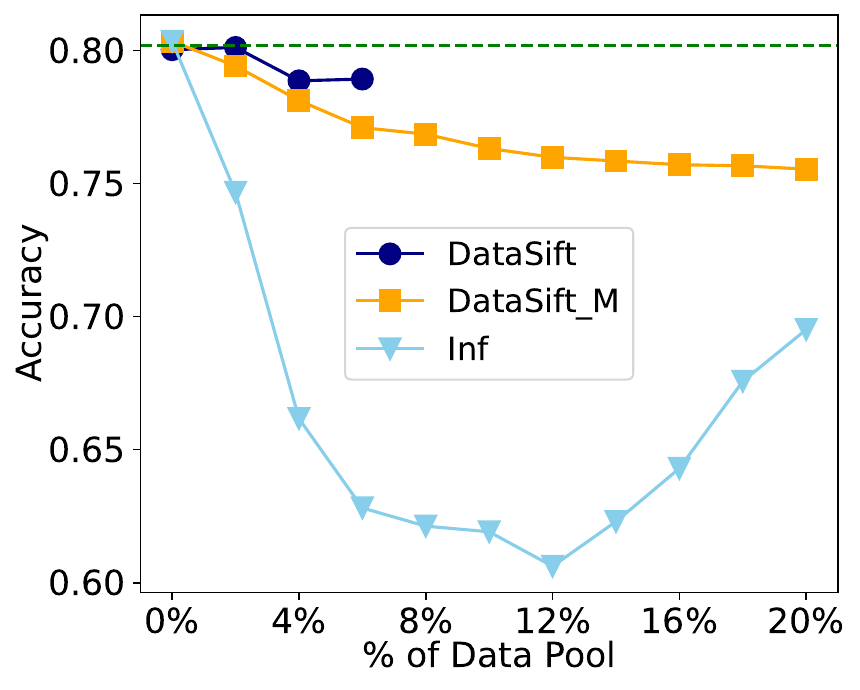}
        \end{minipage}

        \caption{AdultIncome}
        \label{fig: adult_income_inf}
    \end{subfigure}


    \begin{subfigure}[t]{\columnwidth}
        \centering
        \begin{minipage}{0.48\columnwidth}
            \centering
            \includegraphics[width=\linewidth]{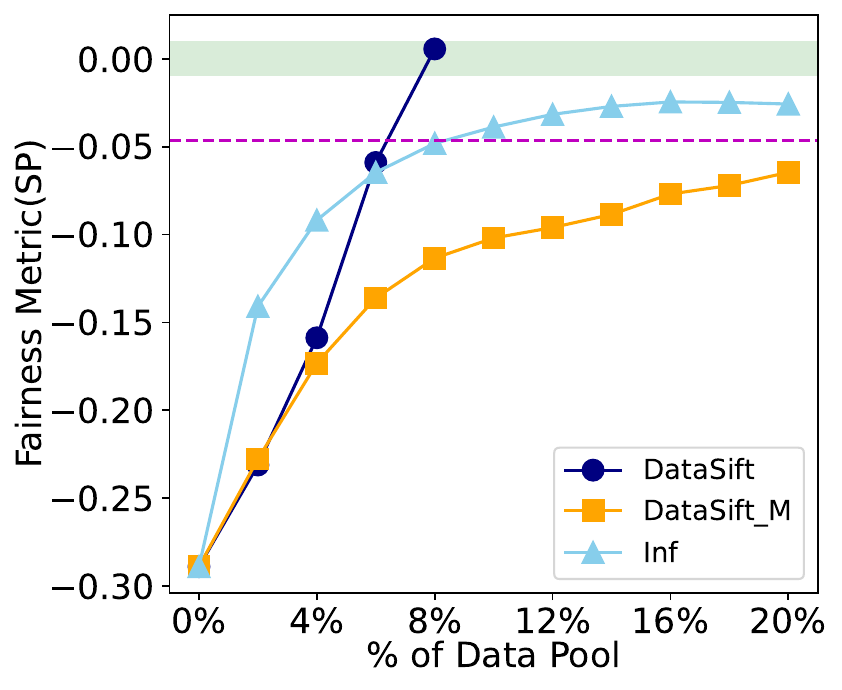}
        \end{minipage}\hfill
        \begin{minipage}{0.48\columnwidth}
            \centering
            \includegraphics[width=\linewidth]{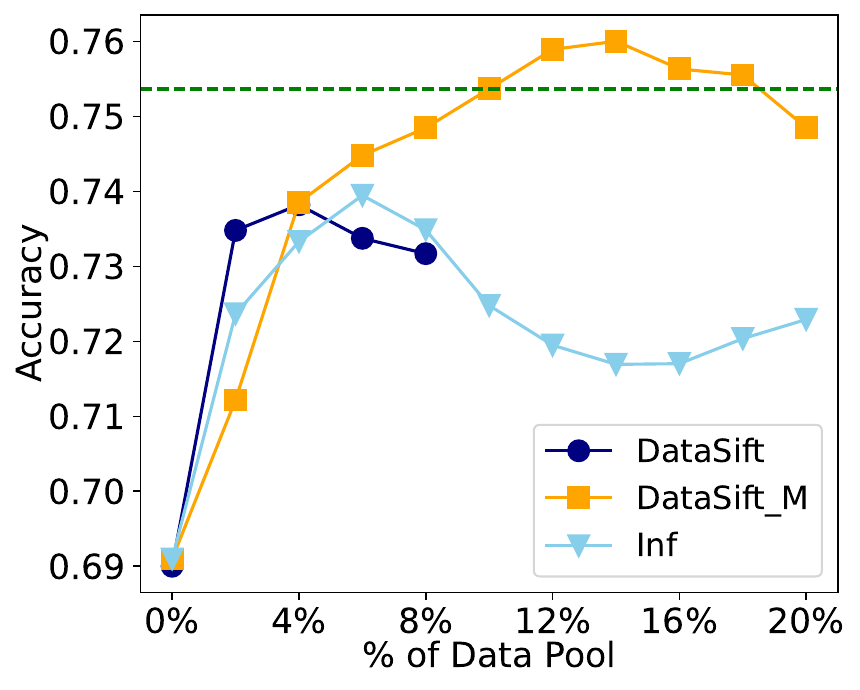}
        \end{minipage}

        \caption{ACSEmployment}
        \label{fig: acsemployment_inf}
    \end{subfigure}


    \begin{subfigure}[t]{\columnwidth}
        \centering
        \begin{minipage}{0.48\columnwidth}
            \centering
            \includegraphics[width=\linewidth]{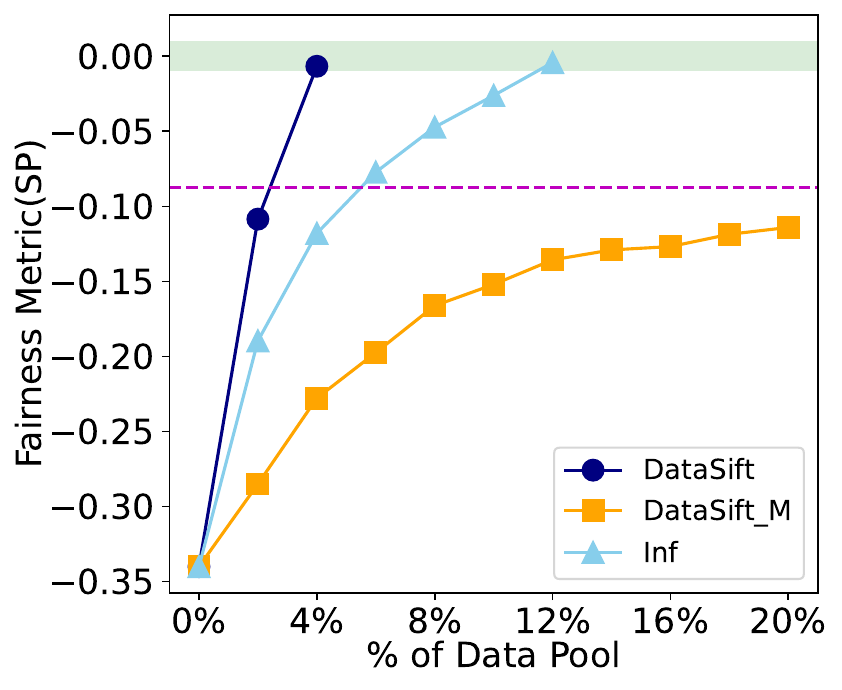}
        \end{minipage}\hfill
        \begin{minipage}{0.48\columnwidth}
            \centering
            \includegraphics[width=\linewidth]{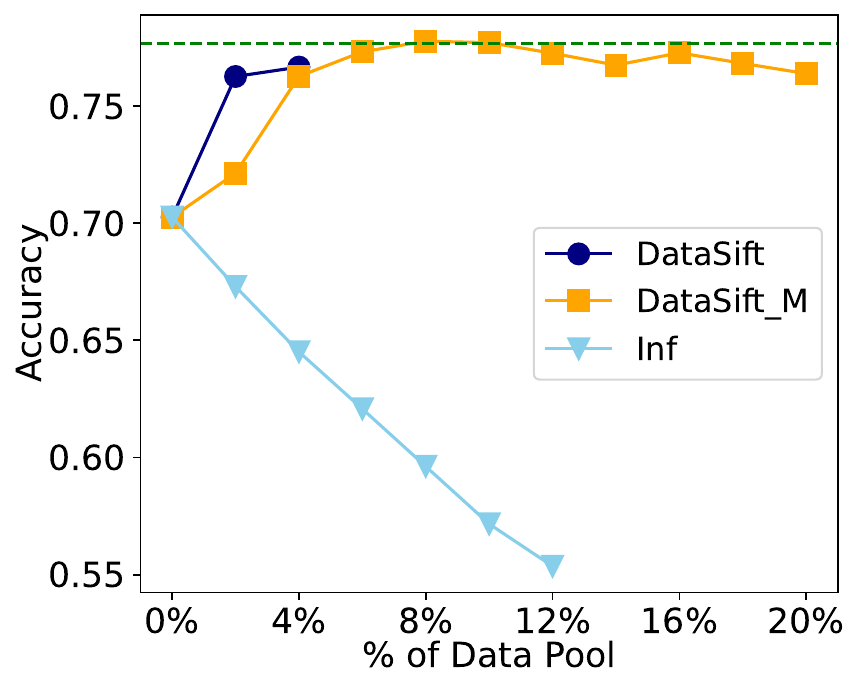}
        \end{minipage}

        \caption{ACSIncome}
        \label{fig: acsincome_inf}
    \end{subfigure}

    \caption{Comparing \sys\ with \sysmab\ and \sysinf\ to highlight the rationale behind \sys's formation. For each dataset, the left panel shows fairness improvement and the right panel shows accuracy. The magenta and green dotted lines indicate fairness and accuracy, respectively, after adding the entire data pool.}
    \label{fig: DataSift_Inf_Vs_Others}
\end{figure}

\subsection{Importance of \sys's components}
\label{sec:valuation_exp}

In this set of experiments, we address RQ2 by investigating the respective contributions of the MAB component and the data valuation component within the \sys framework. Specifically, we compare \sys against two baselines: (i) \sysmab, a variant of \sys, which relies solely on the MAB framework followed by random batch selection instead of data valuation, and (ii) \sysinf, which utilizes only data valuation (no MAB) for data selection. This comparison underscores the critical importance of integrating both MAB and data valuation in achieving the effectiveness of \sys. We report performance results—measured in terms of both accuracy and fairness—on three representative datasets (out of six), chosen for their distinct empirical behaviors observed during experiment.


\begin{figure*}[t]
    \centering
    \begin{subfigure}[b]{0.24\textwidth}
        \centering
        \includegraphics[width=\textwidth]{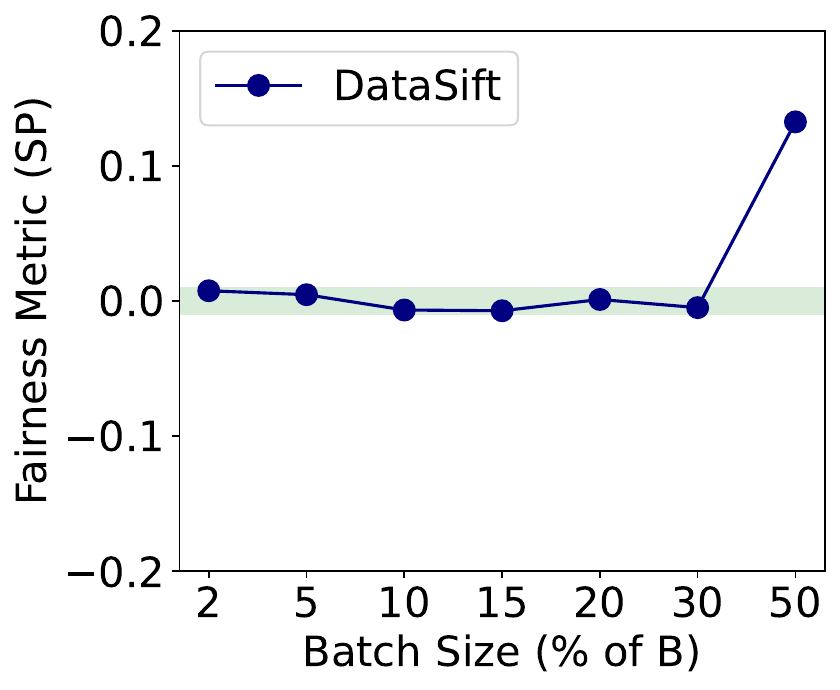}
        \caption{Effect of batch size}
        \label{fig:mini-batch size}
    \end{subfigure}\hfill
    \begin{subfigure}[b]{0.24\textwidth}
        \centering
        \includegraphics[width=\textwidth]{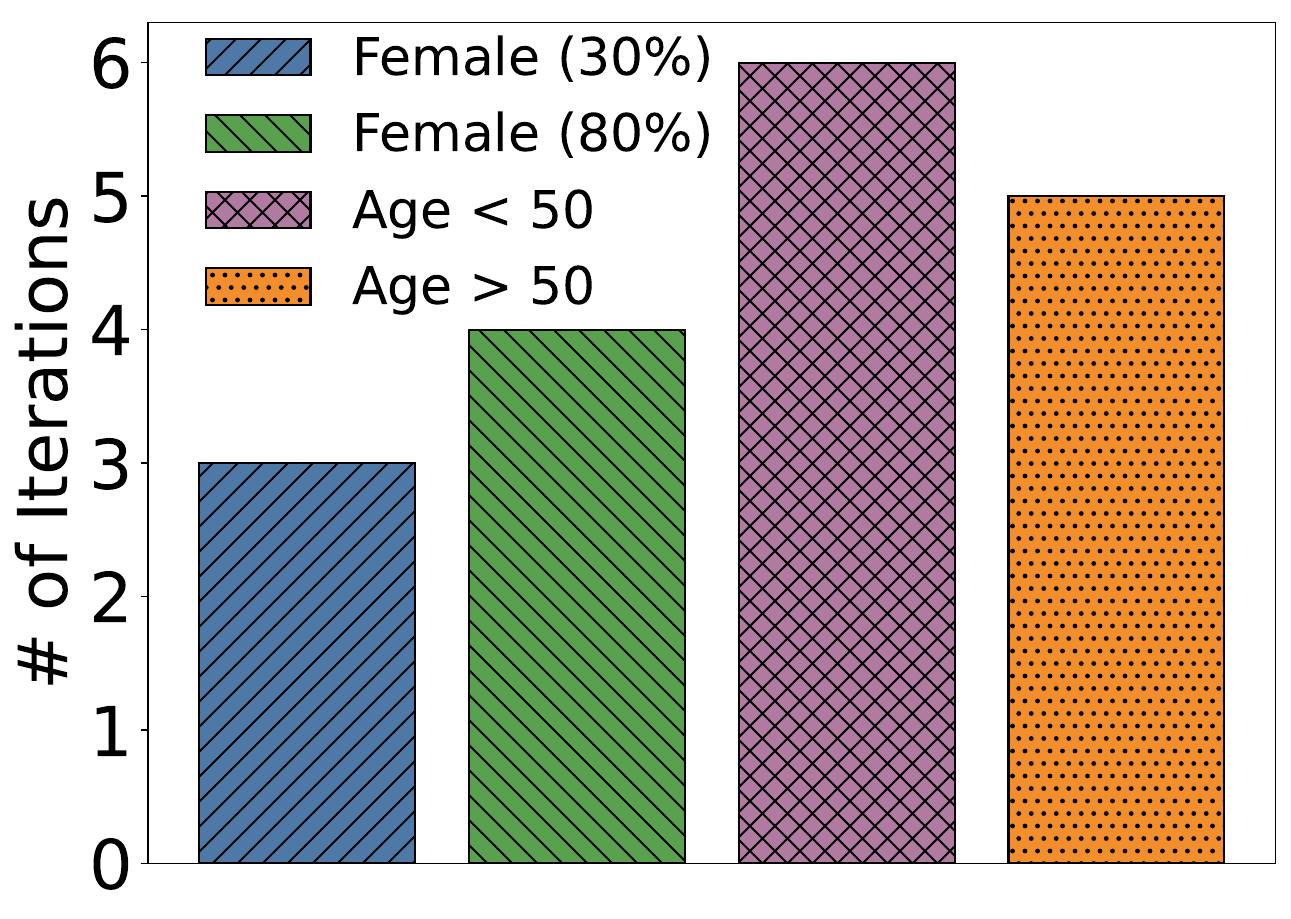}
        \caption{Effect of data pool distribution}
        \label{fig:datapool_dist}
    \end{subfigure}\hfill
    \begin{subfigure}[b]{0.24\textwidth}
        \centering
        \includegraphics[width=\textwidth]{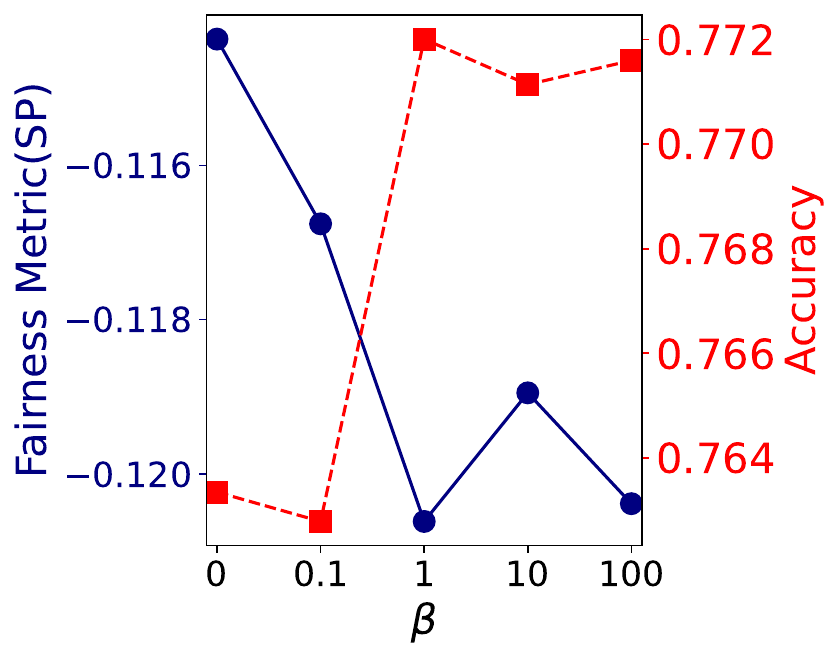}
        \caption{Effect of $\beta$}
        \label{fig:beta}
    \end{subfigure}\hfill
    \begin{subfigure}[b]{0.24\textwidth}
        \centering
        \includegraphics[width=\textwidth]{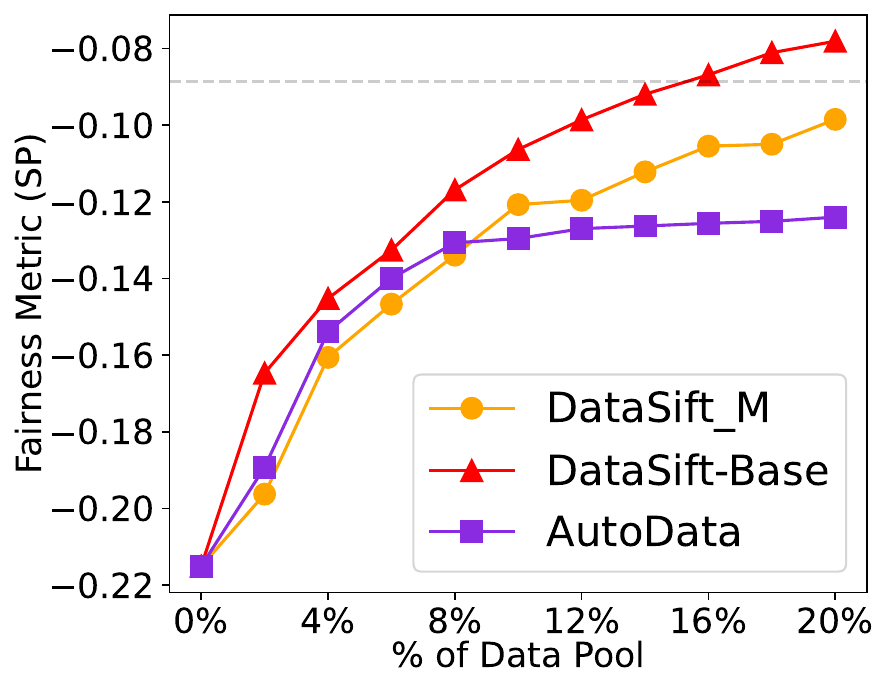}
        \caption{Reward Score selection}
        \label{fig:Reward_score}
    \end{subfigure}
    \caption{Effect of Hyper-parameters and design choices (ACSIncome dataset)}
    \label{fig:effect_hyperparameter_design_choices}
    
\end{figure*}

We report logistic regression results across all datasets. Figure~\ref{fig: DataSift_Inf_Vs_Others} follows the standard layout, with the x-axis denoting the budget fraction and the y-axis showing the evaluation metrics. The first column reports fairness outcomes, where values within the green-shaded region (i.e., close to zero) indicate a fair model, while the second column presents the corresponding classification accuracy.

For the \textbf{AdultIncome} dataset (Figure~\ref{fig: adult_income_inf}), \sys achieves a zero-bias model by incorporating only 6\% of the data pool, whereas \sysmab requires 20\% and, despite substantial improvement, still fails to reach the fairness region. A key limitation of \sysmab is its reliance on random batch selection within the partition with the highest potential reward. 
While the partition itself may be promising, not all points within it are beneficial—some can even worsen fairness. 
Consequently, repeated random sampling causes \sysmab to perform only marginally better than the \random baseline and other expansion methods (Figure~\ref{fig: ACSIncome_DataSift}). In contrast, the valuation-based selection in \sys effectively addresses this issue and drives the model to zero bias.

An interesting pattern emerges with the \sysinf method. Although the model initially exhibits bias favoring the privileged group (negative metric values), as \sysinf continues to exploit the most influential batches from data pool, the model becomes increasingly biased in the opposite direction (positive values). This reversal occurs because the most influential data points often come from the protected group; repeated exploitation of the same information amplifies bias in opposite direction rather than mitigating it. By dynamically balancing exploration and exploitation, \sys avoids this pitfall and converges to a fair model.
Finally, we note that if \sysinf is stopped early (e.g., at 0–2\% expansion), fairness can be temporarily improved, but at the cost of a 13\% drop in accuracy (see corresponding accuracy at Figure~\ref{fig: adult_income_inf}). By contrast, \sys consistently preserves accuracy while driving the fairness metric toward zero across the entire expansion process.

For the \textbf{ACSEmployment} dataset in Figure~\ref{fig: acsemployment_inf}, \sys achieves the fairness goal by incorporating only 8\% of the data pool, whereas \sysinf requires 20\% and still fails to reach the fairness region—though it performs better than \sysmab and the other baselines (Figure~\ref{fig: ACSEmployment_DataSift}). As in earlier cases, \sysmab cannot achieve full fairness due to its influence-agnostic batch selection, but it marginally outperforms the simpler baselines.
For this substantially larger dataset, \sysinf improves fairness slowly, even when adding the most influential points, because batch interdependencies diminish actual gains (refer to Section~\ref{sec:batch_construction}). This shows that batch influence is not simply additive, and repeatedly exploiting top-$K$ influential batches yields only marginal improvements. While this strategy enhances fairness to some extent, it does so at the cost of lower accuracy relative to other methods (see accuracy plot).
In contrast, when the performance of \sys declines compared to \infs in the initial 4\% of data pool, it transitions to a new partition that provides additional information, resulting in a sharp improvement in fairness in next iteration. Moreover, \sys gradually improves accuracy beyond its initial level (around 6\%), while \sysmab attains slightly higher accuracy (around 9\%)  by incorporating more random data points, albeit at the expense of fairness.

For the \textbf{ACSIncome} dataset (Figure~\ref{fig: acsincome_inf}), both \sys and \sysinf achieve the fairness goal, requiring only 4\% and 12\% of the data pool, respectively—again highlighting the efficiency of \sys. While \sysinf substantially outperforms other baselines in fairness (Figure~\ref{fig: ACSIncome_DataSift}), it suffers a drastic accuracy loss of nearly 23\%. As observed for the \textbf{ACSEmployment} dataset, this decline arises from repeatedly exploiting top influential points, which are disproportionately drawn from the protected group. The resulting redundancy improves fairness but fails to enhance predictive power, thereby degrading accuracy. In contrast, \sys leverages the integration of MAB with data valuation to selectively acquire high-quality data points from unexplored regions of the pool. Consequently, \sys attains a 10\% increase in accuracy while maintaining a fair model. Although \sysmab exhibits a comparable accuracy gain, its fairness level remains substantially lower.

\subsection{Ablation analysis}
\label{subsec:hyperparameter}
This section reports the sensitivity of our solutions to the hyper-parameters and other design choices. All experiments are conducted on the \textbf{ACSIncome} dataset in this section.
 


\vspace{1mm}\noindent\textbf{Effect of batch size.}
In our variations of \sys, we define the batch size as a percentage of the expansion threshold, denoted by $B$. Figure~\ref{fig:mini-batch size} illustrates the final fairness metric (y-axis) for a logistic regression model trained on selectively expanded data from the \textbf{ACSIncome} dataset, while varying the batch size from $2\%$ to $50\%$ of the total budget (x-axis).
The results show that the performance of \sys remains stable across batch sizes of up to 30\%, consistently achieving fairness regardless of the specific choice. However, excessively large batch sizes—around 50\% of $\budget$—lead to a decline in performance, introducing bias in the opposite direction. This degradation occurs because \sysinf, under large batch settings, selects predominantly influential data points concentrated in the protected group. With only two large acquisitions possible, the framework cannot sufficiently balance these effects, resulting in residual bias. By contrast, smaller batch sizes (up to 30\% of $\budget$) enable multiple evaluation rounds, allowing \sys to adjust its selections iteratively and ultimately converge to a fair model.

\vspace{1mm}\noindent\textbf{Effect of data pool distribution.} \sys does not assume that the data pool follows the same distribution as the training data. To substantiate this claim, we construct data pools whose distributions differ markedly from the training set. While the original training data has $48\%$ representing the protected group, and ages spanning from $17$ to $94$, we consider the following four scenarios for the data pool: (a)~fraction of protected group $=0.3$ (KL Divergence=0.19), (b)~fraction of protected group $=0.8$ (KL Divergence=0.26), (c)~age-restricted data pool 1 (age $<50$, KL Divergence=0.39), and (d)~age-restricted data pool 2 (age $\geq50$, KL Divergence=0.96). Across all four settings, \sys consistently attains the target fairness threshold; accordingly, Figure~\ref{fig:datapool_dist} reports only the number of acquisition iterations required to reach the goal.
In gender-skewed pools, \sys requires three and four iterations (compared to two when the pool follows the training distribution), whereas age-restricted pools require five and six iterations. The increased number of iterations in these cases are expected, as the pool size is substantially reduced (one-third of the original), making the finding of influential samples more gradual. Nevertheless, \sys consistently improves fairness across all settings, while baseline methods degrade under distributional shifts (see Appendix~\ref{sec: datapool_appendix} for a detailed comparison). These results confirm that \sys acquisition is driven by observed utility signals through influence-based candidate subsets of pool, rather than by assumptions of global distributional alignment.


\vspace{1mm}\noindent\textbf{Effect of
$\beta$.}
In Figure~\ref{fig:beta}, we analyze the effect of varying $\beta$, the parameter that balances fairness and accuracy in the reward function defined in Equation~\ref{eq:reward_eq}. Setting $\beta=0$ places exclusive emphasis on fairness when shaping the reward distribution, whereas larger values of $\beta$ progressively shift the focus toward accuracy. As expected,figure~\ref{fig:beta} shows that increasing $\beta$ reduces fairness while improving accuracy. Hence, the choice of $\beta$ should be guided by the requirements of the specific application, with domain expertise playing a central role in determining the appropriate trade-off.


\vspace{1mm}\noindent\textbf{Effect of choice of reward score.} \sys builds on the MAB paradigm, which requires designing a reward score to approximate the reward distribution across partitions. \autodata employs a distance-centric reward optimized for accuracy, whereas \sys incorporates base-rate differences among sensitive groups into the reward design (Equation~\ref{eq:reward_eq}) to account for both fairness and accuracy. For completeness, we report results for \sysmab.
As shown in Figure~\ref{fig:Reward_score}, \autodata fails to yield stable fairness improvements. Reward scores based solely on base rates in our framework, \sys-Base, substantially improve fairness but often sacrifice accuracy. By contrast, the reward function in \sysmab effectively balances the trade-off, promoting equity and performance.

\begin{figure}[t]
    \centering
    \begin{subfigure}[b]{0.195\textwidth}
        \centering
        \includegraphics[width=\textwidth]{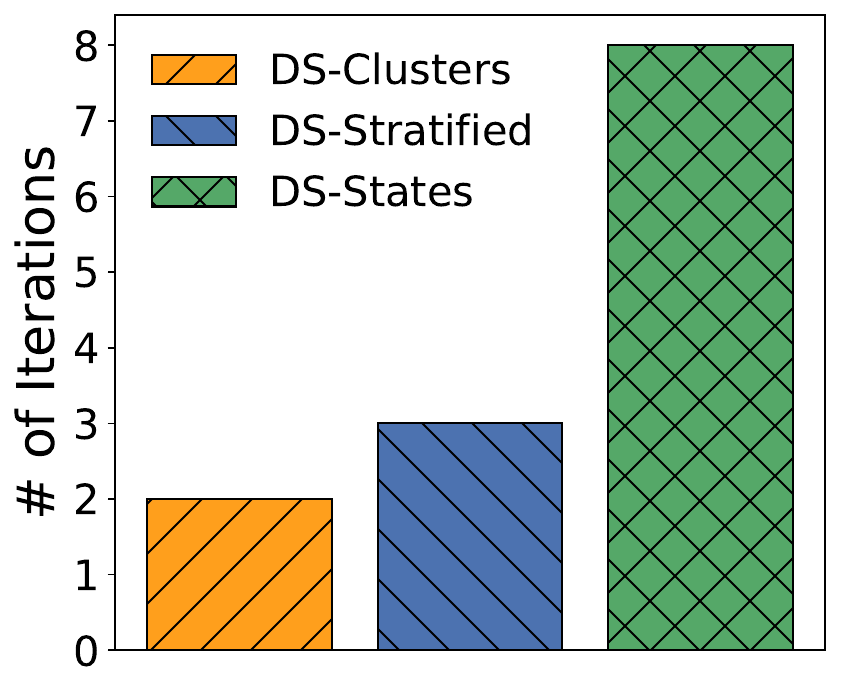}
        \caption{Effect of partitioning.}
        \label{fig:Partitioning VS Clusters}
    \end{subfigure}\hfill
    \begin{subfigure}[b]{0.275\textwidth}
        \centering
        \includegraphics[width=\textwidth]{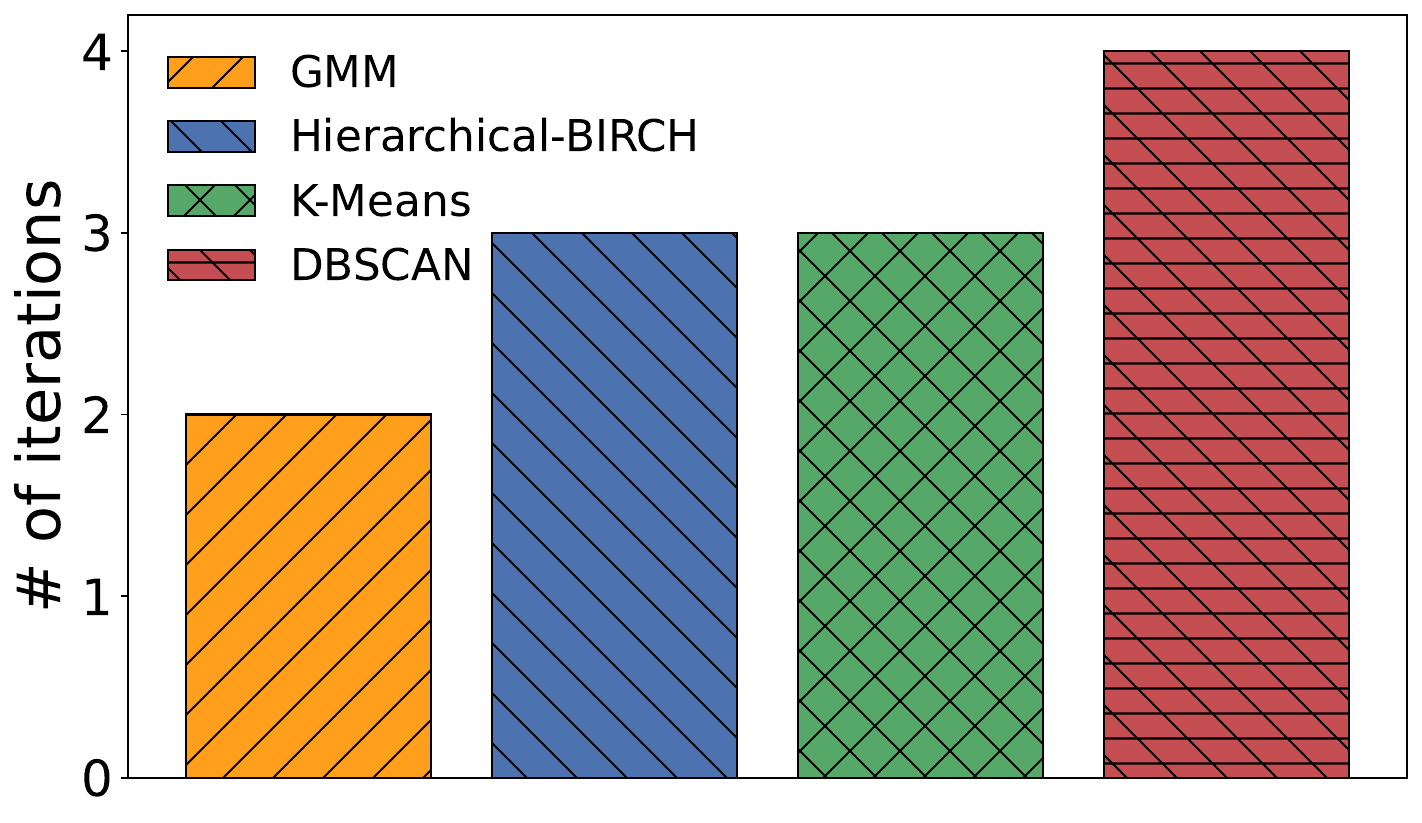}
        \caption{Effect of clustering methods}
        \label{fig:clustering_ablation_main}
    \end{subfigure}
    \caption{Effect of Partitioning and Clustering Sensitivity on Fairness Convergence. (DS: \sys) }
    \label{fig:partitioning_clustering}
\end{figure}

\vspace{1mm}\noindent\textbf{Effect of partitioning.} 
\sys requires partitioning the data space into smaller groups, which can be accomplished either automatically (e.g., via clustering) or by leveraging domain-specific expertise. In Figure~\ref{fig:Partitioning VS Clusters}, we evaluate \sys under three partitioning strategies: (i) clustering into the optimal number of groups using a Gaussian Mixture Model (GMM), (ii) geographic partitioning by state (treating the four largest states as separate partitions), and (iii) stratified partitioning into four subsets defined by the cross-product of a binary sensitive attribute and a binary outcome.

\begin{figure}[b]
    \centering
    \begin{subfigure}[b]{0.23\textwidth}
        \centering
        \includegraphics[width=\textwidth]{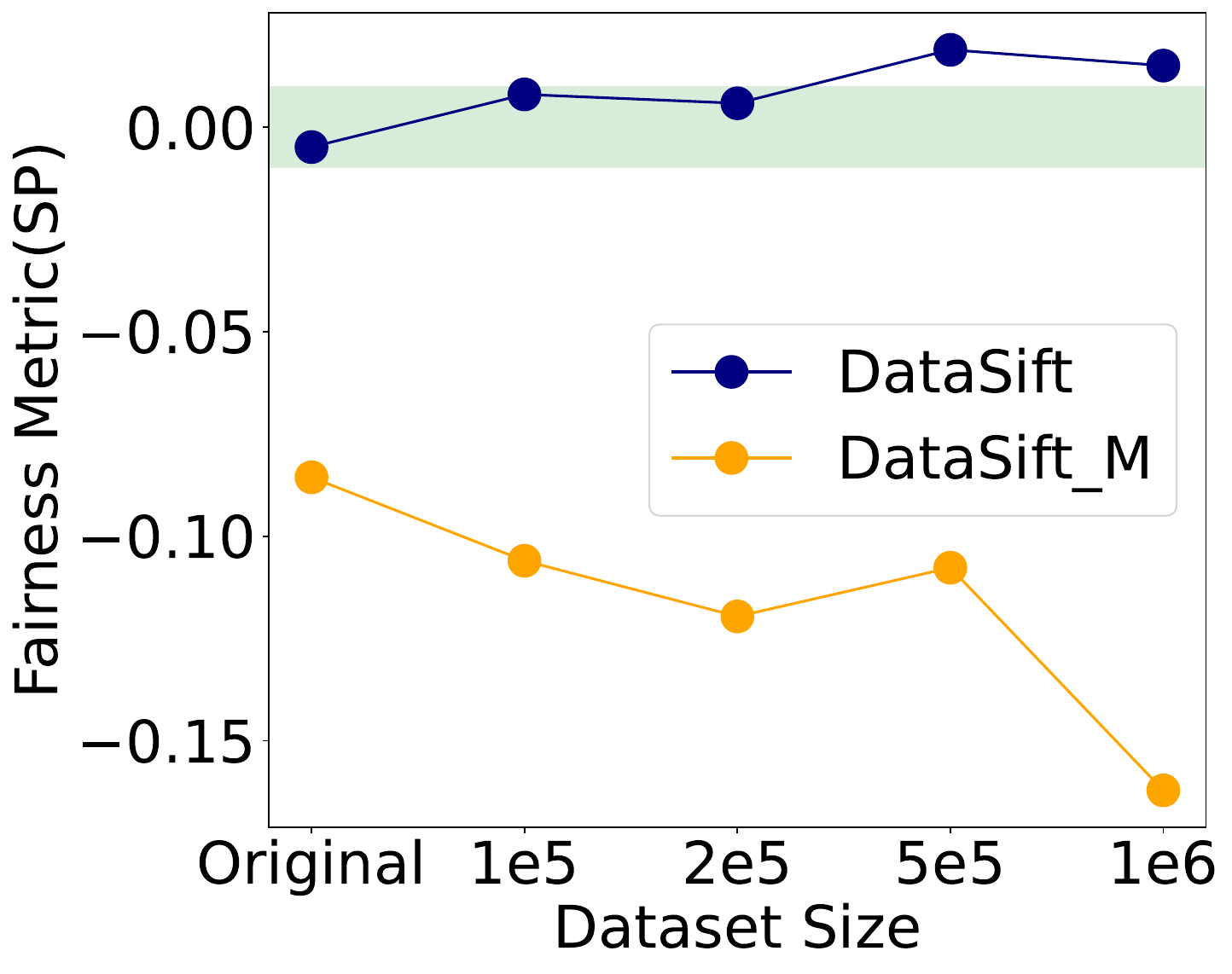}
        \caption{Effectiveness}
        \label{fig: Synthetic Dataset Effectiveness}
    \end{subfigure}\hfill
    \begin{subfigure}[b]{0.23\textwidth}
        \centering
        \includegraphics[width=\textwidth]{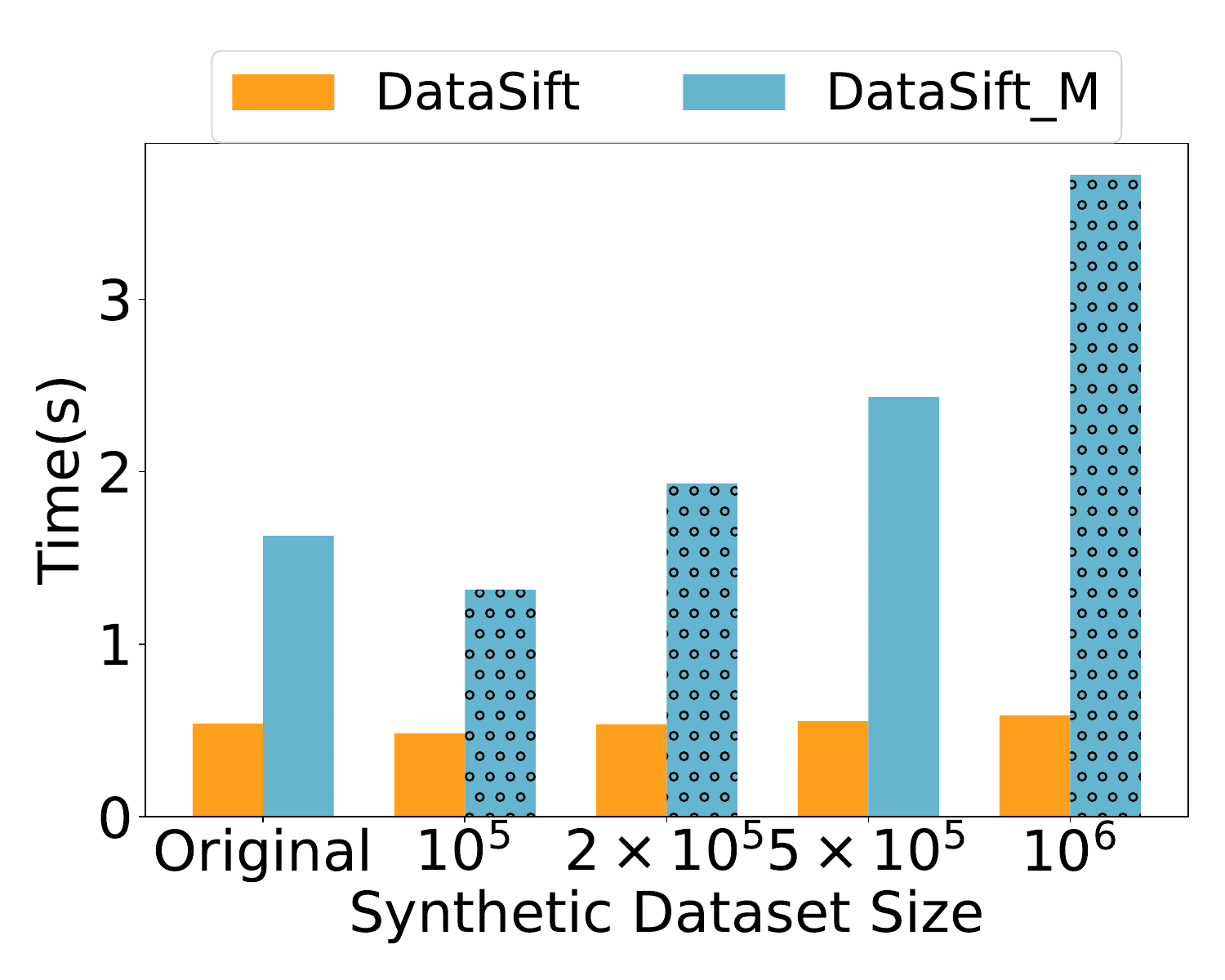}
        \caption{Efficiency}
        \label{fig: Synthetic Dataset Efficiency}
    \end{subfigure}
    \caption{Scalability on synthetic dataset.}
    \label{fig: Synthetic Dataset}
    
\end{figure}

\begin{figure*}[t]
    \centering
        \includegraphics[width=.97\textwidth]{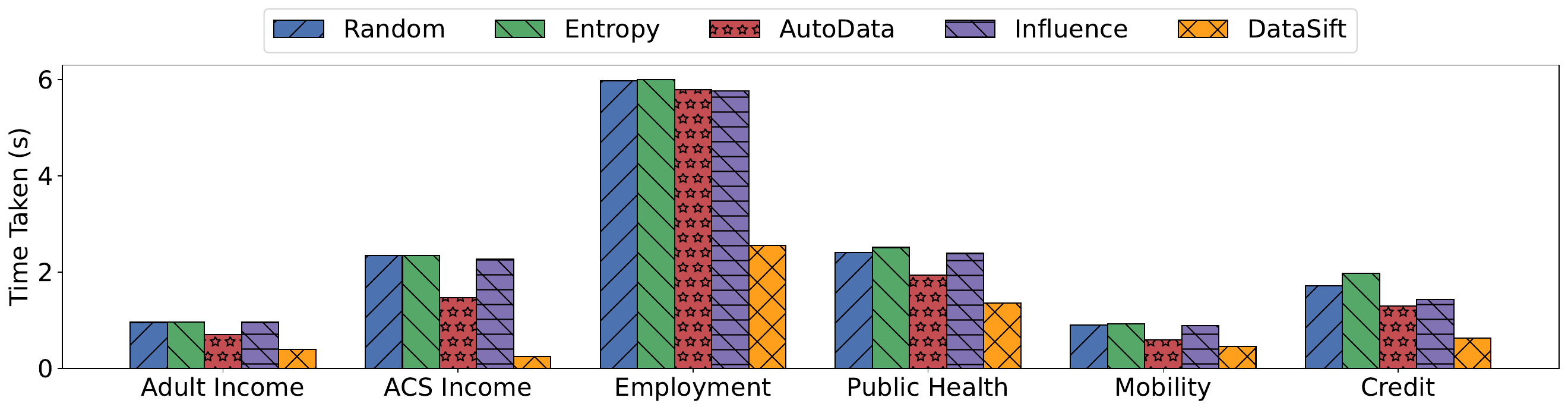}
    \caption{Efficiency on real-world datasets.}
    \label{fig:Efficiency}
\end{figure*}

Across all three strategies, \sys achieves a completely fair model; thus, we report the required number of iterations in Figure~\ref{fig:Partitioning VS Clusters}. \sys-Cluster and \sys-Stratified reach fairness in 2 and 3 iterations, respectively, while \sys-States requires 8. This inefficiency stems from the four largest states having highly similar distributions, limiting the information gained when switching partitions. In contrast, clustering and stratification create more diverse partitions, enabling faster convergence. Accordingly, we base all reported experiments on clustering.

We further conduct an ablation study comparing GMM~\cite{figueiredo2002unsupervised}, k-means~\cite{jain2010data}, DBSCAN~\cite{ester1996density}, and Hierarchical-BIRCH~\cite{zhang1997birch} on the same dataset (Figure~\ref{fig:clustering_ablation_main}). GMM, K-means, and BIRCH show consistent performance, with comparable fairness gains and only minor differences in convergence speed, indicating that \sys is largely partition-agnostic when the pool is divided into a small number of stable, coarse-grained groups. DBSCAN requires slightly more iterations, as density-based clustering often produces many small clusters, expanding the exploration space and increasing reward variance. Overall, simple and scalable methods such as GMM and K-means are sufficient in practice, though \sys is not tied to any specific clustering strategy.

\vspace{1mm}\noindent \textbf{Implementation and parameter guidelines.}
While \sys introduces multiple components, its implementation remains practical, as most parameters have intuitive interpretations and stable defaults. The batch size controls the granularity of data acquisition, the fairness threshold specifies the desired tolerance level, and the exploration parameters follow standard multi-armed bandit practices. In our experiments, a single set of parameters was used across datasets, and in ablation analysis, we observed that performance is robust to moderate parameter variations, indicating limited sensitivity to parameter variations. In practice, we recommend starting from these default settings and adjusting the parameters based on computational budget and fairness requirements.

\subsection{Scalability Analysis}
This section answers RQ4 that evaluates the different methods with respect to variations in dataset and data pool sizes.

Using the \textbf{AdultIncome} dataset, we synthetically generate additional data points to vary the pool size from the original dataset up to $10^6$ records. Throughout this study, the budget and batch size remain fixed. Figure~\ref{fig: Synthetic Dataset Effectiveness} reports the fairness achieved by logistic regression models trained with \sys and \sysmab under different pool sizes. The y-axis represents the ultimate fairness achieved by the model under various scenarios. The results show that \sys maintains nearly consistent performance as the data pool grows, whereas \sysmab experiences a slight decline. Unlike \sysmab, which relies on random batch selection and is thus influenced by the data pool’s distribution, \sys consistently identifies influential points irrespective of pool size, leading to more stable fairness outcomes.

\noindent\textbf{Efficiency analysis.}
Figure~\ref{fig:Efficiency} reports the runtime of the data expansion framework, excluding all pre-computation costs for all methods. All the baselines typically consume the entire expansion budget, evaluating a large number of candidate batches. Since each additional batch evaluation requires retraining the model, this results in substantially higher computational cost. In contrast, \sys incurs a modest overhead for reward updates and influence-based ranking, but achieves significantly lower overall runtime by terminating early once the fairness objective is satisfied.

Figure~\ref{fig: Synthetic Dataset Efficiency} evaluates scalability under increasing data pool sizes using a synthetically enlarged AdultIncome dataset. For \sys, batch construction cost remains nearly constant since influence scores are computed once on the fixed training set and the top-$K$ influential points are selected deterministically. As a result, runtime is largely insensitive to pool size. In contrast, \sysmab relies on random batch construction, making its runtime grow with the size of the data pool.

\section{Related Work}
\label{sec:related}
This work intersects four research areas: algorithmic fairness, data expansion/acquisition, reinforcement learning, and data valuation. While each of these areas has been studied extensively, our work is novel in integrating data valuation with a multi-armed bandit framework to selectively acquire high-quality data for improving model fairness.

\vspace{1mm}\noindent\textbf{Algorithmic fairness.}
With the growing use of ML in high-stakes domains such as criminal justice, healthcare, and finance, fairness violations have become increasingly evident~\cite{fb-housing,amazonhire2018,10.1145/2702123.2702520,self-driving-cars}. Bias mitigation techniques are typically categorized into pre-, in-, and post-processing methods~\cite{mehrabi_survey_2022,10.1145/3616865}. Due to their effectiveness and ease of implementation, pre-processing approaches are widely adopted~\cite{10.1145/3290605.3300830}. Since our method modifies input data, it is most closely related to pre-processing bias mitigation.

\vspace{1mm}\noindent\textbf{Data expansion/acquisition for model fairness.}
Data quality has long been central to data management~\cite{weikum2013data,rekatsinas2016sourcesight,tae2021slice}, and its importance for ML performance has gained increasing attention~\cite{li_data_2021,10.1145/3654934,wang2024optimizing,chai_selective_2022,asudeh_assessing_2019}. We focus on \emph{data expansion}~\cite{NEURIPS2023_0506ad3d}, selectively acquiring high-quality examples. Unlike prior work targeting accuracy or vision-based augmentation~\cite{shorten2019survey}, our objective is fairness. The closest work, AutoData~\cite{chai_selective_2022}, uses bandits for accuracy-driven acquisition but relies on distance-based clustering and random batch selection. In contrast, \sys leverages data valuation for fairness-aware acquisition.

\vspace{1mm}\noindent\textbf{Reinforcement learning in data management.}
Reinforcement learning has been applied to data cleaning, integration, and acquisition~\cite{sutton2018reinforcement,9723570,chai_selective_2022,berti2019learn2clean,10.1145/3394486.3403261}. MAB provides efficient exploration–exploitation trade-offs and near-optimal solutions under limited feedback. However, existing bandit-based acquisition methods remain accuracy-centric and do not optimize fairness.

\vspace{1mm}\noindent\textbf{Data valuation.}
Data valuation methods, including influence functions~\cite{pmlr-v70-koh17a} and data Shapley~\cite{pmlr-v97-ghorbani19c}, quantify the contribution of individual data points and have recently been explored in data management~\cite{bertossi2023shapley,luo2024applications,10.1145/3459637.3482341}. While data Shapley is model-agnostic but computationally expensive, influence functions provide efficient first-order approximations with a one-time offline cost, which we leverage for fairness-aware acquisition.

\section{Conclusions and Future Work}
We present a novel approach for solving the problem of data expansion to improve machine learning model fairness by determining which data points in a data pool must be added to the underlying training data. We introduce \sys, a principled framework that integrates reinforcement learning with data valuation to determine the most valuable data points to incorporate. We demonstrate experimentally that data valuation is crucial in effective discovery of useful data points and that \sys is both effective and efficient in identifying data points valuable for model fairness. 
In the future, we plan to explore several interesting directions: (a)~expand the data valuation module to efficiently cater to non-parametric ML algorithms. (b)~explore alternate reinforcement learning approaches (e.g., Q-learning) and other exploration-exploitation trade-off algorithms (e.g., Thompson sampling) since the MAB-based framework is limited by the choice of identified partitions. Other directions include incorporating the cost of data acquisition in the reward scores and extending the framework to the task of source discovery.

\section*{Acknowledgments}
We would like to thank the anonymous reviewers for their insightful feedback. Funding for this research was provided by the National Science Foundation (NSF) under Grant \#2237149 and the UL Research Institutes through the Center for Advancing Safety of Machine Intelligence.

\section*{Artifacts}
The experiments run on a MacBook Pro (Apple M3 Pro, 36GB LPDDR5) using Python 3.11.5 in Jupyter Notebook. The ready-to-run source code for \sys is available at this link: \href{https://github.com/jahid674/DataSift}{{{\sys}}}. 


\balance
\bibliographystyle{ACM-Reference-Format}
\bibliography{ref}


\begin{thebibliography}{68}


\ifx \showCODEN    \undefined \def \showCODEN     #1{\unskip}     \fi
\ifx \showISBNx    \undefined \def \showISBNx     #1{\unskip}     \fi
\ifx \showISBNxiii \undefined \def \showISBNxiii  #1{\unskip}     \fi
\ifx \showISSN     \undefined \def \showISSN      #1{\unskip}     \fi
\ifx \showLCCN     \undefined \def \showLCCN      #1{\unskip}     \fi
\ifx \shownote     \undefined \def \shownote      #1{#1}          \fi
\ifx \showarticletitle \undefined \def \showarticletitle #1{#1}   \fi
\ifx \showURL      \undefined \def \showURL       {\relax}        \fi
\providecommand\bibfield[2]{#2}
\providecommand\bibinfo[2]{#2}
\providecommand\natexlab[1]{#1}
\providecommand\showeprint[2][]{arXiv:#2}

\bibitem[Agrawal and Goyal(2012)]%
        {agrawal2012analysis}
\bibfield{author}{\bibinfo{person}{Shipra Agrawal} {and} \bibinfo{person}{Navin Goyal}.} \bibinfo{year}{2012}\natexlab{}.
\newblock \showarticletitle{Analysis of Thompson sampling for the multi-armed bandit problem}. In \bibinfo{booktitle}{\emph{Conference on Learning Theory}}. \bibinfo{publisher}{JMLR Workshop and Conference Proceedings}, \bibinfo{address}{Edinburgh, Scotland}, \bibinfo{pages}{39--1}.
\newblock


\bibitem[Angwin et~al\mbox{.}(2016)]%
        {propublica-machine-bias}
\bibfield{author}{\bibinfo{person}{Julia Angwin}, \bibinfo{person}{Jeff Larson}, \bibinfo{person}{Surya Mattu}, {and} \bibinfo{person}{Lauren Kirchner}.} \bibinfo{year}{2016}\natexlab{}.
\newblock \bibinfo{booktitle}{\emph{Machine Bias: There’s Software Used Across the Country to Predict Future Criminals. And It’s Biased Against Blacks.}}
\newblock
\newblock
\shownote{ProPublica, May 23, 2016}.


\bibitem[Asudeh et~al\mbox{.}(2019)]%
        {asudeh_assessing_2019}
\bibfield{author}{\bibinfo{person}{Abolfazl Asudeh}, \bibinfo{person}{Zhongjun Jin}, {and} \bibinfo{person}{H.~V. Jagadish}.} \bibinfo{year}{2019}\natexlab{}.
\newblock \showarticletitle{Assessing and {Remedying} {Coverage} for a {Given} {Dataset}}. In \bibinfo{booktitle}{\emph{2019 {IEEE} 35th {International} {Conference} on {Data} {Engineering} ({ICDE})}}. \bibinfo{publisher}{IEEE}, \bibinfo{address}{Macau, China}, \bibinfo{pages}{554--565}.
\newblock
\href{https://doi.org/10.1109/ICDE.2019.00056}{doi:\nolinkurl{10.1109/ICDE.2019.00056}}
\newblock
\shownote{ISSN: 2375-026X}.


\bibitem[Auer(2000)]%
        {auer2000using}
\bibfield{author}{\bibinfo{person}{Peter Auer}.} \bibinfo{year}{2000}\natexlab{}.
\newblock \showarticletitle{Using upper confidence bounds for online learning}. In \bibinfo{booktitle}{\emph{Proceedings of the 41st Annual Symposium on Foundations of Computer Science (FOCS)}}. IEEE, \bibinfo{publisher}{IEEE}, \bibinfo{address}{Redondo Beach, CA, USA}, \bibinfo{pages}{270--279}.
\newblock


\bibitem[Auer(2002)]%
        {auer2002using}
\bibfield{author}{\bibinfo{person}{Peter Auer}.} \bibinfo{year}{2002}\natexlab{}.
\newblock \showarticletitle{Using confidence bounds for exploitation-exploration trade-offs}.
\newblock \bibinfo{journal}{\emph{Journal of Machine Learning Research}} \bibinfo{volume}{3}, \bibinfo{number}{Nov} (\bibinfo{year}{2002}), \bibinfo{pages}{397--422}.
\newblock


\bibitem[Berti-Equille(2019)]%
        {berti2019learn2clean}
\bibfield{author}{\bibinfo{person}{Laure Berti-Equille}.} \bibinfo{year}{2019}\natexlab{}.
\newblock \showarticletitle{Learn2clean: Optimizing the sequence of tasks for web data preparation}. In \bibinfo{booktitle}{\emph{The world wide web conference}}. \bibinfo{pages}{2580--2586}.
\newblock


\bibitem[Bertossi et~al\mbox{.}(2023)]%
        {bertossi2023shapley}
\bibfield{author}{\bibinfo{person}{Leopoldo Bertossi}, \bibinfo{person}{Benny Kimelfeld}, \bibinfo{person}{Ester Livshits}, {and} \bibinfo{person}{Mika{\"e}l Monet}.} \bibinfo{year}{2023}\natexlab{}.
\newblock \showarticletitle{The Shapley value in database management}.
\newblock \bibinfo{journal}{\emph{ACM Sigmod Record}} \bibinfo{volume}{52}, \bibinfo{number}{2} (\bibinfo{year}{2023}), \bibinfo{pages}{6--17}.
\newblock


\bibitem[Biswas and Rajan(2021)]%
        {10.1145/3468264.3468536}
\bibfield{author}{\bibinfo{person}{Sumon Biswas} {and} \bibinfo{person}{Hridesh Rajan}.} \bibinfo{year}{2021}\natexlab{}.
\newblock \showarticletitle{Fair preprocessing: towards understanding compositional fairness of data transformers in machine learning pipeline}. In \bibinfo{booktitle}{\emph{Proceedings of the 29th ACM Joint Meeting on European Software Engineering Conference and Symposium on the Foundations of Software Engineering}}. \bibinfo{pages}{981–993}.
\newblock
\showISBNx{9781450385626}


\bibitem[Biswas et~al\mbox{.}(2022)]%
        {10.1145/3510003.3510057}
\bibfield{author}{\bibinfo{person}{Sumon Biswas}, \bibinfo{person}{Mohammad Wardat}, {and} \bibinfo{person}{Hridesh Rajan}.} \bibinfo{year}{2022}\natexlab{}.
\newblock \showarticletitle{The art and practice of data science pipelines: A comprehensive study of data science pipelines in theory, in-the-small, and in-the-large}. In \bibinfo{booktitle}{\emph{Proceedings of the 44th International Conference on Software Engineering}}. \bibinfo{pages}{2091–2103}.
\newblock
\showISBNx{9781450392211}


\bibitem[Booker(2019)]%
        {fb-housing}
\bibfield{author}{\bibinfo{person}{Brakkton Booker}.} \bibinfo{year}{2019}\natexlab{}.
\newblock \bibinfo{title}{Housing Department Slaps Facebook With Discrimination Charge. \url{https://www.npr.org/2019/03/28/707614254/hud-slaps-facebook-with-housing-discrimination-charge}}.
\newblock


\bibitem[Bubeck et~al\mbox{.}(2013)]%
        {bubeck2013bandits}
\bibfield{author}{\bibinfo{person}{S{\'e}bastien Bubeck}, \bibinfo{person}{Nicolo Cesa-Bianchi}, {and} \bibinfo{person}{G{\'a}bor Lugosi}.} \bibinfo{year}{2013}\natexlab{}.
\newblock \showarticletitle{Bandits with heavy tail}.
\newblock \bibinfo{journal}{\emph{IEEE Transactions on Information Theory}} \bibinfo{volume}{59}, \bibinfo{number}{11} (\bibinfo{year}{2013}), \bibinfo{pages}{7711--7717}.
\newblock


\bibitem[Cai et~al\mbox{.}(2023)]%
        {9723570}
\bibfield{author}{\bibinfo{person}{Qingpeng Cai}, \bibinfo{person}{Can Cui}, \bibinfo{person}{Yiyuan Xiong}, \bibinfo{person}{Wei Wang}, \bibinfo{person}{Zhongle Xie}, {and} \bibinfo{person}{Meihui Zhang}.} \bibinfo{year}{2023}\natexlab{}.
\newblock \showarticletitle{A Survey on Deep Reinforcement Learning for Data Processing and Analytics}.
\newblock \bibinfo{journal}{\emph{IEEE Transactions on Knowledge and Data Engineering}} \bibinfo{volume}{35}, \bibinfo{number}{5} (\bibinfo{year}{2023}), \bibinfo{pages}{4446--4465}.
\newblock
\href{https://doi.org/10.1109/TKDE.2022.3155196}{doi:\nolinkurl{10.1109/TKDE.2022.3155196}}


\bibitem[Carpentier et~al\mbox{.}(2011)]%
        {carpentier2011upper}
\bibfield{author}{\bibinfo{person}{Alexandra Carpentier}, \bibinfo{person}{Alessandro Lazaric}, \bibinfo{person}{Mohammad Ghavamzadeh}, \bibinfo{person}{R{\'e}mi Munos}, {and} \bibinfo{person}{Peter Auer}.} \bibinfo{year}{2011}\natexlab{}.
\newblock \showarticletitle{Upper-confidence-bound algorithms for active learning in multi-armed bandits}. In \bibinfo{booktitle}{\emph{International Conference on Algorithmic Learning Theory}}. Springer, \bibinfo{pages}{189--203}.
\newblock


\bibitem[Caton and Haas(2024)]%
        {10.1145/3616865}
\bibfield{author}{\bibinfo{person}{Simon Caton} {and} \bibinfo{person}{Christian Haas}.} \bibinfo{year}{2024}\natexlab{}.
\newblock \showarticletitle{Fairness in Machine Learning: A Survey}.
\newblock \bibinfo{journal}{\emph{ACM Comput. Surv.}} \bibinfo{volume}{56}, \bibinfo{number}{7}, Article \bibinfo{articleno}{166} (\bibinfo{date}{April} \bibinfo{year}{2024}), \bibinfo{numpages}{38}~pages.
\newblock
\showISSN{0360-0300}
\href{https://doi.org/10.1145/3616865}{doi:\nolinkurl{10.1145/3616865}}


\bibitem[Chai et~al\mbox{.}(2022)]%
        {chai_selective_2022}
\bibfield{author}{\bibinfo{person}{Chengliang Chai}, \bibinfo{person}{Jiabin Liu}, \bibinfo{person}{Nan Tang}, \bibinfo{person}{Guoliang Li}, {and} \bibinfo{person}{Yuyu Luo}.} \bibinfo{year}{2022}\natexlab{}.
\newblock \showarticletitle{Selective data acquisition in the wild for model charging}.
\newblock \bibinfo{journal}{\emph{Proc. VLDB Endow.}} \bibinfo{volume}{15}, \bibinfo{number}{7} (\bibinfo{date}{March} \bibinfo{year}{2022}), \bibinfo{pages}{1466--1478}.
\newblock
\showISSN{2150-8097}
\href{https://doi.org/10.14778/3523210.3523223}{doi:\nolinkurl{10.14778/3523210.3523223}}


\bibitem[Chouldechova(2017)]%
        {chouldechova2017fair}
\bibfield{author}{\bibinfo{person}{Alexandra Chouldechova}.} \bibinfo{year}{2017}\natexlab{}.
\newblock \showarticletitle{Fair prediction with disparate impact: A study of bias in recidivism prediction instruments}.
\newblock \bibinfo{journal}{\emph{CoRR}}  \bibinfo{volume}{abs/1703.00056} (\bibinfo{year}{2017}).
\newblock


\bibitem[Cook and Weisberg(1980)]%
        {doi:10.1080/00401706.1980.10486199}
\bibfield{author}{\bibinfo{person}{R.~Dennis Cook} {and} \bibinfo{person}{Sanford Weisberg}.} \bibinfo{year}{1980}\natexlab{}.
\newblock \showarticletitle{Characterizations of an Empirical Influence Function for Detecting Influential Cases in Regression}.
\newblock \bibinfo{journal}{\emph{Technometrics}} \bibinfo{volume}{22}, \bibinfo{number}{4} (\bibinfo{year}{1980}), \bibinfo{pages}{495--508}.
\newblock


\bibitem[Credit~Fusion(2011)]%
        {GiveMeSomeCredit}
\bibfield{author}{\bibinfo{person}{Will~Cukierski Credit~Fusion}.} \bibinfo{year}{2011}\natexlab{}.
\newblock \bibinfo{title}{Give Me Some Credit}.
\newblock
\urldef\tempurl%
\url{https://kaggle.com/competitions/GiveMeSomeCredit}
\showURL{%
\tempurl}


\bibitem[Dastin(2018)]%
        {amazonhire2018}
\bibfield{author}{\bibinfo{person}{Jeffrey Dastin}.} \bibinfo{year}{2018}\natexlab{}.
\newblock \showarticletitle{RPT-INSIGHT-Amazon scraps secret AI recruiting tool that showed bias against women}.
\newblock \bibinfo{journal}{\emph{Reuters}} (\bibinfo{year}{2018}).
\newblock


\bibitem[Davis et~al\mbox{.}(1997)]%
        {886461319970101}
\bibfield{author}{\bibinfo{person}{G. Davis}, \bibinfo{person}{S. Mallat}, {and} \bibinfo{person}{M. Avellaneda}.} \bibinfo{year}{1997}\natexlab{}.
\newblock \showarticletitle{Adaptive Greedy Approximations.}
\newblock \bibinfo{journal}{\emph{Constructive Approximation}} \bibinfo{volume}{13}, \bibinfo{number}{1} (\bibinfo{year}{1997}), \bibinfo{pages}{57}.
\newblock
\showISSN{01764276}
\urldef\tempurl%
\url{https://search.ebscohost.com/login.aspx?direct=true&amp;AuthType=sso&amp;db=a9h&amp;AN=8864613&amp;site=ehost-live&amp;custid=purdue}
\showURL{%
\tempurl}


\bibitem[Deng et~al\mbox{.}(2023)]%
        {NEURIPS2023_0506ad3d}
\bibfield{author}{\bibinfo{person}{Yihe Deng}, \bibinfo{person}{Yu Yang}, \bibinfo{person}{Baharan Mirzasoleiman}, {and} \bibinfo{person}{Quanquan Gu}.} \bibinfo{year}{2023}\natexlab{}.
\newblock \showarticletitle{Robust Learning with Progressive Data Expansion Against Spurious Correlation}. In \bibinfo{booktitle}{\emph{Advances in Neural Information Processing Systems}}, \bibfield{editor}{\bibinfo{person}{A.~Oh}, \bibinfo{person}{T.~Naumann}, \bibinfo{person}{A.~Globerson}, \bibinfo{person}{K.~Saenko}, \bibinfo{person}{M.~Hardt}, {and} \bibinfo{person}{S.~Levine}} (Eds.), Vol.~\bibinfo{volume}{36}. \bibinfo{publisher}{Curran Associates, Inc.}, \bibinfo{pages}{1390--1402}.
\newblock
\urldef\tempurl%
\url{https://proceedings.neurips.cc/paper_files/paper/2023/file/0506ad3d1bcc8398a920db9340f27fe4-Paper-Conference.pdf}
\showURL{%
\tempurl}


\bibitem[Deutch et~al\mbox{.}(2021)]%
        {10.1145/3459637.3482341}
\bibfield{author}{\bibinfo{person}{Daniel Deutch}, \bibinfo{person}{Nave Frost}, \bibinfo{person}{Amir Gilad}, {and} \bibinfo{person}{Oren Sheffer}.} \bibinfo{year}{2021}\natexlab{}.
\newblock \showarticletitle{Explanations for Data Repair Through Shapley Values}. In \bibinfo{booktitle}{\emph{Proceedings of the 30th ACM International Conference on Information \& Knowledge Management}} (Virtual Event, Queensland, Australia) \emph{(\bibinfo{series}{CIKM '21})}. \bibinfo{publisher}{Association for Computing Machinery}, \bibinfo{address}{New York, NY, USA}, \bibinfo{pages}{362–371}.
\newblock
\showISBNx{9781450384469}
\href{https://doi.org/10.1145/3459637.3482341}{doi:\nolinkurl{10.1145/3459637.3482341}}


\bibitem[Ding et~al\mbox{.}(2021)]%
        {ding2021retiring}
\bibfield{author}{\bibinfo{person}{Frances Ding}, \bibinfo{person}{Moritz Hardt}, \bibinfo{person}{John Miller}, {and} \bibinfo{person}{Ludwig Schmidt}.} \bibinfo{year}{2021}\natexlab{}.
\newblock \showarticletitle{Retiring adult: New datasets for fair machine learning}.
\newblock \bibinfo{journal}{\emph{Advances in neural information processing systems}}  \bibinfo{volume}{34} (\bibinfo{year}{2021}), \bibinfo{pages}{6478--6490}.
\newblock


\bibitem[Dua et~al\mbox{.}(2017)]%
        {dua2017uci}
\bibfield{author}{\bibinfo{person}{Dheeru Dua}, \bibinfo{person}{Casey Graff}, {et~al\mbox{.}}} \bibinfo{year}{2017}\natexlab{}.
\newblock \showarticletitle{UCI machine learning repository}.
\newblock  (\bibinfo{year}{2017}).
\newblock


\bibitem[Ester et~al\mbox{.}(1996)]%
        {ester1996density}
\bibfield{author}{\bibinfo{person}{Martin Ester}, \bibinfo{person}{Hans-Peter Kriegel}, \bibinfo{person}{J{\"o}rg Sander}, \bibinfo{person}{Xiaowei Xu}, {et~al\mbox{.}}} \bibinfo{year}{1996}\natexlab{}.
\newblock \showarticletitle{A density-based algorithm for discovering clusters in large spatial databases with noise}. In \bibinfo{booktitle}{\emph{kdd}}, Vol.~\bibinfo{volume}{96}. \bibinfo{pages}{226--231}.
\newblock


\bibitem[Fernandez et~al\mbox{.}(2018)]%
        {fernandez2018aurum}
\bibfield{author}{\bibinfo{person}{Raul~Castro Fernandez}, \bibinfo{person}{Ziawasch Abedjan}, \bibinfo{person}{Famien Koko}, \bibinfo{person}{Gina Yuan}, \bibinfo{person}{Samuel Madden}, {and} \bibinfo{person}{Michael Stonebraker}.} \bibinfo{year}{2018}\natexlab{}.
\newblock \showarticletitle{Aurum: A data discovery system}. In \bibinfo{booktitle}{\emph{2018 IEEE 34th International Conference on Data Engineering (ICDE)}}. IEEE, \bibinfo{pages}{1001--1012}.
\newblock


\bibitem[Figueiredo and Jain(2002)]%
        {figueiredo2002unsupervised}
\bibfield{author}{\bibinfo{person}{Mario A.~T. Figueiredo} {and} \bibinfo{person}{Anil~K. Jain}.} \bibinfo{year}{2002}\natexlab{}.
\newblock \showarticletitle{Unsupervised learning of finite mixture models}.
\newblock \bibinfo{journal}{\emph{IEEE Transactions on pattern analysis and machine intelligence}} \bibinfo{volume}{24}, \bibinfo{number}{3} (\bibinfo{year}{2002}), \bibinfo{pages}{381--396}.
\newblock


\bibitem[Garivier and Moulines(2011)]%
        {garivier2011upper}
\bibfield{author}{\bibinfo{person}{Aur{\'e}lien Garivier} {and} \bibinfo{person}{Eric Moulines}.} \bibinfo{year}{2011}\natexlab{}.
\newblock \showarticletitle{On upper-confidence bound policies for switching bandit problems}. In \bibinfo{booktitle}{\emph{International conference on algorithmic learning theory}}. Springer, \bibinfo{pages}{174--188}.
\newblock


\bibitem[Ghorbani and Zou(2019)]%
        {pmlr-v97-ghorbani19c}
\bibfield{author}{\bibinfo{person}{Amirata Ghorbani} {and} \bibinfo{person}{James Zou}.} \bibinfo{year}{2019}\natexlab{}.
\newblock \showarticletitle{Data Shapley: Equitable Valuation of Data for Machine Learning}. In \bibinfo{booktitle}{\emph{Proceedings of the 36th International Conference on Machine Learning}}, Vol.~\bibinfo{volume}{97}. \bibinfo{publisher}{PMLR}, \bibinfo{pages}{2242--2251}.
\newblock


\bibitem[Hao(2019)]%
        {self-driving-cars}
\bibfield{author}{\bibinfo{person}{Karen Hao}.} \bibinfo{year}{2019}\natexlab{}.
\newblock \bibinfo{title}{Self-driving cars more likely to hit blacks. \url{https://www.technologyreview.com/2019/03/01/136808/self-driving-cars-are-coming-but-accidents-may-not-be-evenly-distributed/}}.
\newblock


\bibitem[Heffetz et~al\mbox{.}(2020)]%
        {10.1145/3394486.3403261}
\bibfield{author}{\bibinfo{person}{Yuval Heffetz}, \bibinfo{person}{Roman Vainshtein}, \bibinfo{person}{Gilad Katz}, {and} \bibinfo{person}{Lior Rokach}.} \bibinfo{year}{2020}\natexlab{}.
\newblock \showarticletitle{DeepLine: AutoML Tool for Pipelines Generation using Deep Reinforcement Learning and Hierarchical Actions Filtering}. In \bibinfo{booktitle}{\emph{Proceedings of the 26th ACM SIGKDD International Conference on Knowledge Discovery \& Data Mining}} (Virtual Event, CA, USA) \emph{(\bibinfo{series}{KDD '20})}. \bibinfo{publisher}{Association for Computing Machinery}, \bibinfo{address}{New York, NY, USA}, \bibinfo{pages}{2103–2113}.
\newblock
\showISBNx{9781450379984}
\href{https://doi.org/10.1145/3394486.3403261}{doi:\nolinkurl{10.1145/3394486.3403261}}


\bibitem[Holstein et~al\mbox{.}(2019)]%
        {10.1145/3290605.3300830}
\bibfield{author}{\bibinfo{person}{Kenneth Holstein}, \bibinfo{person}{Jennifer Wortman~Vaughan}, \bibinfo{person}{Hal Daum\'{e}}, \bibinfo{person}{Miro Dudik}, {and} \bibinfo{person}{Hanna Wallach}.} \bibinfo{year}{2019}\natexlab{}.
\newblock \showarticletitle{Improving Fairness in Machine Learning Systems: What Do Industry Practitioners Need?}. In \bibinfo{booktitle}{\emph{Proceedings of the 2019 CHI Conference on Human Factors in Computing Systems}}. \bibinfo{pages}{1–16}.
\newblock
\showISBNx{9781450359702}


\bibitem[Jain(2010)]%
        {jain2010data}
\bibfield{author}{\bibinfo{person}{Anil~K Jain}.} \bibinfo{year}{2010}\natexlab{}.
\newblock \showarticletitle{Data clustering: 50 years beyond K-means}.
\newblock \bibinfo{journal}{\emph{Pattern recognition letters}} \bibinfo{volume}{31}, \bibinfo{number}{8} (\bibinfo{year}{2010}), \bibinfo{pages}{651--666}.
\newblock


\bibitem[Kay et~al\mbox{.}(2015)]%
        {10.1145/2702123.2702520}
\bibfield{author}{\bibinfo{person}{Matthew Kay}, \bibinfo{person}{Cynthia Matuszek}, {and} \bibinfo{person}{Sean~A. Munson}.} \bibinfo{year}{2015}\natexlab{}.
\newblock \showarticletitle{Unequal Representation and Gender Stereotypes in Image Search Results for Occupations}. In \bibinfo{booktitle}{\emph{Proceedings of the 33rd Annual ACM Conference on Human Factors in Computing Systems}}. \bibinfo{pages}{3819–3828}.
\newblock
\showISBNx{9781450331456}


\bibitem[Koh and Liang(2017)]%
        {pmlr-v70-koh17a}
\bibfield{author}{\bibinfo{person}{Pang~Wei Koh} {and} \bibinfo{person}{Percy Liang}.} \bibinfo{year}{2017}\natexlab{}.
\newblock \showarticletitle{Understanding Black-box Predictions via Influence Functions}. In \bibinfo{booktitle}{\emph{Proceedings of the 34th International Conference on Machine Learning}}, Vol.~\bibinfo{volume}{70}. \bibinfo{pages}{1885--1894}.
\newblock


\bibitem[Kuleshov and Precup(2014)]%
        {kuleshov2014algorithms}
\bibfield{author}{\bibinfo{person}{Volodymyr Kuleshov} {and} \bibinfo{person}{Doina Precup}.} \bibinfo{year}{2014}\natexlab{}.
\newblock \showarticletitle{Algorithms for multi-armed bandit problems}.
\newblock \bibinfo{journal}{\emph{arXiv preprint arXiv:1402.6028}} (\bibinfo{year}{2014}).
\newblock


\bibitem[Li et~al\mbox{.}(2021)]%
        {li_data_2021}
\bibfield{author}{\bibinfo{person}{Yifan Li}, \bibinfo{person}{Xiaohui Yu}, {and} \bibinfo{person}{Nick Koudas}.} \bibinfo{year}{2021}\natexlab{}.
\newblock \showarticletitle{Data {Acquisition} for {Improving} {Machine} {Learning} {Models}}.
\newblock \bibinfo{journal}{\emph{Proceedings of the VLDB Endowment}} \bibinfo{volume}{14}, \bibinfo{number}{10} (\bibinfo{date}{June} \bibinfo{year}{2021}), \bibinfo{pages}{1832--1844}.
\newblock
\showISSN{2150-8097}
\href{https://doi.org/10.14778/3467861.3467872}{doi:\nolinkurl{10.14778/3467861.3467872}}
\newblock
\shownote{arXiv:2105.14107 [cs]}.


\bibitem[Li et~al\mbox{.}(2024)]%
        {10.1145/3654934}
\bibfield{author}{\bibinfo{person}{Yifan Li}, \bibinfo{person}{Xiaohui Yu}, {and} \bibinfo{person}{Nick Koudas}.} \bibinfo{year}{2024}\natexlab{}.
\newblock \showarticletitle{Data Acquisition for Improving Model Confidence}.
\newblock \bibinfo{journal}{\emph{Proc. ACM Manag. Data}} \bibinfo{volume}{2}, \bibinfo{number}{3}, Article \bibinfo{articleno}{131} (\bibinfo{date}{May} \bibinfo{year}{2024}), \bibinfo{numpages}{25}~pages.
\newblock
\href{https://doi.org/10.1145/3654934}{doi:\nolinkurl{10.1145/3654934}}


\bibitem[Liu et~al\mbox{.}(2021)]%
        {pmlr-v139-liu21f}
\bibfield{author}{\bibinfo{person}{Evan~Z Liu}, \bibinfo{person}{Behzad Haghgoo}, \bibinfo{person}{Annie~S Chen}, \bibinfo{person}{Aditi Raghunathan}, \bibinfo{person}{Pang~Wei Koh}, \bibinfo{person}{Shiori Sagawa}, \bibinfo{person}{Percy Liang}, {and} \bibinfo{person}{Chelsea Finn}.} \bibinfo{year}{2021}\natexlab{}.
\newblock \showarticletitle{Just Train Twice: Improving Group Robustness without Training Group Information}. In \bibinfo{booktitle}{\emph{Proceedings of the 38th International Conference on Machine Learning}} \emph{(\bibinfo{series}{Proceedings of Machine Learning Research}, Vol.~\bibinfo{volume}{139})}, \bibfield{editor}{\bibinfo{person}{Marina Meila} {and} \bibinfo{person}{Tong Zhang}} (Eds.). \bibinfo{publisher}{PMLR}, \bibinfo{pages}{6781--6792}.
\newblock
\urldef\tempurl%
\url{https://proceedings.mlr.press/v139/liu21f.html}
\showURL{%
\tempurl}


\bibitem[Luo and Pei(2024)]%
        {luo2024applications}
\bibfield{author}{\bibinfo{person}{Xuan Luo} {and} \bibinfo{person}{Jian Pei}.} \bibinfo{year}{2024}\natexlab{}.
\newblock \showarticletitle{Applications and computation of the Shapley value in databases and machine learning}. In \bibinfo{booktitle}{\emph{Companion of the 2024 International Conference on Management of Data}}. \bibinfo{pages}{630--635}.
\newblock


\bibitem[Mehrabi et~al\mbox{.}(2022)]%
        {mehrabi_survey_2022}
\bibfield{author}{\bibinfo{person}{Ninareh Mehrabi}, \bibinfo{person}{Fred Morstatter}, \bibinfo{person}{Nripsuta Saxena}, \bibinfo{person}{Kristina Lerman}, {and} \bibinfo{person}{Aram Galstyan}.} \bibinfo{year}{2022}\natexlab{}.
\newblock \showarticletitle{A {Survey} on {Bias} and {Fairness} in {Machine} {Learning}}.
\newblock \bibinfo{journal}{\emph{Comput. Surveys}} \bibinfo{volume}{54}, \bibinfo{number}{6} (\bibinfo{date}{July} \bibinfo{year}{2022}), \bibinfo{pages}{1--35}.
\newblock
\showISSN{0360-0300, 1557-7341}
\href{https://doi.org/10.1145/3457607}{doi:\nolinkurl{10.1145/3457607}}


\bibitem[Mishra et~al\mbox{.}(2021)]%
        {mishra2021accelerating}
\bibfield{author}{\bibinfo{person}{Asit Mishra}, \bibinfo{person}{Jorge~Albericio Latorre}, \bibinfo{person}{Jeff Pool}, \bibinfo{person}{Darko Stosic}, \bibinfo{person}{Dusan Stosic}, \bibinfo{person}{Ganesh Venkatesh}, \bibinfo{person}{Chong Yu}, {and} \bibinfo{person}{Paulius Micikevicius}.} \bibinfo{year}{2021}\natexlab{}.
\newblock \showarticletitle{Accelerating sparse deep neural networks}.
\newblock \bibinfo{journal}{\emph{arXiv preprint arXiv:2104.08378}} (\bibinfo{year}{2021}).
\newblock


\bibitem[Murtagh and Contreras(2012)]%
        {murtagh2012algorithms}
\bibfield{author}{\bibinfo{person}{Fionn Murtagh} {and} \bibinfo{person}{Pedro Contreras}.} \bibinfo{year}{2012}\natexlab{}.
\newblock \showarticletitle{Algorithms for hierarchical clustering: an overview}.
\newblock \bibinfo{journal}{\emph{Wiley interdisciplinary reviews: data mining and knowledge discovery}} \bibinfo{volume}{2}, \bibinfo{number}{1} (\bibinfo{year}{2012}), \bibinfo{pages}{86--97}.
\newblock


\bibitem[Nargesian et~al\mbox{.}(2021)]%
        {nargesian2021tailoring}
\bibfield{author}{\bibinfo{person}{Fatemeh Nargesian}, \bibinfo{person}{Abolfazl Asudeh}, {and} \bibinfo{person}{HV Jagadish}.} \bibinfo{year}{2021}\natexlab{}.
\newblock \showarticletitle{Tailoring data source distributions for fairness-aware data integration}.
\newblock \bibinfo{journal}{\emph{Proceedings of the VLDB Endowment}} \bibinfo{volume}{14}, \bibinfo{number}{11} (\bibinfo{year}{2021}), \bibinfo{pages}{2519--2532}.
\newblock


\bibitem[Nargesian et~al\mbox{.}(2020)]%
        {nargesian2020organizing}
\bibfield{author}{\bibinfo{person}{Fatemeh Nargesian}, \bibinfo{person}{Ken~Q Pu}, \bibinfo{person}{Erkang Zhu}, \bibinfo{person}{Bahar Ghadiri~Bashardoost}, {and} \bibinfo{person}{Ren{\'e}e~J Miller}.} \bibinfo{year}{2020}\natexlab{}.
\newblock \showarticletitle{Organizing data lakes for navigation}. In \bibinfo{booktitle}{\emph{Proceedings of the 2020 ACM SIGMOD International Conference on Management of Data}}. \bibinfo{pages}{1939--1950}.
\newblock


\bibitem[Nargesian et~al\mbox{.}(2018)]%
        {nargesian2018table}
\bibfield{author}{\bibinfo{person}{Fatemeh Nargesian}, \bibinfo{person}{Erkang Zhu}, \bibinfo{person}{Ken~Q Pu}, {and} \bibinfo{person}{Ren{\'e}e~J Miller}.} \bibinfo{year}{2018}\natexlab{}.
\newblock \showarticletitle{Table union search on open data}.
\newblock \bibinfo{journal}{\emph{Proceedings of the VLDB Endowment}} \bibinfo{volume}{11}, \bibinfo{number}{7} (\bibinfo{year}{2018}), \bibinfo{pages}{813--825}.
\newblock


\bibitem[Neath and Cavanaugh(2012)]%
        {neath2012bayesian}
\bibfield{author}{\bibinfo{person}{Andrew~A Neath} {and} \bibinfo{person}{Joseph~E Cavanaugh}.} \bibinfo{year}{2012}\natexlab{}.
\newblock \showarticletitle{The Bayesian information criterion: background, derivation, and applications}.
\newblock \bibinfo{journal}{\emph{Wiley Interdisciplinary Reviews: Computational Statistics}} \bibinfo{volume}{4}, \bibinfo{number}{2} (\bibinfo{year}{2012}), \bibinfo{pages}{199--203}.
\newblock


\bibitem[Nguyen-Thanh et~al\mbox{.}(2019)]%
        {nguyen2019recommendation}
\bibfield{author}{\bibinfo{person}{Nhan Nguyen-Thanh}, \bibinfo{person}{Dana Marinca}, \bibinfo{person}{Kinda Khawam}, \bibinfo{person}{David Rohde}, \bibinfo{person}{Flavian Vasile}, \bibinfo{person}{Elena~Simona Lohan}, \bibinfo{person}{Steven Martin}, {and} \bibinfo{person}{Dominique Quadri}.} \bibinfo{year}{2019}\natexlab{}.
\newblock \showarticletitle{Recommendation system-based upper confidence bound for online advertising}.
\newblock \bibinfo{journal}{\emph{arXiv preprint arXiv:1909.04190}} (\bibinfo{year}{2019}).
\newblock


\bibitem[Paszke et~al\mbox{.}(2019)]%
        {paszke2019pytorch}
\bibfield{author}{\bibinfo{person}{Adam Paszke}, \bibinfo{person}{Sam Gross}, \bibinfo{person}{Francisco Massa}, \bibinfo{person}{Adam Lerer}, \bibinfo{person}{James Bradbury}, \bibinfo{person}{Gregory Chanan}, \bibinfo{person}{Trevor Killeen}, \bibinfo{person}{Zeming Lin}, \bibinfo{person}{Natalia Gimelshein}, \bibinfo{person}{Luca Antiga}, {et~al\mbox{.}}} \bibinfo{year}{2019}\natexlab{}.
\newblock \showarticletitle{Pytorch: An imperative style, high-performance deep learning library}.
\newblock \bibinfo{journal}{\emph{Advances in neural information processing systems}}  \bibinfo{volume}{32} (\bibinfo{year}{2019}).
\newblock


\bibitem[Pedregosa et~al\mbox{.}(2011)]%
        {pedregosa2011scikit}
\bibfield{author}{\bibinfo{person}{Fabian Pedregosa}, \bibinfo{person}{Ga{\"e}l Varoquaux}, \bibinfo{person}{Alexandre Gramfort}, \bibinfo{person}{Vincent Michel}, \bibinfo{person}{Bertrand Thirion}, \bibinfo{person}{Olivier Grisel}, \bibinfo{person}{Mathieu Blondel}, \bibinfo{person}{Peter Prettenhofer}, \bibinfo{person}{Ron Weiss}, \bibinfo{person}{Vincent Dubourg}, {et~al\mbox{.}}} \bibinfo{year}{2011}\natexlab{}.
\newblock \showarticletitle{Scikit-learn: Machine learning in Python}.
\newblock \bibinfo{journal}{\emph{the Journal of machine Learning research}}  \bibinfo{volume}{12} (\bibinfo{year}{2011}), \bibinfo{pages}{2825--2830}.
\newblock


\bibitem[Polyzotis and Zaharia(2021)]%
        {DBLP:journals/corr/abs-2112-06439}
\bibfield{author}{\bibinfo{person}{Neoklis Polyzotis} {and} \bibinfo{person}{Matei Zaharia}.} \bibinfo{year}{2021}\natexlab{}.
\newblock \showarticletitle{What can Data-Centric {AI} Learn from Data and {ML} Engineering?}
\newblock \bibinfo{journal}{\emph{CoRR}}  \bibinfo{volume}{abs/2112.06439} (\bibinfo{year}{2021}).
\newblock
\showeprint[arXiv]{2112.06439}
\urldef\tempurl%
\url{https://arxiv.org/abs/2112.06439}
\showURL{%
\tempurl}


\bibitem[Rekatsinas et~al\mbox{.}(2016)]%
        {rekatsinas2016sourcesight}
\bibfield{author}{\bibinfo{person}{Theodoros Rekatsinas}, \bibinfo{person}{Amol Deshpande}, \bibinfo{person}{Xin~Luna Dong}, \bibinfo{person}{Lise Getoor}, {and} \bibinfo{person}{Divesh Srivastava}.} \bibinfo{year}{2016}\natexlab{}.
\newblock \showarticletitle{Sourcesight: Enabling effective source selection}. In \bibinfo{booktitle}{\emph{Proceedings of the 2016 International Conference on Management of Data}}. \bibinfo{pages}{2157--2160}.
\newblock


\bibitem[Shahbazi et~al\mbox{.}(2024)]%
        {shahbazi2024coverage}
\bibfield{author}{\bibinfo{person}{Nima Shahbazi}, \bibinfo{person}{Mahdi Erfanian}, {and} \bibinfo{person}{Abolfazl Asudeh}.} \bibinfo{year}{2024}\natexlab{}.
\newblock \showarticletitle{Coverage-based Data-centric Approaches for Responsible and Trustworthy AI.}
\newblock \bibinfo{journal}{\emph{IEEE Data Engineering Bulletin}} (\bibinfo{year}{2024}).
\newblock


\bibitem[Shahbazi et~al\mbox{.}(2023)]%
        {shahbazi2023representation}
\bibfield{author}{\bibinfo{person}{Nima Shahbazi}, \bibinfo{person}{Yin Lin}, \bibinfo{person}{Abolfazl Asudeh}, {and} \bibinfo{person}{HV Jagadish}.} \bibinfo{year}{2023}\natexlab{}.
\newblock \showarticletitle{Representation bias in data: A survey on identification and resolution techniques}.
\newblock \bibinfo{journal}{\emph{Comput. Surveys}} \bibinfo{volume}{55}, \bibinfo{number}{13s} (\bibinfo{year}{2023}), \bibinfo{pages}{1--39}.
\newblock


\bibitem[Shannon(1948)]%
        {6773024}
\bibfield{author}{\bibinfo{person}{C.~E. Shannon}.} \bibinfo{year}{1948}\natexlab{}.
\newblock \showarticletitle{A mathematical theory of communication}.
\newblock \bibinfo{journal}{\emph{The Bell System Technical Journal}} \bibinfo{volume}{27}, \bibinfo{number}{3} (\bibinfo{year}{1948}), \bibinfo{pages}{379--423}.
\newblock
\href{https://doi.org/10.1002/j.1538-7305.1948.tb01338.x}{doi:\nolinkurl{10.1002/j.1538-7305.1948.tb01338.x}}


\bibitem[Shorten and Khoshgoftaar(2019)]%
        {shorten2019survey}
\bibfield{author}{\bibinfo{person}{Connor Shorten} {and} \bibinfo{person}{Taghi~M Khoshgoftaar}.} \bibinfo{year}{2019}\natexlab{}.
\newblock \showarticletitle{A survey on image data augmentation for deep learning}.
\newblock \bibinfo{journal}{\emph{Journal of big data}} \bibinfo{volume}{6}, \bibinfo{number}{1} (\bibinfo{year}{2019}), \bibinfo{pages}{1--48}.
\newblock


\bibitem[Slivkins et~al\mbox{.}(2019)]%
        {slivkins2019introduction}
\bibfield{author}{\bibinfo{person}{Aleksandrs Slivkins} {et~al\mbox{.}}} \bibinfo{year}{2019}\natexlab{}.
\newblock \showarticletitle{Introduction to multi-armed bandits}.
\newblock \bibinfo{journal}{\emph{Foundations and Trends{\textregistered} in Machine Learning}} \bibinfo{volume}{12}, \bibinfo{number}{1-2} (\bibinfo{year}{2019}), \bibinfo{pages}{1--286}.
\newblock


\bibitem[Sutton and Barto(2018)]%
        {sutton2018reinforcement}
\bibfield{author}{\bibinfo{person}{Richard~S Sutton} {and} \bibinfo{person}{Andrew~G Barto}.} \bibinfo{year}{2018}\natexlab{}.
\newblock \bibinfo{booktitle}{\emph{Reinforcement learning: An introduction}}.
\newblock \bibinfo{publisher}{MIT press}.
\newblock


\bibitem[Tae and Whang(2021a)]%
        {10.1145/3448016.3452792}
\bibfield{author}{\bibinfo{person}{Ki~Hyun Tae} {and} \bibinfo{person}{Steven~Euijong Whang}.} \bibinfo{year}{2021}\natexlab{a}.
\newblock \showarticletitle{Slice Tuner: A Selective Data Acquisition Framework for Accurate and Fair Machine Learning Models}. In \bibinfo{booktitle}{\emph{Proceedings of the 2021 International Conference on Management of Data}} (Virtual Event, China) \emph{(\bibinfo{series}{SIGMOD '21})}. \bibinfo{publisher}{Association for Computing Machinery}, \bibinfo{address}{New York, NY, USA}, \bibinfo{pages}{1771–1783}.
\newblock
\showISBNx{9781450383431}
\href{https://doi.org/10.1145/3448016.3452792}{doi:\nolinkurl{10.1145/3448016.3452792}}


\bibitem[Tae and Whang(2021b)]%
        {tae2021slice}
\bibfield{author}{\bibinfo{person}{Ki~Hyun Tae} {and} \bibinfo{person}{Steven~Euijong Whang}.} \bibinfo{year}{2021}\natexlab{b}.
\newblock \showarticletitle{Slice tuner: A selective data acquisition framework for accurate and fair machine learning models}. In \bibinfo{booktitle}{\emph{Proceedings of the 2021 International Conference on Management of Data}}. \bibinfo{pages}{1771--1783}.
\newblock


\bibitem[Verma and Rubin(2018)]%
        {verma2018fairness}
\bibfield{author}{\bibinfo{person}{Sahil Verma} {and} \bibinfo{person}{Julia Rubin}.} \bibinfo{year}{2018}\natexlab{}.
\newblock \showarticletitle{Fairness definitions explained}. In \bibinfo{booktitle}{\emph{Proceedings of the international workshop on software fairness}}. \bibinfo{publisher}{Association for Computing Machinery}, \bibinfo{pages}{1--7}.
\newblock


\bibitem[Vermorel and Mohri(2005)]%
        {vermorel2005multi}
\bibfield{author}{\bibinfo{person}{Joannes Vermorel} {and} \bibinfo{person}{Mehryar Mohri}.} \bibinfo{year}{2005}\natexlab{}.
\newblock \showarticletitle{Multi-armed bandit algorithms and empirical evaluation}. In \bibinfo{booktitle}{\emph{European conference on machine learning}}. Springer, \bibinfo{pages}{437--448}.
\newblock


\bibitem[Wang et~al\mbox{.}(2024)]%
        {wang2024optimizing}
\bibfield{author}{\bibinfo{person}{Tingting Wang}, \bibinfo{person}{Shixun Huang}, \bibinfo{person}{Zhifeng Bao}, \bibinfo{person}{J~Shane Culpepper}, \bibinfo{person}{Volkan Dedeoglu}, {and} \bibinfo{person}{Reza Arablouei}.} \bibinfo{year}{2024}\natexlab{}.
\newblock \showarticletitle{Optimizing Data Acquisition to Enhance Machine Learning Performance}.
\newblock \bibinfo{journal}{\emph{Proceedings of the VLDB Endowment}} \bibinfo{volume}{17}, \bibinfo{number}{6} (\bibinfo{year}{2024}), \bibinfo{pages}{1310--1323}.
\newblock


\bibitem[Weikum(2013)]%
        {weikum2013data}
\bibfield{author}{\bibinfo{person}{Gerhard Weikum}.} \bibinfo{year}{2013}\natexlab{}.
\newblock \showarticletitle{Data discovery}.
\newblock \bibinfo{journal}{\emph{Data Science Journal}}  \bibinfo{volume}{12} (\bibinfo{year}{2013}).
\newblock


\bibitem[Whang et~al\mbox{.}(2023)]%
        {10.1007/s00778-022-00775-9}
\bibfield{author}{\bibinfo{person}{Steven~Euijong Whang}, \bibinfo{person}{Yuji Roh}, \bibinfo{person}{Hwanjun Song}, {and} \bibinfo{person}{Jae-Gil Lee}.} \bibinfo{year}{2023}\natexlab{}.
\newblock \showarticletitle{Data collection and quality challenges in deep learning: a data-centric AI perspective}.
\newblock \bibinfo{journal}{\emph{The VLDB Journal}} \bibinfo{volume}{32}, \bibinfo{number}{4} (\bibinfo{date}{Jan.} \bibinfo{year}{2023}), \bibinfo{pages}{791–813}.
\newblock
\showISSN{1066-8888}
\href{https://doi.org/10.1007/s00778-022-00775-9}{doi:\nolinkurl{10.1007/s00778-022-00775-9}}


\bibitem[Zha et~al\mbox{.}({[n.\,d.]})]%
        {doi:10.1137/1.9781611977653.ch106}
\bibfield{author}{\bibinfo{person}{Daochen Zha}, \bibinfo{person}{Zaid~Pervaiz Bhat}, \bibinfo{person}{Kwei-Herng Lai}, \bibinfo{person}{Fan Yang}, {and} \bibinfo{person}{Xia Hu}.} \bibinfo{year}{[n.\,d.]}\natexlab{}.
\newblock \bibinfo{booktitle}{\emph{Data-centric AI: Perspectives and Challenges}}.
\newblock \bibinfo{pages}{945--948}.
\newblock
\href{https://doi.org/10.1137/1.9781611977653.ch106}{doi:\nolinkurl{10.1137/1.9781611977653.ch106}}
\showeprint{https://epubs.siam.org/doi/pdf/10.1137/1.9781611977653.ch106}


\bibitem[Zha et~al\mbox{.}(2023)]%
        {10.1145/3580305.3599553}
\bibfield{author}{\bibinfo{person}{Daochen Zha}, \bibinfo{person}{Kwei-Herng Lai}, \bibinfo{person}{Fan Yang}, \bibinfo{person}{Na Zou}, \bibinfo{person}{Huiji Gao}, {and} \bibinfo{person}{Xia Hu}.} \bibinfo{year}{2023}\natexlab{}.
\newblock \showarticletitle{Data-centric AI: Techniques and Future Perspectives}. In \bibinfo{booktitle}{\emph{Proceedings of the 29th ACM SIGKDD Conference on Knowledge Discovery and Data Mining}} (Long Beach, CA, USA) \emph{(\bibinfo{series}{KDD '23})}. \bibinfo{publisher}{Association for Computing Machinery}, \bibinfo{address}{New York, NY, USA}, \bibinfo{pages}{5839–5840}.
\newblock
\showISBNx{9798400701030}
\href{https://doi.org/10.1145/3580305.3599553}{doi:\nolinkurl{10.1145/3580305.3599553}}


\bibitem[Zhang et~al\mbox{.}(1997)]%
        {zhang1997birch}
\bibfield{author}{\bibinfo{person}{Tian Zhang}, \bibinfo{person}{Raghu Ramakrishnan}, {and} \bibinfo{person}{Miron Livny}.} \bibinfo{year}{1997}\natexlab{}.
\newblock \showarticletitle{BIRCH: A new data clustering algorithm and its applications}.
\newblock \bibinfo{journal}{\emph{Data mining and knowledge discovery}} \bibinfo{volume}{1}, \bibinfo{number}{2} (\bibinfo{year}{1997}), \bibinfo{pages}{141--182}.
\newblock


\end{thebibliography}
\appendix
\section{Appendix}
\label{sec:Appendix}


\subsection{Generalizability across fairness metrics}
\label{sec: Appendix_metric}

\begin{figure}[h]
    \centering

    \begin{subfigure}[b]{0.49\columnwidth}
        \centering
        \includegraphics[width=\linewidth]{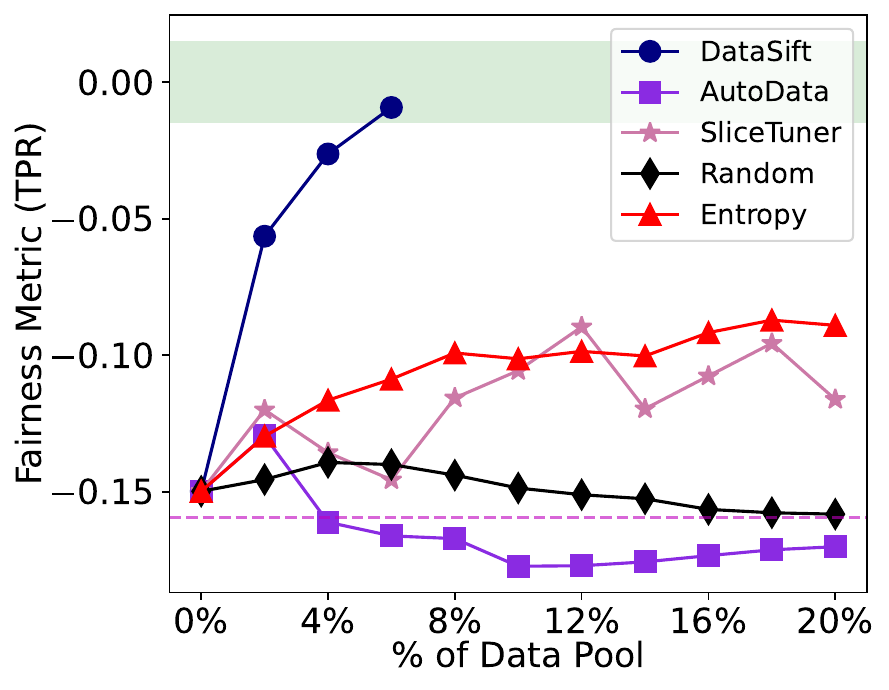}
        \caption{AdultIncome}
    \end{subfigure}\hfill
    \begin{subfigure}[b]{0.49\columnwidth}
        \centering
        \includegraphics[width=\linewidth]{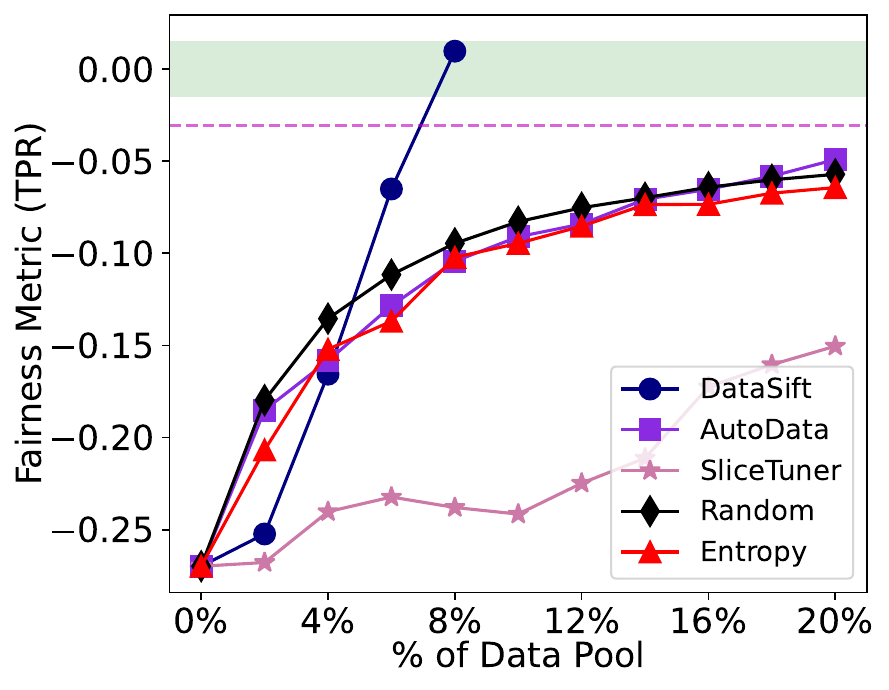}
        \caption{ACSEmployment}
    \end{subfigure}\hfill
    \begin{subfigure}[b]{0.49\columnwidth}
        \centering
        \includegraphics[width=\linewidth]{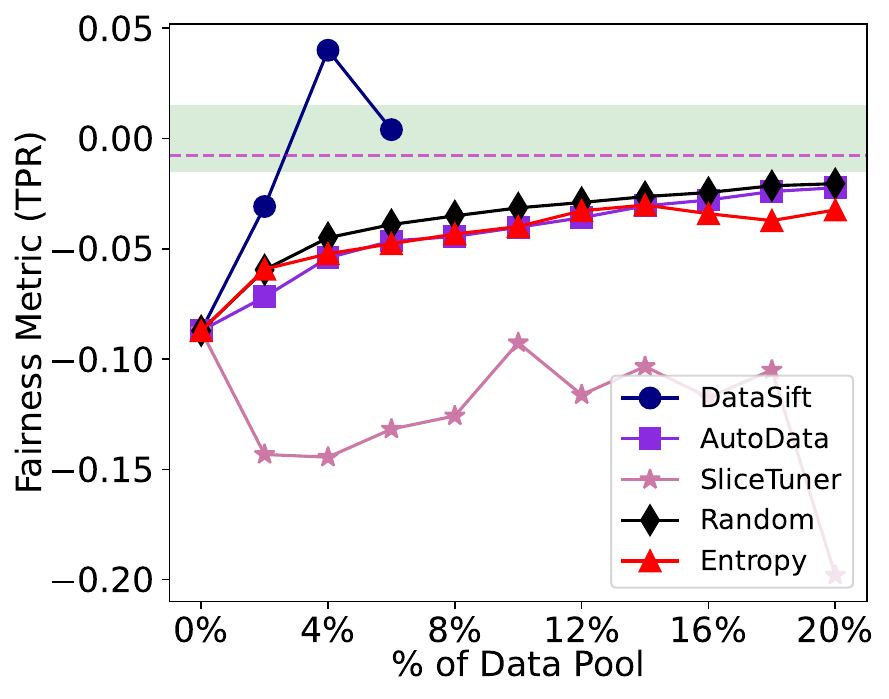}
        \caption{ACSMobility}
    \end{subfigure}\hfill
    \begin{subfigure}[b]{0.49\columnwidth}
        \centering
        \includegraphics[width=\linewidth]{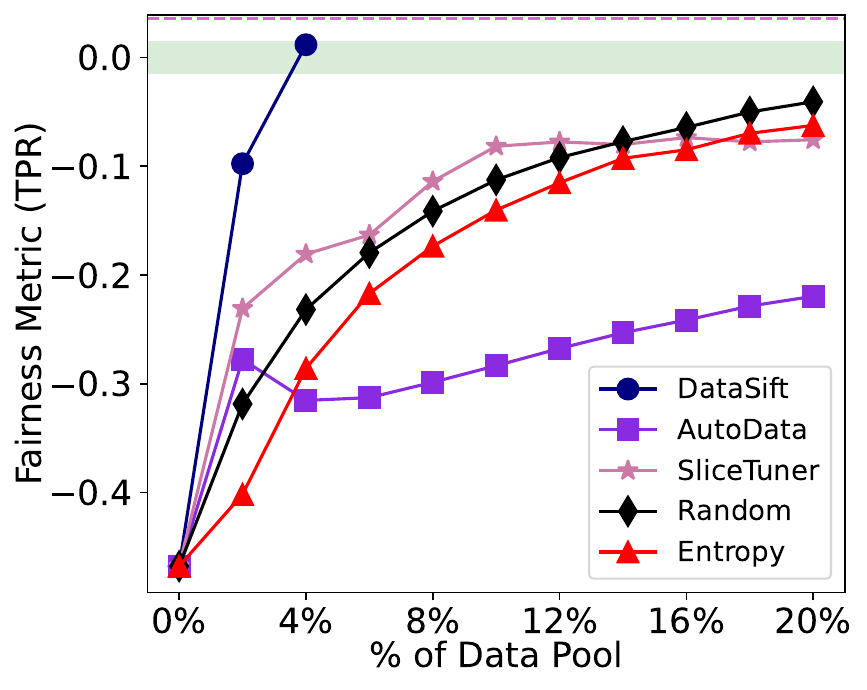}
        \caption{Credit}
    \end{subfigure}

    \caption{\sys vs. baselines (True Positive Rate).}
    \label{fig: DataSift_TPR_extended}
\end{figure}

Algorithmic group fairness requires that individuals belonging to different demographic groups be treated comparably by a decision-making system. Several formal metrics have been proposed to quantify such notions of fairness, including statistical (demographic) parity, true positive rate (TPR) parity, predictive parity, and equalized odds. While the main body of this paper focuses on statistical parity—under which \sys consistently outperforms the baselines—we emphasize that the proposed framework is not limited to a single fairness definition. We report results for the following metrics:

\vspace{1mm}
\noindent \textbf{True Positive Rate (TPR) Parity.}
We first consider \emph{true positive rate parity} (also known as \emph{equal opportunity}), which requires equal true positive rates across groups. Formally, TPR parity is satisfied if
\[
\Pr(\hat{Y} = 1 \mid Y = 1, A = a) = \Pr(\hat{Y} = 1 \mid Y = 1, A = b),
\]
for all protected groups $A \in \{a,b\}$. This metric focuses on ensuring that qualified individuals have equal chances of receiving a positive outcome regardless of group membership. Figure~\ref{fig: DataSift_TPR_extended} reports the corresponding results. \sys outperforms all the baselines across datasets--reach the fairness goal within less number of iterations exploiting only 4 to 8\% of data pool.

\vspace{1mm}

\begin{figure}[b]
    \centering


    \begin{subfigure}[b]{0.48\columnwidth}
        \centering
        \includegraphics[width=\linewidth]{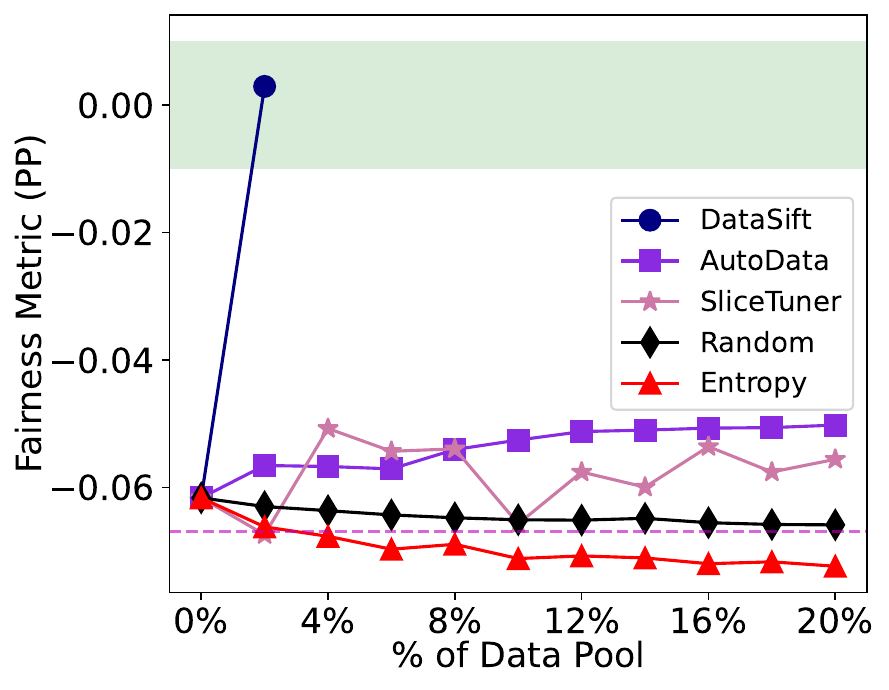}
        \caption{ACSEmployment}
        \label{fig:PP_Employment}
    \end{subfigure}\hfill
    \begin{subfigure}[b]{0.48\columnwidth}
        \centering
        \includegraphics[width=\linewidth]{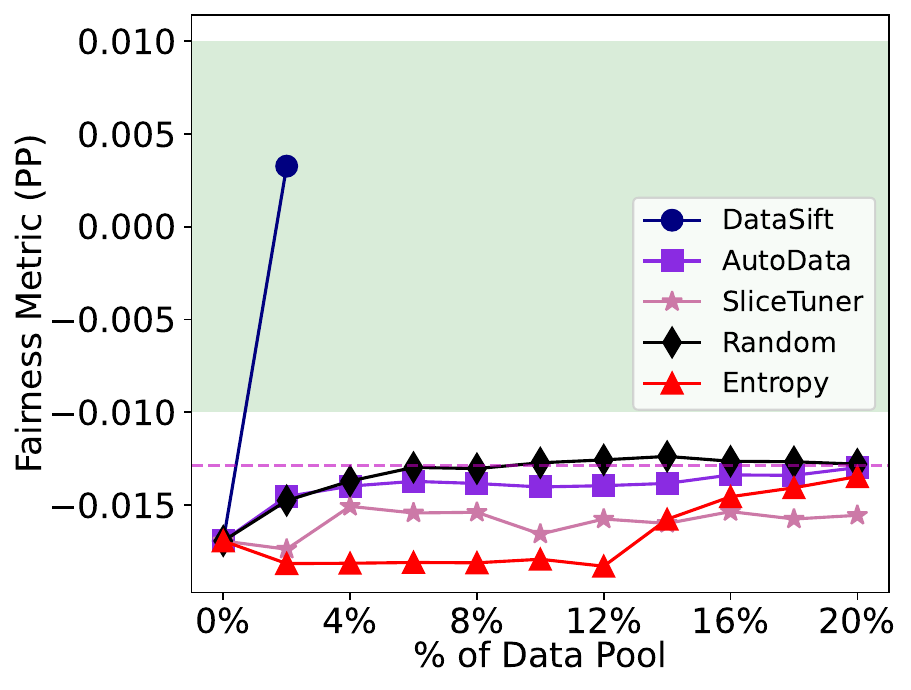}
        \caption{ACSMobility}
        \label{fig:PP_Mobility}
    \end{subfigure}

    \caption{\sys vs. baselines (Predictive Parity).}
    \label{fig: Datasift_PP_single_column}
\end{figure}

\noindent \textbf{Predictive Parity.}
We further evaluate \sys under \emph{predictive parity}, which requires that a model’s precision be equal across groups. Formally, predictive parity holds if, for all protected groups $A \in \{a,b\}$,
\[
\Pr(Y = 1 \mid \hat{Y} = 1, A = a) = \Pr(Y = 1 \mid \hat{Y} = 1, A = b),
\]
where $Y$ denotes the true label and $\hat{Y}$ denotes the predicted label. This criterion ensures that, among individuals receiving a positive prediction, the likelihood of being correctly classified is the same across groups. Figure~\ref{fig: Datasift_PP_single_column} reproduces the main experimental setting of Figure~3 using predictive parity as the fairness objective for two datasets. For the remaining datasets, the initial training data already meets the predefined fairness threshold, and thus the expansion process is omitted. The results demonstrate that \sys remains effective in reducing group-level disparities under this metric. As with  TPR parity , \sys consistently improves fairness under predictive parity while maintaining competitive predictive performance.

Taken together, these additional experiments confirm that the effectiveness of \sys is not tied to a specific fairness metric, rather it works as like a \emph{plug-and-play} method. By leveraging reward-driven partition selection and influence-based acquisition, the framework generalizes naturally across different definitions of group fairness, including outcome-based and error-rate–based criteria. This highlights the flexibility of \sys and its applicability in settings where different fairness notions may be required.

\subsection{Ablation Analysis}
\label{sec: Appendix_ablation}

\subsubsection{Datapool Distribution}
\label{sec: datapool_appendix}

To evaluate that \sys does not assume the data pool follows the same distribution as the initial training data, we conduct an additional sensitivity analysis by deliberately introducing distributional mismatches between the training set and the pool. In the original training data, the fraction of female samples is approximately $0.48$. We construct two alternative pool distributions by fixing the female fraction to $0.3$ and $0.8$, respectively, while keeping the training data unchanged. The corresponding results are shown in Appendix Figures~\ref{fig:gender_03} and~\ref{fig:gender_08}.

Across both scenarios, \sys consistently reduces the statistical parity gap and attains the target fairness threshold by selectively acquiring at most $10\%$ of the data pool, even when the pool distribution is substantially skewed relative to the training data. In contrast, the baseline methods exhibit slower convergence or unstable behavior under these distribution shifts. Notably, when the representation of the protected group is severely limited (Figure~\ref{fig:gender_03}), the baselines struggle to identify fairness-improving samples, whereas \sys is able to extract informative instances even from sparsely represented subpopulations. These results demonstrate that \sys does not depend on the data pool matching the training distribution, but instead dynamically adapts its acquisition strategy to the current model state.

We further assess robustness to shifts in non-sensitive attributes by varying the age distribution of the data pool. In particular, we consider two extreme scenarios in which the pool consists exclusively of individuals older than $50$ or exclusively of individuals younger than $50$, while the training data spans ages from $17$ to $94$. The corresponding results are reported in Appendix Figures~\ref{fig:age_gt_50} and~\ref{fig:age_lt_50}. Consistent with the gender-based experiments, \sys continues to achieve steady reductions in the fairness gap under both settings, demonstrating robustness to substantial changes in the pool’s feature composition.

Overall, these experiments confirm that \sys does not assume the data pool and training data share the same underlying distribution. Instead, the method leverages reward-driven partition selection and influence-based acquisition to remain effective even when the pool is biased, skewed, or otherwise distributionally different from the initial training set.

\begin{figure}[h]
    \centering

    \begin{subfigure}[b]{0.48\columnwidth}
        \centering
        \includegraphics[width=\linewidth]{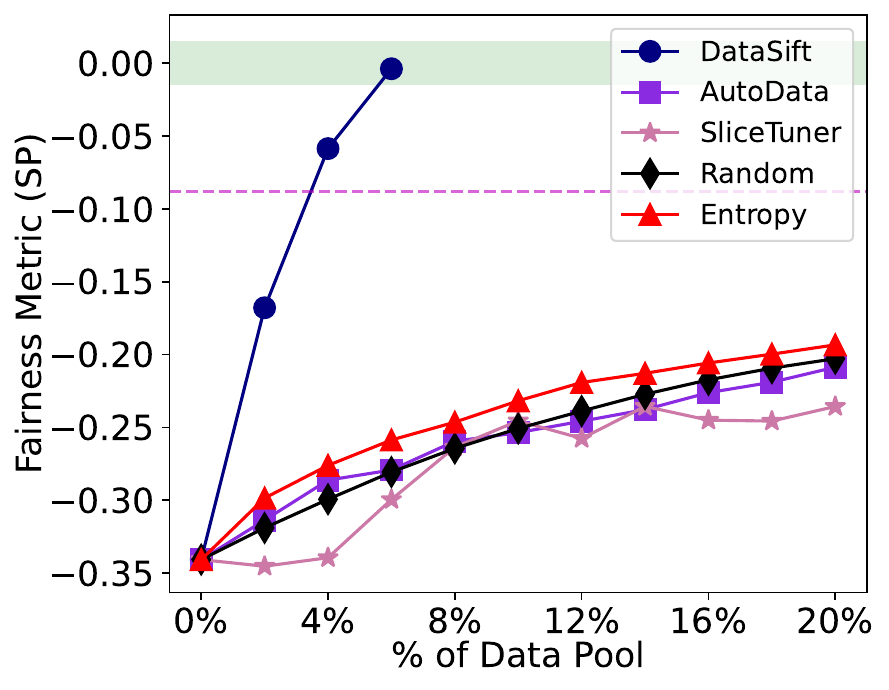}
        \caption{Protected fraction = 0.3, \\KL divergence = 0.19}
        \label{fig:gender_03}
    \end{subfigure}\hfill
    \begin{subfigure}[b]{0.48\columnwidth}
        \centering
        \includegraphics[width=\linewidth]{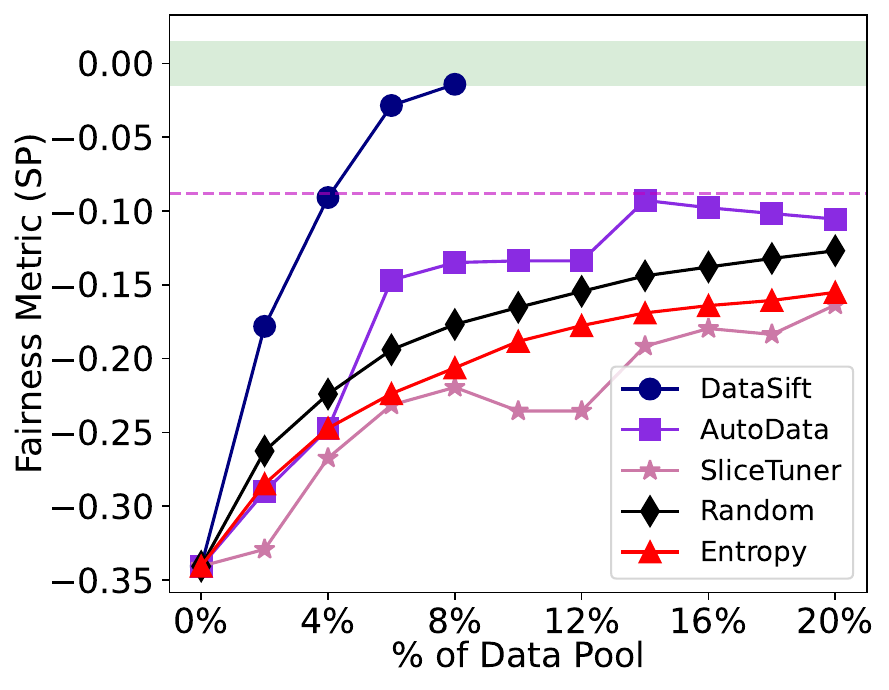}
        \caption{Protected fraction = 0.8, \\KL divergence = 0.26}
        \label{fig:gender_08}
    \end{subfigure}


    \begin{subfigure}[b]{0.48\columnwidth}
        \centering
        \includegraphics[width=\linewidth]{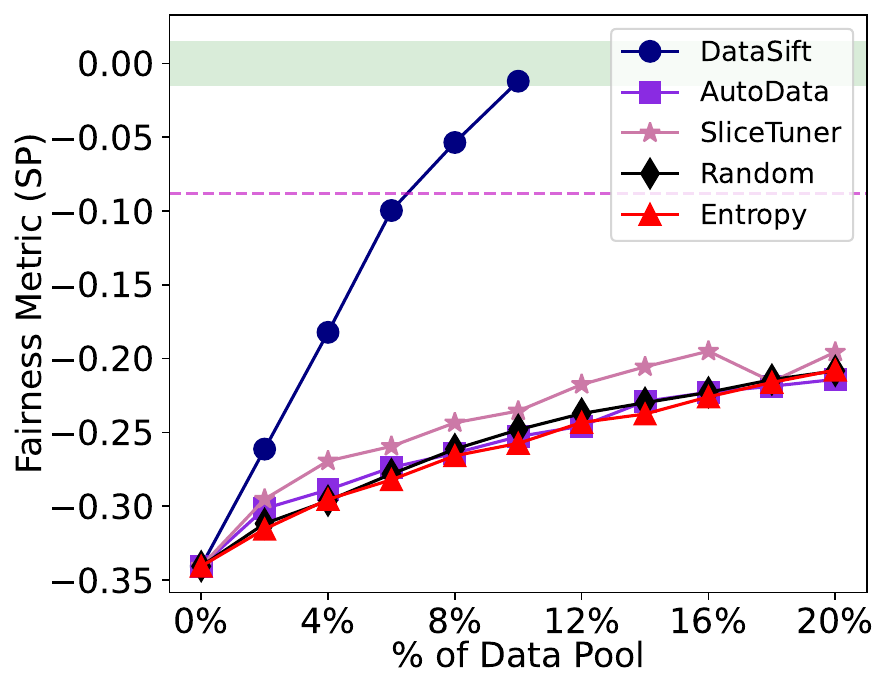}
        \caption{Age $>$ 50, \\ KL divergence = 0.96}
        \label{fig:age_gt_50}
    \end{subfigure}\hfill
    \begin{subfigure}[b]{0.48\columnwidth}
        \centering
        \includegraphics[width=\linewidth]{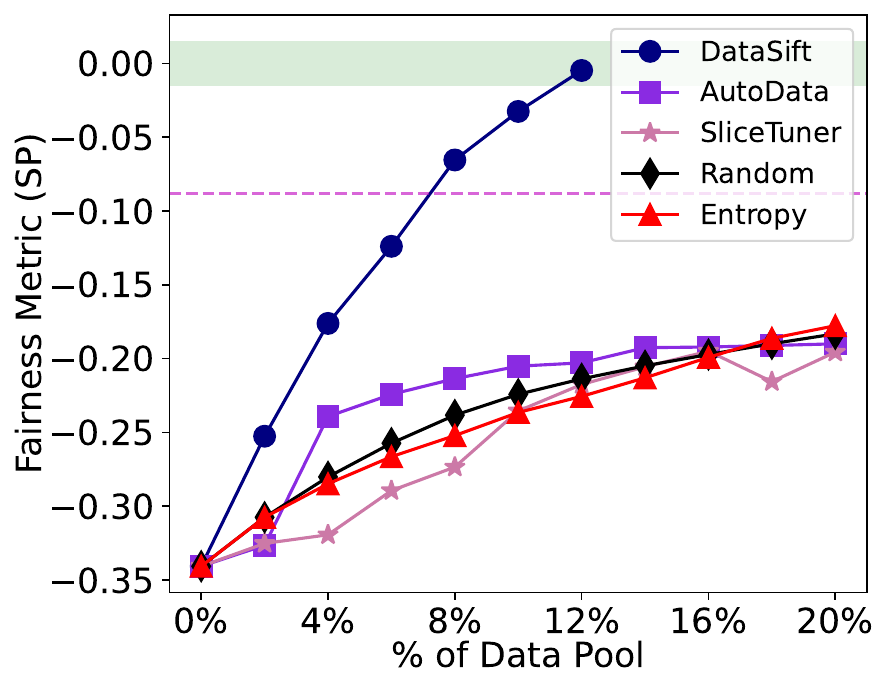}
        \caption{Age $<$ 50, \\KL divergence = 0.39}
        \label{fig:age_lt_50}
    \end{subfigure}

    \caption{Varying data pool distributions.}
    \label{fig:appendix_datapool_distributions}
\end{figure}

\clearpage

\end{document}